\PassOptionsToPackage{table,xcdraw}{xcolor}

\documentclass{uia}

\usepackage[utf8]{inputenc}
\usepackage{amsmath}
\usepackage{amssymb}
\usepackage{amsfonts}
\usepackage{graphicx}
\usepackage{url}
\usepackage[colorinlistoftodos,prependcaption,textsize=tiny]{todonotes}
\usepackage{cite}
\usepackage{color}
\usepackage{svg}
\usepackage{cite}
\usepackage{glossaries}
\usepackage{algorithm}
\usepackage{algpseudocode}
\usepackage{booktabs} 
\usepackage{multirow}
\usepackage[toc,page]{appendix}
\usepackage{pdfpages}
\usepackage{footnote}
\usepackage{hyperref}
\usepackage{lipsum}
\usepackage{emptypage}
\usepackage{titlesec}

\newcommand{\sectionbreak}{\clearpage}

\usepackage{tocloft}
\newcommand{\listpublicationname}{List of Publications}
\newlistof{publication}{pub}{\listpublicationname}
\newcommand{\publication}[1]{%
	\refstepcounter{publication}
	\addcontentsline{pub}{publication}
	{\protect\numberline{\thechapter.}#1}\par}

\hypersetup{
    colorlinks,
    citecolor=black,
    filecolor=black,
    linkcolor=black,
    urlcolor=black
}

\makeglossaries

\begin{document}
\pagenumbering{roman}
	
\newglossaryentry{GAN}{name={GAN},description={Generative Adversarial Networks}}
\newglossaryentry{DQN}{name={DQN},description={Deep Q-Network}}
\newglossaryentry{DCCGAQN}{name={DCCGAQN},description={Deep Conditional Convolutional Generative Adversarial Q Networks}}
\newglossaryentry{RL}{name={RL},description={Reinforcement Learning}}
\newglossaryentry{DRL}{name={DRL},description={Deep Reinforcement Learning}}
\newglossaryentry{RTS}{name={RTS},description={Real Time Strategy Games}}
\newglossaryentry{AI}{name={AI},description={Artificial Intelligence}}
\newglossaryentry{SDG}{name={SDG},description={Stochastic Gradient Decent}}
\newglossaryentry{DL}{name={DL},description={Deep Learning}}
\newglossaryentry{DNN}{name={DNN},description={Deep Neural Network}}
\newglossaryentry{MLP}{name={MLP},description={Multilayer Perceptron}}
\newglossaryentry{DDQN}{name={DDQN},description={Double Deep Q-Learning}}
\newglossaryentry{MDP}{name={MDP},description={Markov Decision Process}}
\newglossaryentry{CapsNet}{name={CapsNet},description={Capsule Network}}
\newglossaryentry{ConvNet}{name={ConvNet},description={Convolutional Neural Network}}
\newglossaryentry{ML}{name={ML},description={Machine Learning}}
\newglossaryentry{ANN}{name={ANN},description={Artificial Neural Network}}
\newglossaryentry{MCGS}{name={MCGS},description={Monte Carlo Graph Search}}
\newglossaryentry{MCTS}{name={MCTS},description={Monte Carlo Tree Search}}
\newglossaryentry{MCM}{name={MCM},description={Monte Carlo Methods}}
\newglossaryentry{ReLU}{name={ReLU},description={Rectified Linear Unit}}
\newglossaryentry{FCN}{name={FCN},description={Fully-Connected Network}}
\newglossaryentry{AGI}{name={AGI},description={Artificial General Intelligence}}
\newglossaryentry{CCDN}{name={CCDN},description={Conditional Convolution Deconvolution Network}}
\newglossaryentry{MSE}{name={MSE},description={Mean Squared Error}}
\newglossaryentry{CGAN}{name={CGAN},description={Conditional Generative Adversarial Network}}

%

\newcommand{\ThesisTitle}{Deep Reinforcement Learning using Capsules in Advanced Game Environments}
\title{\ThesisTitle}
\author{Per-Arne Andersen}
\faculty{Faculty of Engineering and Science}
\department{Department of ICT}
\copyrightnotice{All rights reserved}
\isbnprinted{}
\isbnelectronic{}
\serialnumber{}
\setyear{2018}
\setmonth{January}
\supervisor{Morten Goodwin\\Ole-Christoffer Granmo}
\date{\today}

\maketitle

\begin{abstract}
	Reinforcement Learning (\gls{RL}) is a research area that has blossomed tremendously in recent years and has shown remarkable potential for artificial intelligence based opponents in computer games. This success is primarily due to vast capabilities of Convolutional Neural Networks (\gls{ConvNet}), enabling algorithms to extract useful information from noisy environments. Capsule Network (\gls{CapsNet}) is a recent introduction to the Deep Learning algorithm group and has only barely begun to be explored. The network is an architecture for image classification, with superior performance for classification of the MNIST dataset. \gls{CapsNet}s have not been explored beyond image classification.
	
	\noindent
	\\
	This thesis introduces the use of \gls{CapsNet} for Q-Learning based game algorithms. To successfully apply \gls{CapsNet} in advanced game play, three main contributions follow. First, the introduction of four new game environments as frameworks for \gls{RL} research with increasing complexity, namely Flash RL, Deep Line Wars, Deep RTS, and Deep Maze. These environments fill the gap between relatively simple and more complex game environments available for \gls{RL} research and are in the thesis used to test and explore the \gls{CapsNet} behavior.
	
	\noindent
	\\
	Second, the thesis introduces a generative modeling approach to produce artificial training data for use in Deep Learning models including \gls{CapsNet}s. We empirically show that conditional generative modeling can successfully generate game data of sufficient quality to train a Deep Q-Network well.
	
	\noindent
	\\
	Third, we show that \gls{CapsNet} is a reliable architecture for Deep Q-Learning based algorithms for game AI. A capsule is a group of neurons that determine the presence of objects in the data and is in the literature shown to increase the robustness of training and predictions while lowering the amount training data needed. It should, therefore, be ideally suited for game plays. We conclusively show that capsules can be applied to Deep Q-Learning, and present experimental results of this method in the environments introduced. We further show that capsules do not scale as well as convolutions, indicating that \gls{CapsNet}-based algorithms alone will not be able to play even more advanced games without improved scalability.

\end{abstract}
\cleardoublepage


\tableofcontents

\printglossaries
\addcontentsline{toc}{chapter}{\numberline{}Glossary}%
\cleardoublepage

\listoffigures
\addcontentsline{toc}{chapter}{\numberline{}List of Figures}%
\cleardoublepage

\listoftables
\addcontentsline{toc}{chapter}{\numberline{}List of Tables}
\cleardoublepage

\addcontentsline{toc}{chapter}{\numberline{}List of Publications}
\listofpublication
\cleardoublepage

\pagenumbering{arabic}

\setlength{\parindent}{0em}
\setlength{\parskip}{1em}

\part{Research Overview}
\chapter{Introduction}
\section{Motivation}
Despite many advances in Artificial Intelligence (\gls{AI}) for games, no universal Reinforcement Learning (\gls{RL}) algorithm can be applied to advanced game environments without extensive data manipulation or customization. This includes traditional Real-Time Strategy (\gls{RTS}) games such as Warcraft III, Starcraft II, and Age of Empires. \gls{RL} has been applied to simpler games such as the Atari 2600 platform but is to the best of our knowledge not successfully applied to more advanced games. Further, existing game environments that target AI research are either overly simplistic such as Atari 2600 or complex such as Starcraft II.

\gls{RL} has in recent years had tremendous progress in learning how to control agents from high-dimensional sensory inputs like images. In simple environments, this has been proven to work well~\cite{Mnih2013}, but are still an issue for advanced environments with large state and action spaces~\cite{Mirowski2016}. In environments where the objective is easily observable, there is a short distance between the action and the reward which fuels the learning~\cite{Kaelbling1996}. This is because the consequence of any action is quickly observed, and then easily learned. When the objective is complicated, the game objectives still need to be mapped to a reward, but it becomes far less trivial~\cite{Konidaris2006}. For the Atari 2600 game Ms. Pac-Man this was solved through a hybrid reward architecture that transforms the objective to a low-dimensional representation~\cite{VanSeijen2017}. Similarly, the OpenAI's bot is able to beat world's top professionals at one versus one in DotA 2. It uses an \gls{RL} algorithm and trains this with self-play methods, learning how to predict the opponents next move.

Applying \gls{RL} to advanced environments is challenging because the algorithm must be able to learn features from a high-dimensional input, in order act correctly within the environment~\cite{Gupta2016}. This is solved by doing trial and error to gather knowledge about the mechanics of the environment. This process is slow and unstable~\cite{Mnih2015}. Tree-Search algorithms have been successfully applied to board games such as Tic-Tac-Toe and Chess, but fall short for environments with large state-spaces \cite{Ciardo1997}. This is a problem because the grand objective is to use these algorithms in real-world environments, that are often complex by nature. Convolutional Neural Networks (\gls{ConvNet})~\cite{Lecun1998} solves complexity problems but faces several challenges when it comes to interpreting the environment data correctly.




The primary motivation of this thesis is to create a foundation for \gls{RL}  research in advanced environments, Using generative modeling to train artificial neural networks, and to use the Capsule Network architecture in \gls{RL} algorithms.

\section{Thesis definition}
\label{sec:intro:definition}
The primary objective of this thesis is to perform \textbf{\ThesisTitle}. The research is split into six goals following the thesis hypotheses.

\subsection{Thesis Goals}

\textbf{Goal 1: } \textit{Investigate the state-of-the-art research in the field of Deep Learning, and learn how Capsule Networks function internally.}

\textbf{Goal 2: } \textit{Design and develop game environments that can be used for research into \gls{RL} agents for the \gls{RTS} game genre.}

\textbf{Goal 3: } \textit{Research generative modeling and implement an experimental architecture for generating artificial training data for games.}

\textbf{Goal 4: } \textit{Research the novel \gls{CapsNet} architecture for MNIST classification and combine this with \gls{RL} problems.}

\textbf{Goal 5: } \textit{Combine Deep-Q Learning and \gls{CapsNet} and perform experiments on environments from Achievement 2.}

\textbf{Goal 6: } \textit{Combine the elements of Goal 3 and Goal 5. The goal is to train an \gls{RL} agent with artificial training data successfully.}

\subsection{Hypotheses}

\textbf{Hypothesis 1: } \textit{Generative modeling using deep learning is capable of generating artificial training data for games with a sufficient quality.}

\textbf{Hypothesis 2: } \textit{\gls{CapsNet} can be used in Deep Q-Learning with comparable performance to \gls{ConvNet} based models.}

\subsection{Summary}
The first goal of this thesis is to create a learning platform for \gls{RTS} game research. Second, to use generative modeling to produce artificial training data for \gls{RL} algorithms. The third goal is to apply \gls{CapsNet}s to Deep Reinforcement Learning algorithms. The hypothesis is that its possible to produce artificial training data, and that \gls{CapsNet}s can be applied to Deep Q-Learning algorithms.

\section{Contributions}
This thesis introduces four new game environments, \textit{Flash RL}\footnote{Proceedings of the \(30^{th}\) Norwegian Informatics Conference, Oslo, Norway 2017}, \textit{Deep Line Wars}\footnote{Proceedings of the \(37^{th}\) SGAI International Conference on Artificial Intelligence, Cambridge, UK, 2017}, \textit{Deep RTS}, and \textit{Deep Maze}. These environments integrates well with OpenAI GYM, creating a novel learning platform that targets \textit{Deep Reinforcement Learning for Advanced Games}.

\gls{CapsNet} is applied to \gls{RL} algorithms and provides new insight on how \gls{CapsNet} performs in problems beyond object recognition. This thesis presents a novel method that use generative modeling to train \gls{RL} agents using artificial training data.

There is to the best of our knowledge no documented research on using \gls{CapsNet} in \gls{RL} problems, nor are there environments specifically targeted \gls{RTS} \gls{AI} research.

\section{Thesis outline}
Chapter~\ref{chap:bg} provides preliminary background research for Artificial Neural Networks (\ref{sec:bg:ann}, \ref{sec:bg:convnet}), Generative Models~(\ref{sec:bg:generative}), Markov Decision Process (\ref{sec:bg:mdp}), and Reinforcement Learning (\ref{sec:bg:rl}).

Chapter~\ref{chap:sota} investigates the current state-of-the-art in Deep Neural Networks (\ref{sec:sota:nn}), \gls{RL} (\ref{sec:sota:rl}), \gls{GAN} (\ref{sec:sota:gan}) and Game environments (\ref{sec:sota:gameenv}).

Chapter~\ref{chap:env} outlines the technical specifications for the new game environments Flash RL (\ref{sec:env:flashrl}), Deep Line Wars (\ref{sec:env:deeplinewars}), Deep RTS (\ref{sec:env:deeprts}), and Maze (\ref{sec:env:maze}). In addition, a well established game environment (Section~\ref{sec:env:flappybird}) is introduced to validate experiments conducted in this thesis.

Chapter~\ref{chap:solutions} introduces the proposed solutions for the goals defined in Section~\ref{sec:intro:definition}. Section~\ref{sec:solutions:environments} outlines how the environments are presented as a learning platform. Section~\ref{sec:solutions:capsnet} introduces the proposal to use Capsules in \gls{RL}. Section~\ref{sec:solutions:dqn} describes the Deep Q-Learning algorithm and the implementations used for the experiments in this thesis. Finally, the artificial training data generator is outlined in Section~\ref{sec:solutions:datagen}.

Chapter~\ref{chap:results:ccdn} and \ref{chap:results:dqn} shows experimental results from the work presented in Chapter~\ref{chap:solutions}. 

Chapter~\ref{chap:conclusion} concludes the thesis hypotheses and provides a summary of the work done in this thesis. Section~\ref{sec:conclusion:future_work} outlines the road-map for future research related to the thesis.

\chapter{Background}
\label{chap:bg}
Deep Learning (\gls{DL}) is a branch of machine learning algorithms that recently became popularized due to the exponential growth in available computing power. \gls{DL} is unique in that it is designed to learn data representations, as opposed to task-specific algorithms. Methods from \gls{DL} are frequently used in \gls{RL} algorithms, creating a new branch called \textit{Deep Reinforcement Learning} (\gls{DRL}). Artificial Neural Networks (\gls{ANN}) are used at its core, utilizing the most novel \gls{DL} techniques to gain state-of-the-art capabilities.

This chapter outlines background theory for topics related to the research performed later in this thesis. Section \ref{sec:bg:ann} shows how Artificial Neural Networks work, moving onto computer vision with Convolutional Neural Networks in Section \ref{sec:bg:convnet}. Section~\ref{sec:bg:mdp} outlines the theory behind the Markov Decision Process (\gls{MDP}) and how it is used in \gls{RL}.

\section{Artificial Neural Networks}
\label{sec:bg:ann}

\begin{figure}[!htb]
	\centering
	\includegraphics[width=0.90\linewidth]{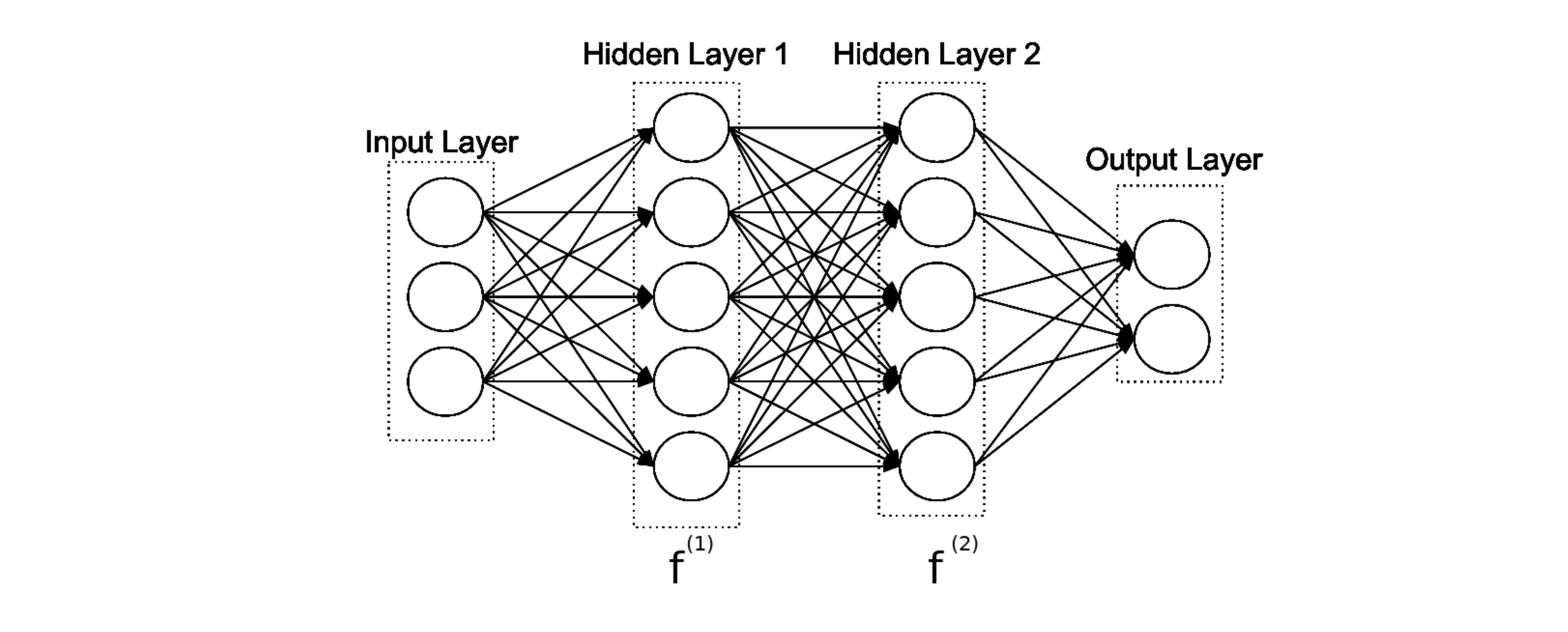}
	\caption{Deep Neural network with two hidden layers}
	\label{fig:ffdnn}
\end{figure}

An Artificial Neural Network (\gls{ANN}) is a computing system that is inspired by how the biological nervous systems, such as the brain, function~\cite{Goodfellow2016}. \gls{ANN}s are composed of an interconnected network of neurons that pass data to its next layer when stimulated by an activation signal. When a network consists of several hidden layers, it is considered a \textit{Deep Neural Network} (\gls{DNN}). Figure~\ref{fig:ffdnn} illustrates a Deep Multi-Layer Perceptron (\gls{MLP}) with two hidden layers.

\begin{figure}[!htb]
	\centering
	\includegraphics[width=0.40\linewidth]{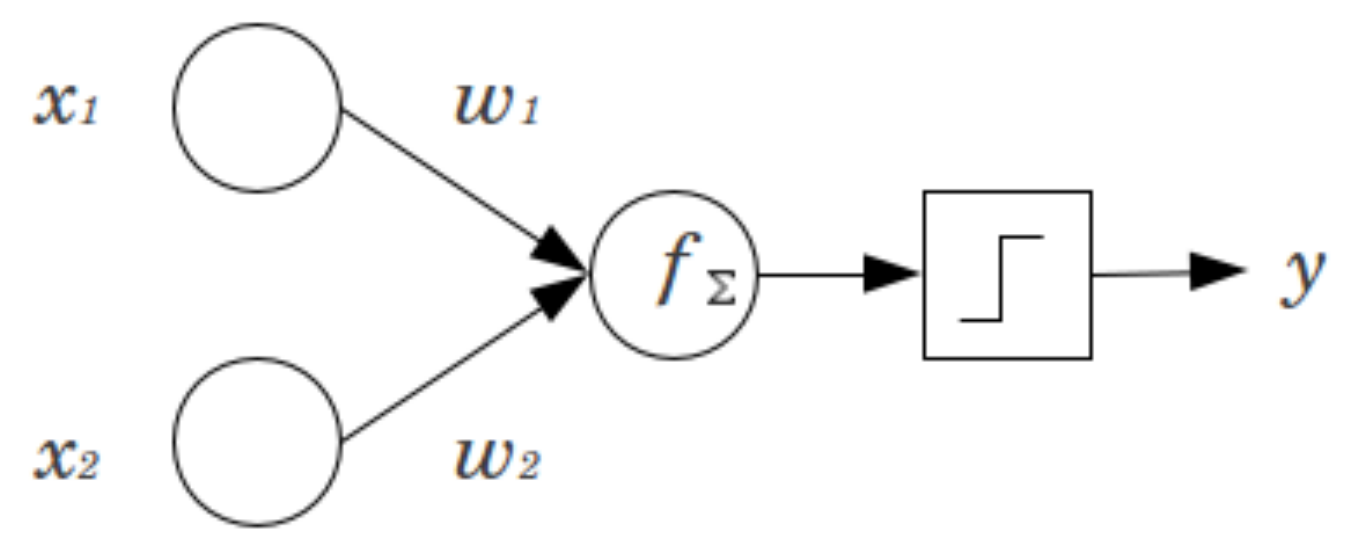}
	\caption{Single Perceptron}
	\label{fig:perceptron}
\end{figure}

\begin{equation}
	f(x) = \left\{\begin{matrix}
	1  & \textup{if} & \sum_{i=1}^{n} (w_i\cdot x_i) + b &  > 0&\\ 
	0  & \textup{otherwise}
	\end{matrix}\right.
	\label{eq:binary_perceptron}
\end{equation} 

\gls{MLP}s are considered a network because they are composed of many different functions. Each of these functions is represented as a \textit{perceptron}. The combination of these functions gives us the ability to represent complex and high-dimensional functions~\cite{Goodfellow2016}. Figure~\ref{fig:perceptron} illustrates a single perceptron from an \gls{MLP} where \(x_1, x_2 \cdots x_n\) are inputs to the perceptron. Each of these inputs has a weight \(w_1, w_2 \cdots w_n\). Input \(x_n\) and weight \(w_n\) are multiplied into \(z_n = x_n \cdot w_n\) and \(z = \sum_{i=1}^{n}(z_n) + b\) where \(b\) is the bias value and \(z\) is the perceptron value. In Figure~\ref{fig:perceptron}, the perceptron has a binary activation function (Equation~\ref{eq:binary_perceptron}), the neuron produce the value 1 for all \(z\) above 1, and 0 otherwise. There are several different activation functions that can be used in a perceptron network, see Section~\ref{sec:bg:ann:activation}.

\subsection{Activation Functions}
\label{sec:bg:ann:activation}
{\renewcommand{\arraystretch}{2.5}
\begin{table}[]
	\centering
	\begin{tabular}{l|l}
	\textbf{Name}                & \textbf{Equation} \\ \bottomrule
	TanH                            & \(\tanh(z) = \frac{2}{1 + e^{-2z}}-1\)\\
	Softmax                          & \(\sigma (z)_{j} = \frac{e^{z_{j}}}{\sum_{k=1}^{K}e^{z_{k}}} \text{for}~j=1\cdots K\) \\
	Sigmoid                      & \(f(z) = \frac{1}{1+e^{-z}}\) \\
	Rectified Linear Unit (\gls{ReLU})    & \(f(z) = \begin{cases}
		0 & \text{ for } z < 0 \\ 
		z & \text{ otherwise }
		\end{cases}\) \\
	LeakyReLU                       & \(f(z) = \begin{cases}
		z & \text{ if } z > 0 \\ 
		0.01z & \text{ otherwise }
		\end{cases}\) \\
	Binary                         & \(f(z) = \begin{cases}
		0 & \text{ if } z < 0 \\ 
		1 & \text{ if } z \geq 0
		\end{cases}\) \\
	\end{tabular}
	\caption{Equations of activation function}
	\label{tbl:activation_functions}
\end{table}}
The purpose of an \textit{Activation function} is often to introduce non-linearity into the network. It is proven that an \gls{DNN} using only linear activations are equal to a single-layered network~\cite{Pralyt1994}. It is therefore natural to use non-linear activation functions in the hidden layers of an \gls{ANN} if the goal is to predict non-linear functions. TanH and Rectified Linear Unit (\gls{ReLU}) has proven to work well in \gls{ANN}s~\cite{Nair2010, Yann1998, LeCun1991}, but there exist several other alternatives as illustrated in Table~\ref{tbl:activation_functions}. Researchers do not understand to the full extent why an activation function works better for a particular problem and is why trial and error is used to find the best fit~\cite{Radford2015}.

\subsection{Optimization}
\label{sec:bg:ann:optimization}
Optimization in \gls{ANN}s is the process of updating the weights of neurons in a network. In the optimization process, a \textit{loss function} is defined. This function calculates the error/cost value of the network at the output layer. The error value describes the distance between the ground truth and the predicted value. For the network to improve, this error is \textit{backpropagated} back through the network until each neuron has an error value that reflects its positive or negative contribution to the ground truth. Each neuron also calculates the gradient of its weights by multiplying output delta together with the input activation value. Weights are updated using \textit{stochastic gradient descent} (\gls{SDG}), which is a method of gradually descending the weight loss until reaching the optimal value.

\subsection{Loss Functions}
To measure the inconsistency between the predicted value and the ground truth, a loss function is used in \gls{ANN}s. The loss function calculates a positive number that is minimized throughout the optimization of the parameters\footnote{Weights and Parameters are used interchangeably throughout the thesis} (Section~\ref{sec:bg:ann:optimization}). A loss function can be any mathematical formula, but there exist several well established functions. The performance varies on the classification task.

\begin{figure}
	\centering
	\includegraphics[width=0.99\linewidth]{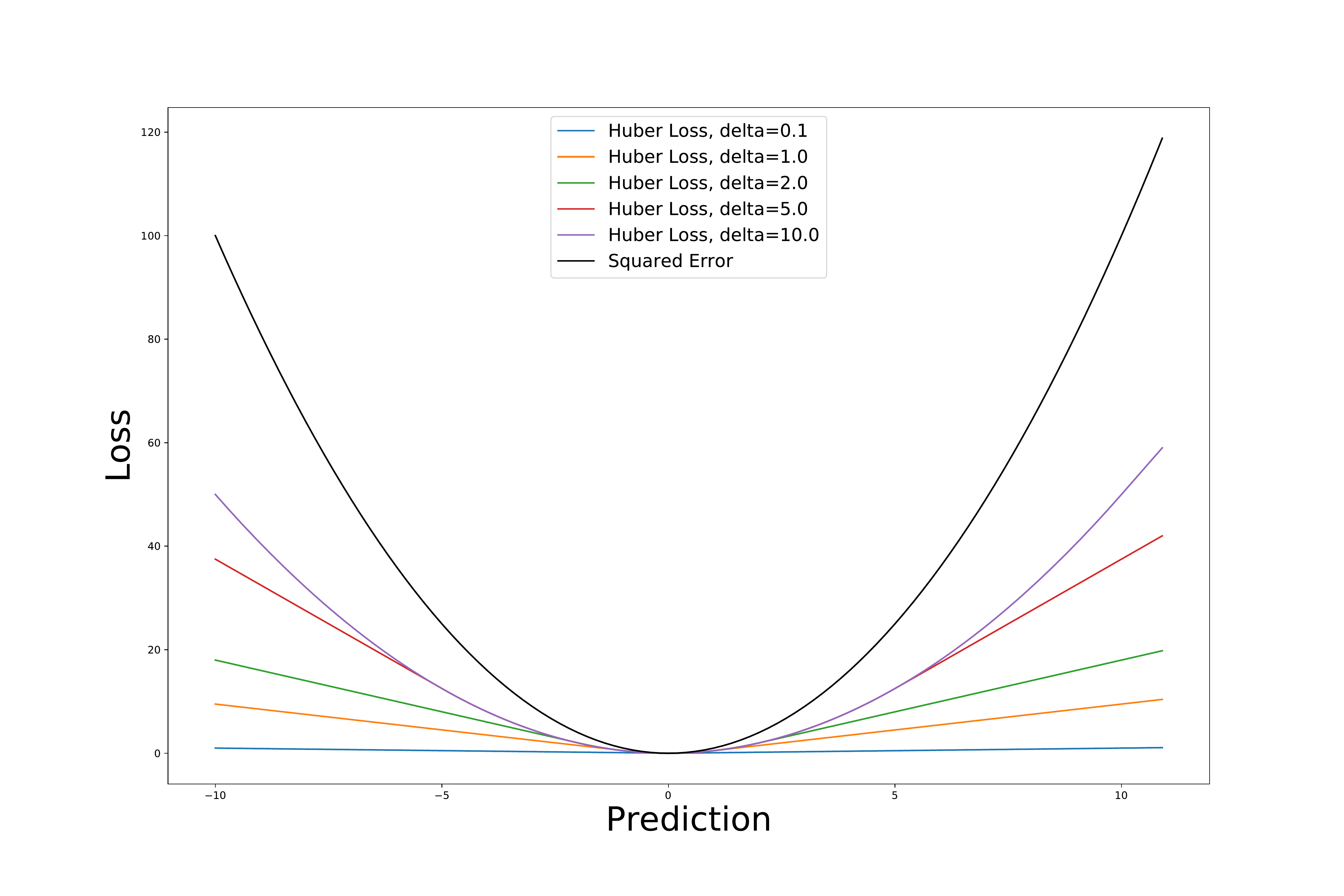}
	\caption{Loss functions}
	\label{fig:loss_functions}
\end{figure}

\textbf{Mean Squared Error} (\gls{MSE}) is a quadratic loss function widely used in linear regression, and are also used in this thesis. Equation~\ref{eq:loss:mse} is the standard form of \gls{MSE}, where the goal is to minimize the residual squares \((y^{(i)} - \hat{y}^{(i)})\).
\begin{equation}
L = \frac{1}{n}\sum_{i=1}^{n}(y^{(i)} - \hat{y}^{(i)})^{2}
\label{eq:loss:mse}
\end{equation}

\begin{equation}
L_{\delta }(a) = \left\{\begin{matrix}
\frac{1}{2}a^{2} & \text{for} \left | a \right | \leq \delta ,\\ 
\delta (\left | a \right | - \frac{1}{2}\delta ), & \text{otherwise} 
\end{matrix}\right.
\label{eq:loss:huber}
\end{equation}

\textbf{Huber Loss} is a loss function that is widely used in \gls{DRL}. It is similar to \gls{MSE}, but are less sensitive to data far apart from the ground truth. Equation~\ref{eq:loss:huber} defines the function where \textit{a} refers to the residuals and \(\delta\) refers to its sensitivity. Figure~\ref{fig:loss_functions} illustrates the difference between \gls{MSE} and Huber Loss using different \(\delta\) configurations.

\subsection{Hyper-parameters}
Hyper-parameters are tunable variables in \gls{ANN}s. These parameters include learning rate, learning rate decay, loss function, and optimization algorithm like Adam, and \gls{SDG}.

\section{Convolutional Neural Networks}
\label{sec:bg:convnet}
A Convolutional Neural Network is a novel \gls{ANN} architecture that primarily reduces the compute power required to learn weights and biases for three-dimensional inputs. 
\gls{ConvNet}s are split into three layers:
\begin{enumerate}
	\item Convolution layer
	\item Activation layer
	\item Pooling (Optional)
\end{enumerate} 
A Convolution layer has two primary components, \textit{kernel} (parameters) and \textit{stride}. The kernel consists of a weight matrix that is multiplied by the input values in its \textit{receptive field}. The receptive field is the area of the input that the kernel is focused on. The kernel then slides over the input with a fixed stride. The stride value determines how fast this sliding happens. With a stride of 1, the receptive field move in the direction \(x + 1\), and when at the end of the input x-axis, \(y + 1\).

\begin{figure}[!hb]
	\centering
	\includegraphics[width=0.90\linewidth]{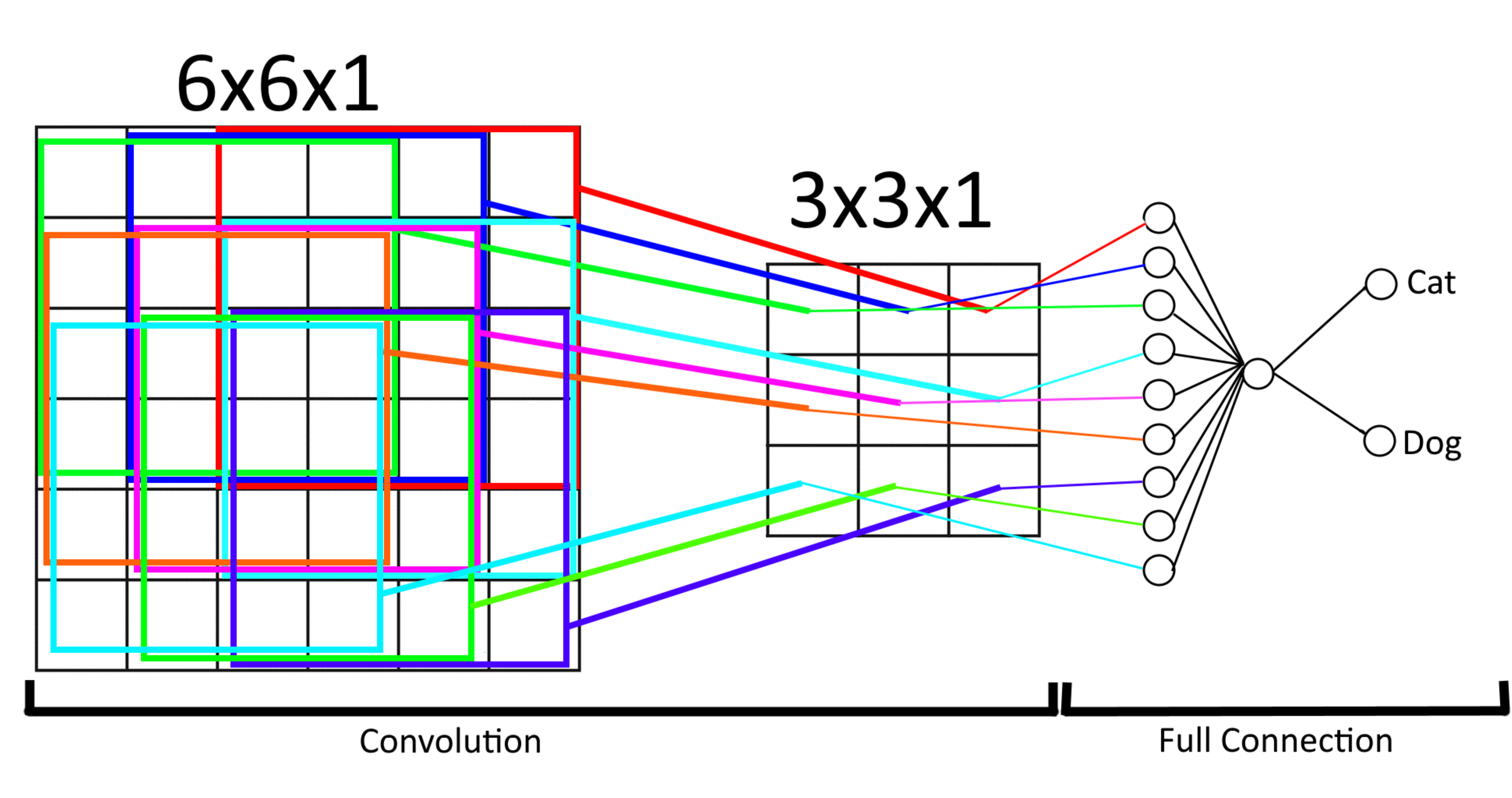}
	\caption{Convolutional Neural Network for classification}
	\label{fig:6x6_cnn}
\end{figure}

\begin{figure}[!hb]
	\centering
	\includegraphics[width=0.90\linewidth]{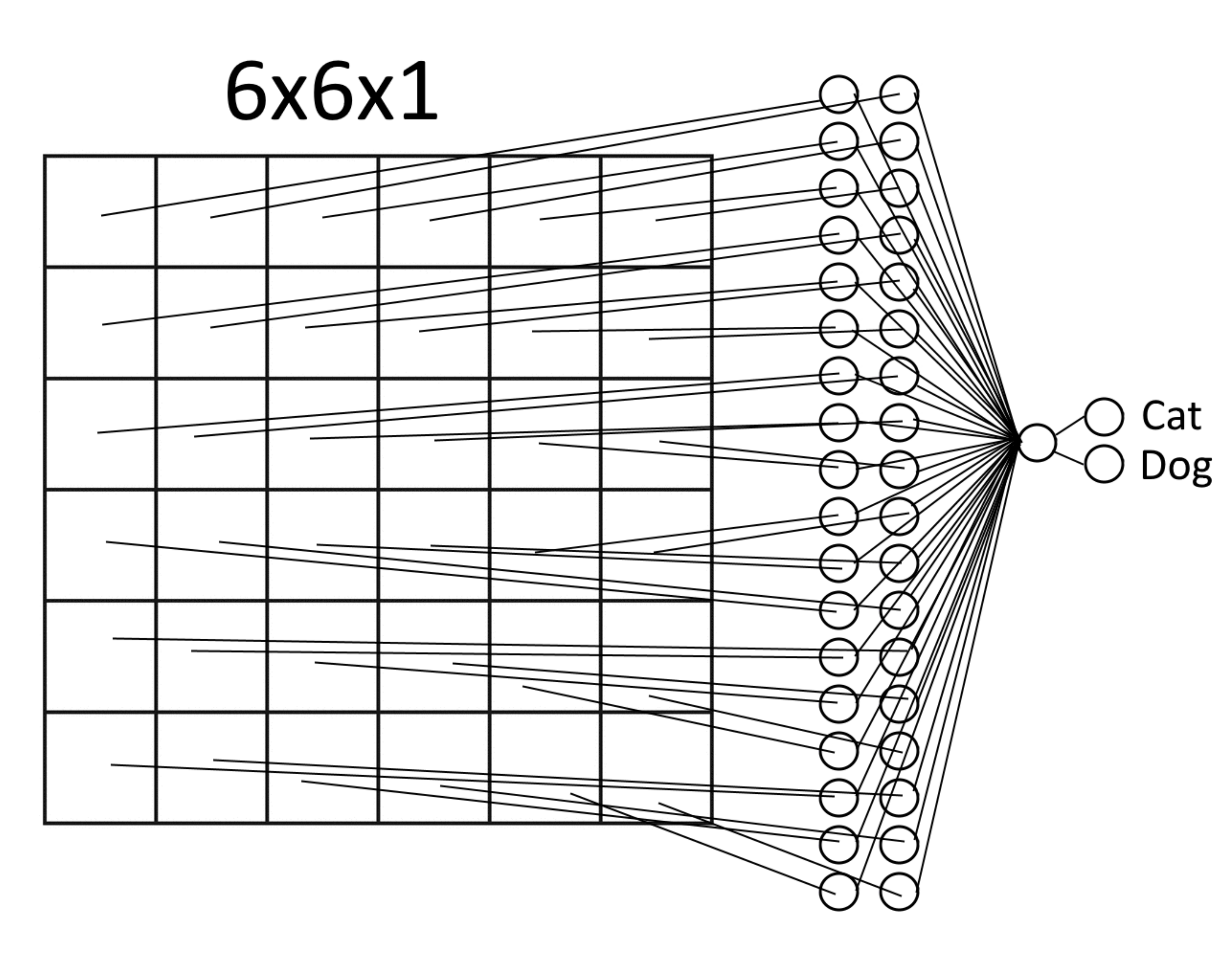}
	\caption{Fully-Connected Neural Network for classification}
	\label{fig:6x6_fc}
\end{figure}

Consider a three-dimensional matrix representing an image of size \(28 \times 28 \times 3\). In this example, the goal is to classify the image to be either a cat or dog. By using hyperparameters \(kernel = 3 \times 3\) and \(stride = 1 \times 1\), there are 32 \textit{shared} parameters to be optimized. In contrast, a Fully-Connected network (\gls{FCN}) with a single neuron layer, would have 2357 parameters to optimize. The reason why convolutions work is that it exploits what is called \textit{feature locality}. \gls{ConvNet}s use \textit{filters} that learn a specific feature of the input, for example, horizontal and vertical lines. For every convolutional layer added to the network, the information becomes more abstract, identifying objects and shapes. Figures~\ref{fig:6x6_cnn} and~\ref{fig:6x6_fc} illustrate how a simple \gls{ConvNet} is modeled compared to an \gls{FCN}. The \gls{ConvNet} use a stride of \(1 \times 1\) and a kernel size of \(4 \times 4\) yielding a \(3 \times 3\) output. This produces a total of 31 parameters to optimize, compared to 41 parameters in the \gls{FCN}.

\subsection{Pooling}
\textit{Pooling} is the operation of reducing the data resolution, often subsequent a convolution layer. This is beneficial because it reduces the number of parameters to optimize, hence decreasing the computational requirement. Pooling also controls overfitting by generalizing features. This makes the network capable of better handling spatial invariance~\cite{Scherer2010}.

\begin{figure}[!h]
	\centering
	\includegraphics[width=0.90\linewidth]{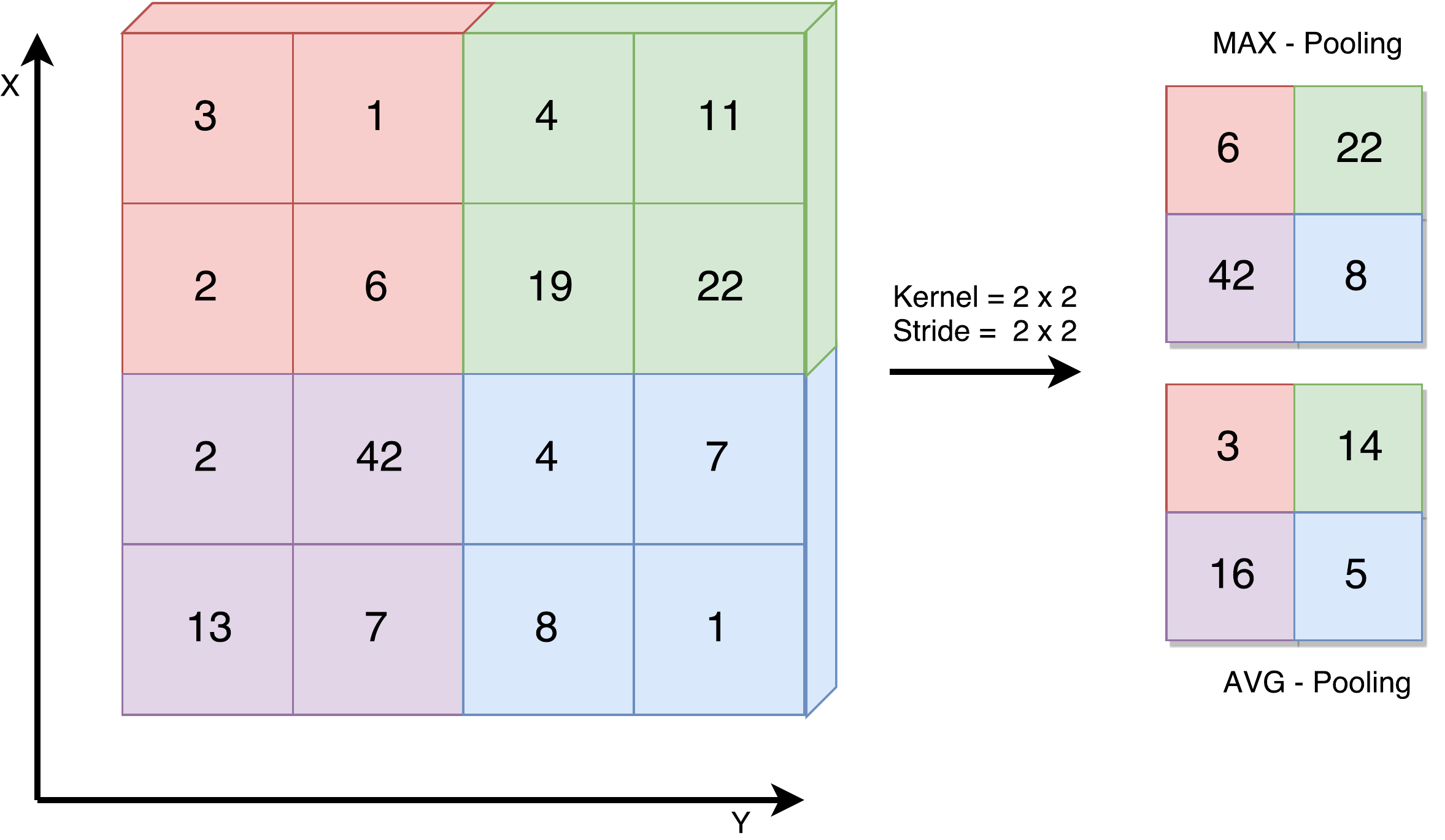}
	\caption{MAX and AVG Pooling operation}
	\label{fig:max_avg_pooling}
\end{figure}

There are several ways to perform pooling. \textit{Max} and \textit{Average} pooling are considered the most stable methods in whereas Max pooling is most used in state-of-the-art research~\cite{Lee2015}. Figure~\ref{fig:max_avg_pooling} illustrates the pooling process using Max and Average pooling on a \(4 \times 4 \times X\)\footnote{\(X = \)Depth of the input volume} input volume. The hyperparameters for the pooling operation is \(kernel = 2 \times 2\) and \(stride = 2 \times 2\) applied to the input vector yields the resulting \(2 \times 2 \times X\) output volume. This operation performed independently for each depth slice of the input volume.

\subsection{Summary}
Historically, \gls{ConvNet}s drastically improved the performance of image recognition because it successfully reduced the number of parameters required, and at the same time preserving important features in the image. There are however several challenges, most notably that they are not rotation invariant. \gls{ConvNet}s are much more complicated then covered in this section, but this beyond the scope of this thesis. For an in-depth survey of the \gls{ConvNet} architecture, refer to \href{https://arxiv.org/abs/1512.07108}{Recent Advances in Convolutional Neural Networks}~\cite{Gu2015}.

\section{Generative Models}
\label{sec:bg:generative}
Generative Models are a series of algorithms trying to generate an artificial output based on some input, often randomized. Generative Adversarial Networks and Variational Autoencoder is two methods that have shown excellent results in this task. These methods have primarily been used in generating realistic images from various datasets like MNIST and CIFAR-10. This section will outline the theory in understanding the underlying architecture of generative models.

\begin{figure}
	\centering
	\includegraphics[width=\linewidth]{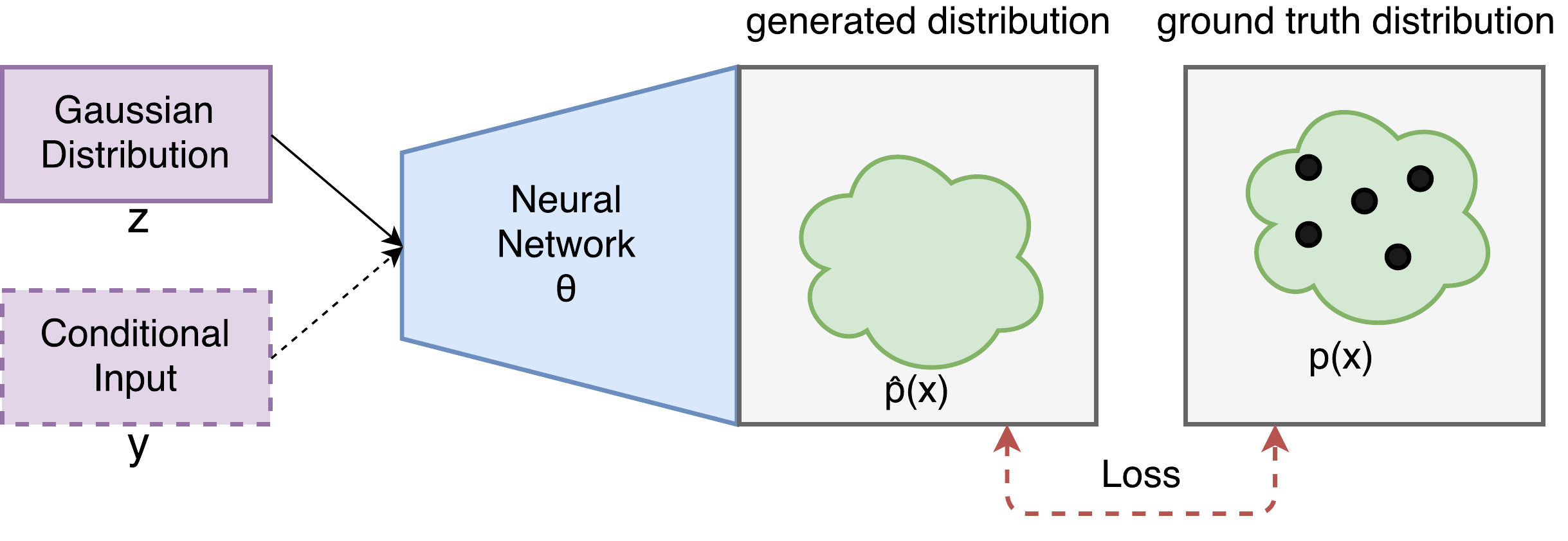}
	\caption{Overview: Generative Model}
	\label{fig:generative_models}
\end{figure}

The objective of most Generative Models is to generate a distribution of data, that is close to the ground-truth distribution (the dataset). The Generative Model takes a Gaussian distribution \(z\), as input, and outputs \(\hat{p}(x)\) as illustrated in Figure \ref{fig:generative_models}. The goal is to find parameters \(\theta\) that best matches the ground truth distribution with the generated distribution. Convolutional Neural Networks are often used in Generative Modeling, typically for models using noise as input. The model has several hidden parameters \(\theta\) that is tuned via backpropagation methods like stochastic gradient descent. 
If the model reaches optimal parameters, \(\hat{p}(x) = p(x)\) is considered true.

\section{Markov Decision Process}
\label{sec:bg:mdp}
\gls{MDP} is a mathematical method of modeling decision-making within an \textit{environment}. An environment defines a real or virtual world, with a set of rules. This thesis focuses on virtual environments, specifically, games with the corresponding game mechanic limitations. The core problem of \gls{MDP}s is to find an optimal \textit{policy} function for the decision maker (hereby referred to as an agent).
\begin{equation}
	\underbrace{a}_{\text{Action }}= \underbrace{\pi(s)}_{\text{Policy}~\boldsymbol{\pi}~\text{for state }\boldsymbol{s}}
	\label{eq:a_pi_s}
\end{equation}
Equation~\ref{eq:a_pi_s} illustrates how a decision/action is made using observed knowledge of the environmental state. The goal of the policy function is to find the decision that yields the best cumulative reward from  the environment. \gls{MDP} behaves like a Markov chain, hence gaining the \textit{Markov Property}. The Markov property describes a system where future states only depend on the present and not the past. This enables \gls{MDP} based algorithms to do iterative learning~\cite{Bellman1957}. \gls{MDP} is the foundation of how \gls{RL} algorithms operate to learn the optimal behavior in an environment.

\section{Reinforcement Learning}
\label{sec:bg:rl}
Reinforcement learning is a process where an agent performs actions in an environment, trying to maximize some cumulative reward~\cite{Sutton1998} (see Section~\ref{sec:bg:mdp}). \gls{RL} differs from supervised learning because the ground truth is never presented directly. In \gls{RL} there are \textit{model-free} and \textit{model-based} algorithms. In model-free \gls{RL}, the algorithm must learn the environmental properties (the model) without guidance. In contrast, model-based \gls{RL} is defined manually describing the features of an environment \cite{doya2002}. For model-free algorithms, the learning only happens in present time and the future must be \textit{explored} before knowledge about the environment can be learned~\cite{Watkins1989, Even-dar2003}.

This thesis focuses on \textit{Q-Learning} algorithms, a model-free \gls{RL} technique that may potentially solve difficult game environments. This section investigates the background theory of Q-Learning and extends this method to Deep Q-Learning (\gls{DQN}), a novel algorithm that combines \gls{RL} and \gls{ANN}.

\subsection{Q-Learning}
\label{sec:bg:rl:ql}
Q-Learning is a model-free algorithm. This means that the MDP stays hidden throughout the learning process. The objective is to learn the optimal policy by estimating the action-value function \(Q^*(s, a)\), yielding maximum expected reward in state \textit{s} performing action \textit{a} in an environment. The optimal policy can then be found by 

\begin{equation}
\pi(s) = argmax_aQ^*(s,a)
\label{eq:optimal_policy}
\end{equation}
Equation~\ref{eq:optimal_policy} is derived from finding the optimal utility of a state \(U(s) = \max_{a}Q(s,a)\). Since the utility is the maximum value, the argmax of that same value qualifies as the optimal policy. The update rule for Q-Learning is based on value iteration:

\begin{equation}
\label{eq:qupdate}
Q(s,a) \leftarrow Q(s,a) + \underbrace{\alpha}_\text{LR} \Bigg( \underbrace{R(s)}_\text{Reward} + \underbrace{\gamma}_\text{Discount} \underbrace{max_{a^{'}} Q(s^{'},a^{'})}_\text{New Q} - \underbrace{Q(s,a)}_\text{Old Q}  \Bigg)
\end{equation}

Equation~\ref{eq:qupdate} shows the iterative process of propagating back the estimated Q-value for each discrete time-step in the environment. 
\(\alpha\) is the learning rate of the algorithm, usually low number between 0.001 and 0.00001. The reward function \(R(s) \in \mathbb{R}\), and is often between \(-1 < x < 1\) to increase learning stability. \(\gamma\) is the discount factor, discounting the importance of future states. The "old Q" is the estimated Q-Value of the starting state while the "new Q" estimates the future state. Equation~\ref{eq:qupdate} is guaranteed to converge towards the optimal action-value function, \(Q_i \rightarrow Q^*\) as i $\rightarrow \infty$ \cite{Sutton1998, Mnih2013}.

\subsection{Deep Q-Learning}
\label{sec:bg:rl:dql}
At the most basic level, Q-Learning utilizes a table for storing $(s,a,r,s^{'})$ pairs. Instead, a non-linear function approximation can be used to approximate $Q(s,a;\theta)$. This is called \textbf{Deep-Q Learning}. $\theta$ describes tunable parameters (weights) for the approximation.\gls{ANN}s are used as an approximation method for retrieving values from the Q-Table but at the cost of stability. Using \gls{ANN} is much like compression found in JPEG images. The compression is \textit{lossy}, and information is lost at compression time. This makes \gls{DQN} unstable, since values may be wrongfully encoded under training. In addition to value iteration, a loss function must be defined for the backpropagation process of updating the parameters.

\begin{equation}
\label{eq:loss_function}
L(\theta_{i}) = E\Big[(r + \gamma max_{a^{'}}Q(s^{'},a^{'}; \theta_{i}) - Q(s,a;\theta_{i}))^2\Big]
\end{equation}
Equation~\ref{eq:loss_function} illustrates the loss function proposed by Minh et al \cite{Mnih2015}. It uses Bellmans equation to calculate the loss in gradient descent.
To increase training stability, \textit{Experience Replay} is used. This is a memory module that store memories from already explored parts of the state space. Experiences are often selected at random and then replayed to the neural network as training data. \cite{Mnih2013}.

\chapter{State-of-the-art}
\label{chap:sota}
This thesis focus on topics that are in active research, meaning that the state-of-the-art methods quickly advances. There have been many achievements in Deep Learning, primarily related to Computer Vision topics. This chapter investigates recent advancements in Deep Learning (\ref{sec:sota:nn}), Deep Reinforcement Learning (\ref{sec:sota:rl}), Generative Modeling (\ref{sec:sota:gan}), Capsule Networks (\ref{sec:sota:capsnet}) and Game Learning Platforms (\ref{sec:sota:gameenv}). In the success of Deep Learning, there have been several breakthroughs in popular game environments. Section~\ref{sec:sota:games} outlines the state-of-the-art of applying \gls{RL} algorithms to game environments. 

\section{Deep Learning}
\label{sec:sota:nn}
Deep Learning has a long history, dating back to late 1980's. One of the first relevant papers on the area is \textit{Learning representations by backpropagating errors} from Rumelhart et al.~\cite{Rumelhart1986} In this paper, they illustrated that a deep neural network could be trained using backpropagation. The deep architecture proved that a neural network could successfully learn non-linear functions.

Yann LeCun started in the early 1990's research into Convolutional Neural Networks (ConvNet), with handwritten zip code classification as the primary goal~\cite{LeCun1989}. He created the famous MNIST dataset, which is still widely used in the literature~\cite{Lecun1998}. After ten years of research, LeCun et al. achieved state-of-the-art results on the MNIST dataset using \gls{ConvNet}s similar to those found in literature today~\cite{Lecun1998}. But due to scaling issues with Deep \gls{ANN}s, they were outperformed by classifiers like Support Vector Machines. It was not until 2006 with the paper \textit{A fast learning algorithm for deep belief nets} by Hinton et al. that Deep Learning would appear again~\cite{Hinton2006}. This paper showed how ectively train a deep neural network, by training one layer at a time. This was the beginning of Deep Neural Networks as they are known today.


For this thesis, Computer Vision is the most interesting architecture. There have been many advances in computer vision in the last couple of years. AlexNet~\cite{Krizhevsky2012}, VGGNet~\cite{Simonyan2014} and ResNet~\cite{He2015} are models achieving state-of-the-art results in the ImageNet competition. These models are complex, but does a good job in image recognition. For \gls{DRL}, there is to best of our knowledge no abstract model, that works for all environments. Therefore the model must be adapted to fit the environment at hand best.

\section{Deep Reinforcement Learning}
\label{sec:sota:rl}
The earliest work found related to Deep Reinforcement Learning is \textit{Reinforcement Learning for Robots Using Neural Networks}. This PhD thesis illustrated how an ANN could be used in RL to perform actions in an environment with delayed reward signals successfully. \cite{Lin1993}

With several breakthroughs in computer vision in early 2010's, researchers started work on integrating ConvNets into RL algorithms. Q-Learning together with Deep Learning was a game-changing moment, and has had tremendous success in many single agent environments on the Atari 2600 platform. Deep Q-Learning (\gls{DQN}) as proposed by Mnih et al. used ConvNets to predict the Q function. This architecture outperformed human expertise in over half of the games. \cite{Mnih2013}

Hasselt et al. proposed \textit{Double DQN} (\gls{DDQN}), which reduced the overestimation of action values in the Deep Q-Network. This led to improvements in some of the games on the Atari platform. \cite{VanHasselt2015}

Wang et al. then proposed a dueling architecture of DQN which introduced estimation of the value function and advantage function. These two functions were then combined to obtain the Q-Value. Dueling DQN were implemented with the previous work of van Hasselt et al. \cite{Wang2015}. 

Harm van Seijen et al.\ recently published an algorithm called Hybrid Reward Architecture (HRA) which is a divide and conquer method where several agents estimate a reward and a Q-value for each state. The algorithm performed above human expertise in Ms. Pac-Man, which is considered one of the hardest games in the Atari 2600 collection and is currently state-of-the-art in the reinforcement learning domain. The drawback of this algorithm is that generalization of Minh et al. approach is lost due to a huge number of separate agents that have domain-specific sensory input. \cite{VanSeijen2017}

There have been few attempts at using Deep Q-Learning on advanced simulators made explicitly for machine-learning. It is probable that this is because there are very few environments created for this purpose.

\section{Generative Modeling}
\label{sec:sota:gan}

There are primarily three Generative models that are actively used in recent literature, \gls{GAN}, Variational Autoencoders and Autoregressive Modeling. \gls{GAN} show far better results than any other generative model and is the primary field of research for this thesis.

\begin{figure}
	\centering
	\includegraphics[width=5cm]{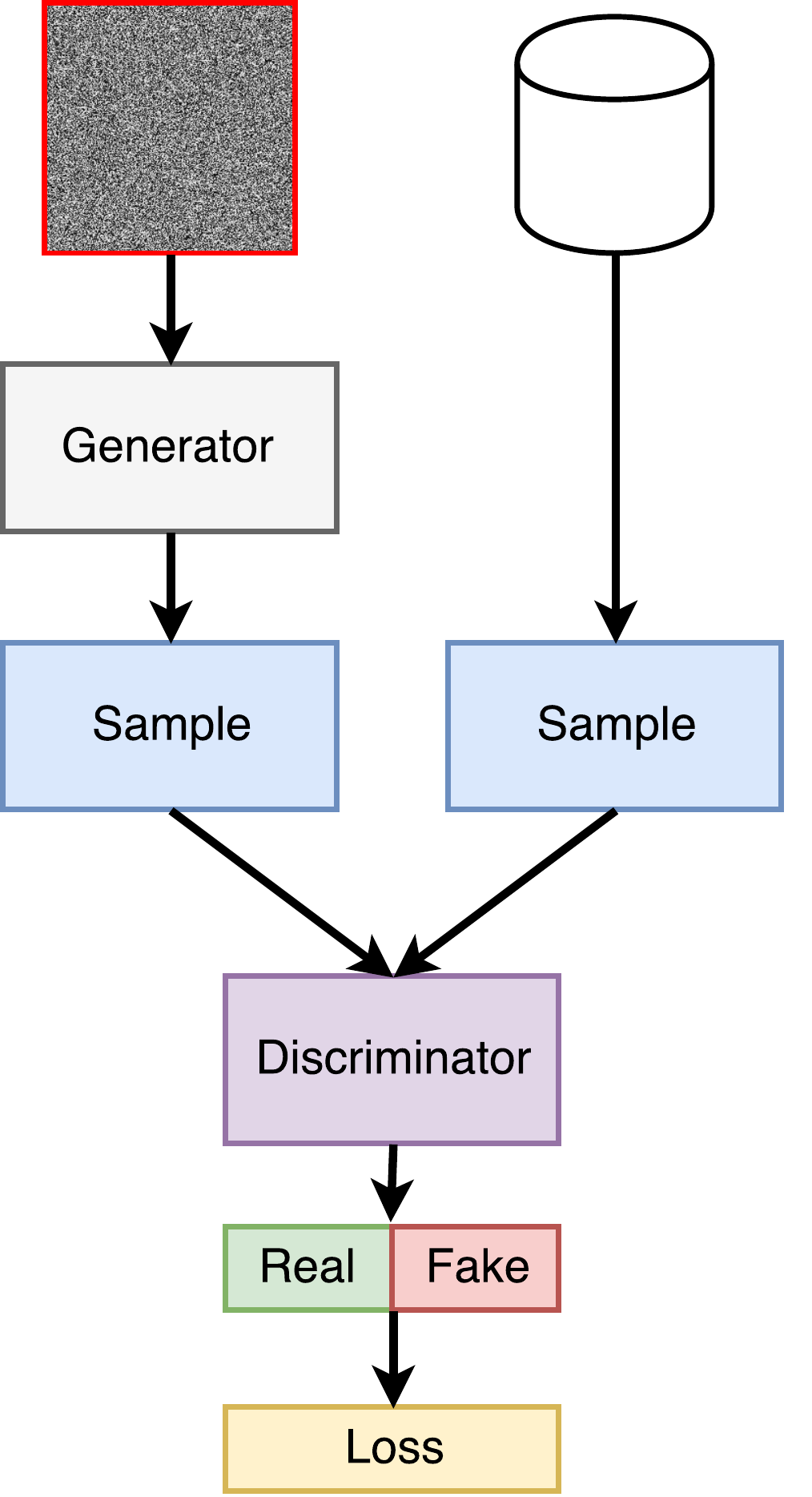}
	\caption{Illustration of Generative Adversarial Network Model}
	\label{fig:gan_architecture}
\end{figure}

\gls{GAN} show great potential when it comes to generating artificial images from real samples. The first occurrence of \gls{GAN} was introduced in the paper \textit{Generative Adversarial Networks} from Ian J. Goodfellow et al.~\cite{Goodfellow2014}. This paper proposed a framework using a generator and discriminator neural network. The general idea of the framework is a two-player game where the generator generates synthetic images from noise and tries to fool the discriminator by learning to create authentic images, see Figure~\ref{fig:gan_architecture}.

\begin{figure}
	\centering
	\includegraphics[width=5cm]{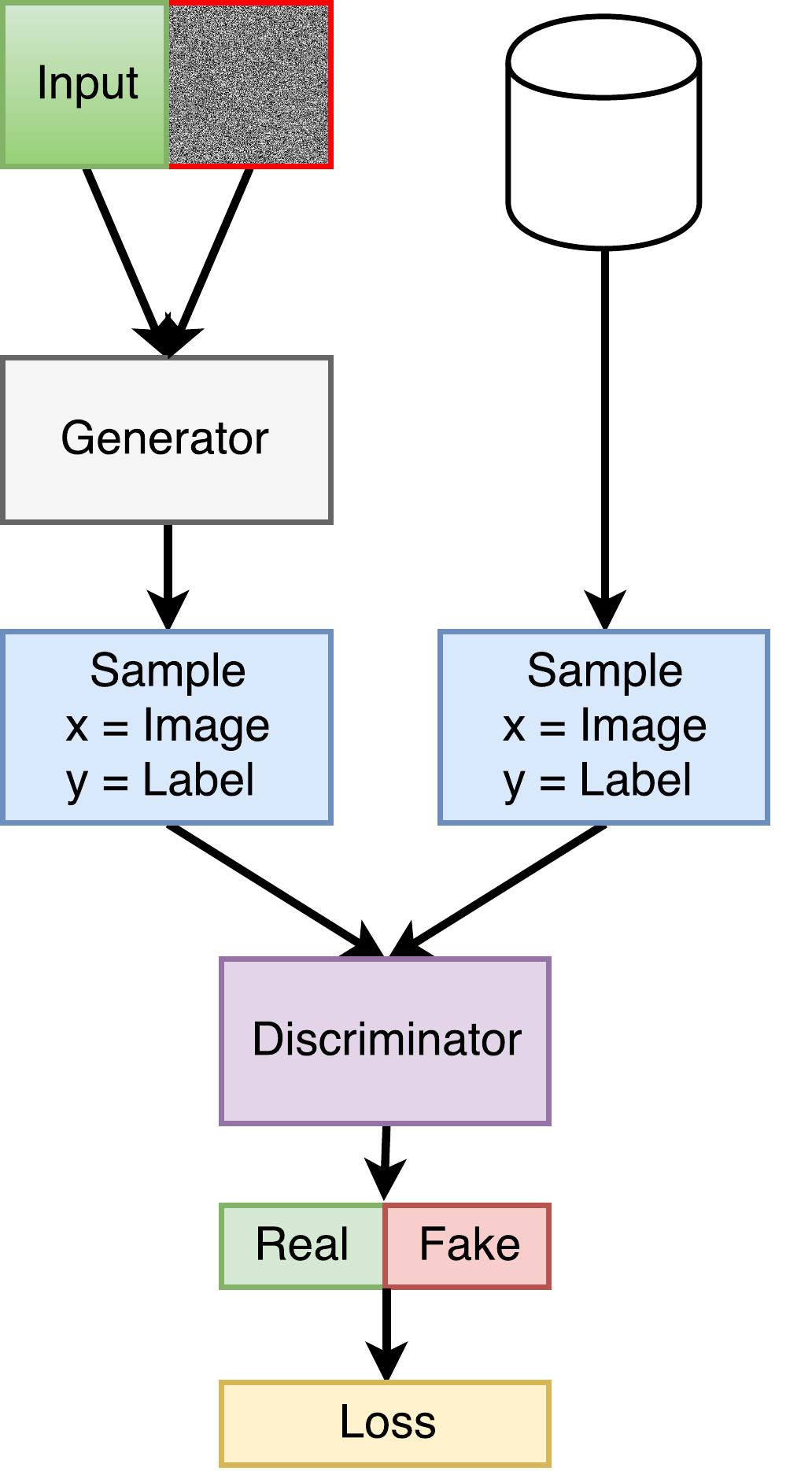}
	\caption{Illustration of Conditional Generative Adversarial Network Model}
	\label{fig:cgan_architecture}
\end{figure}

In future work, it was specified that the proposed framework could be extended from \(p(x) \rightarrow p(x~|~c)\). This was later proposed in the paper \textit{Conditional Generative Adversarial Nets} (\gls{CGAN}) by Mirza et al.\cite{Mirza2014}. \gls{GAN} is extended to a conditional model by demanding additional information \textit{y} as input for the generator and discriminator. This enabled to condition the generated images on information like labels illustrated in Figure~\ref{fig:cgan_architecture}. 

Radford et al.~\cite{Radford2015} proposed \textit{Deep Convolutional Generative Adversarial Networks} (DCGAN) in \textit{Unsupervised Representation Learning with Deep Convolutional Generative Adversarial Networks}. This paper improved on using \gls{ConvNet}s in unsupervised settings. Several architectural constraints were set to make training of DCGAN stable in most scenarios. This paper illustrated many great examples of images generated with DCGAN, for instance, state-of-the-art bedroom images.

In summer 2016, Salimans et al. (Goodfellow) presented \textit{Improved Techniques for Training GANs} achieving state-of-the-art results in the classification of MNIST, CIFAR-10, and SVHN~\cite{Salimans2016}. This paper introduced minibatch discrimination, historical averaging, one-sided label smoothing and virtual batch normalization.

There have been many advances in \gls{GAN} between and after these papers. Throughout the research process of \gls{GAN}s, the most prominent architecture for our problem is Conditional GANs which enables us to condition the input variable \textit{x} on variable \textit{y}. The most recent paper on this topic is \textit{Towards Diverse and Natural Image Descriptions via a Conditional GAN} from Dai et al.~\cite{Dai2017}. This paper focuses on captioning images using Conditional \gls{GAN}s. It produced captions that were of similar quality to human-made captions. In \gls{RL} terms it is successfully able to learn a good policy for the dataset. 

\section{Capsule Networks}
\label{sec:sota:capsnet}
Capsule Neural Networks (\gls{CapsNet}) is a novel deep learning architecture that attempts to improve the performance of image and object recognition. \gls{CapsNet} is theorized to be far better at detecting rotated objects and requires less training data than traditional \gls{ConvNet}. Instead of creating deep networks like for example ResNet-50, a Capsule layer is created, containing several sub-layers in depth. Each of these capsules has a group of neurons, where the objective is to learn a specific object or part of an object. When an image is inserted into the Capsule Layer, an iterative process of identifying objects begins. The higher dimension layers receive a signal from the lower dimensions. The higher dimension layer then determines which signal is the strongest and a connection is made between the winning signal (betting). This method is called \textit{dynamic routing}. This routing-by-agreement ensures that features are mapped to the output, and preserves all input information at the same time.

Pooling in \gls{ConvNet} is also a primitive form of routing, but information about the input is lost in the process. This makes pooling much more vulnerable to attacks compared to dynamic routing. In current state-of-the-art, \gls{CapsNet} is explained as inverse graphics, where a capsule tries to learn an activity vector describing the probability that an object exists.

Capsule Networks are still only in infancy, and there is not well-documented research on this topic yet apart from state-of-the-art paper \textit{ Dynamic Routing Between Capsules} by Sabour et al. \cite{Sabour2017}.

\section{Game Learning Platforms}
\label{sec:sota:gameenv}

There exists several exciting game learning platform used to research state-of-the-art \gls{AI} algorithms. The goal of these platforms is generally to provide the necessary platform for studying \textit{Artificial General Intelligence} (\gls{AGI}). \gls{AGI} is a term used for \gls{AI} algorithms that can perform well across several environments without training. \gls{DRL} is currently the most promising branch of algorithms to solve \gls{AGI}.

Bellemare et al. provided in 2012 a learning platform \textit{Arcade Learning Environment} (ALE) that enabled scientists to conduct edge research in general deep learning \cite{Bellemare2012}. The package provided hundreds of Atari 2600 environments that in 2013 allowed Minh et al. to do a breakthrough with Deep Q-Learning and A3C. The platform has been a critical component in several advances in RL research. \cite{Mnih2013, Mnih2015, Mnih2016}

The Malmo project is a platform built atop of the popular game \textit{Minecraft}. This game is set in a 3D environment where the object is to survive in a world of dangers. The paper \textit{The Malmo Platform for Artificial Intelligence Experimentation} by Johnson et al. claims that the platform had all characteristics qualifying it to be a platform for \gls{AGI} research.\cite{Johnson2016}

ViZDoom is a platform for research in Visual Reinforcement Learning. With the paper \textit{ViZDoom: A Doom-based AI Research Platform for Visual Reinforcement Learning} Kempka et al. illustrated that an \gls{RL} agent could successfully learn to play the game \textit{Doom}, a first-person shooter game, with behavior similar to humans.~\cite{Kempka2016}

With the paper \textit{DeepMind Lab}, Beattie et al. released a platform for 3D navigation and puzzle solving tasks. The primary purpose of Deepmind Lab is to act as a platform for \gls{DRL} research.\cite{Beattie2016}

In 2016, Brockman et al. from OpenAI released GYM which they referred to as \textit{"a toolkit for developing and comparing reinforcement learning algorithms"}. GYM provides various types of environments from following technologies: Algorithmic tasks, Atari 2600, Board games, Box2d physics engine, MuJoCo physics engine, and Text-based environments.
OpenAI also hosts a website where researchers can submit their performance for comparison between algorithms. GYM is open-source and encourages researchers to add support for their environments. \cite{Brockman2016}

OpenAI recently released a new learning platform called \textit{Universe}. This environment further adds support for environments running inside VNC. It also supports running Flash games and browser applications. However, despite OpenAI's open-source policy, they do not allow researchers to add new environments to the repository. This limits the possibilities of running any environment. Universe is, however, a significant learning platform as it also has support for desktop games like Grand Theft Auto IV, which allow for research in autonomous driving \cite{Li2017}.

Very recently Extensive Lightweight Flexible (ELF) research platform was released with the NIPS paper \textit{ELF: An Extensive, Lightweight and Flexible Research Platform for Real-time Strategy Games}. This paper focuses on \gls{RTS} game research and is the first platform officially targeting these types of games. \cite{Tian2017}

\subsection{Summary}

Multiple interesting observations about current state-of-the-art in learning platforms for \gls{RL} algorithms were found during our research. Table~\ref{tbl:sota_env_summary} describes the capabilities of each of the learning platform in the interest of fulfilling the requirements of this thesis. GYM-CAIR is included in this comparison and is further described in Chapter~\ref{chap:env} and \ref{chap:solutions}. 

\begin{table}[]
	\centering
	\begin{tabular}{|ll
			>{\columncolor[HTML]{D5E8D4}}l 
			>{\columncolor[HTML]{D5E8D4}}l |}
		\hline
		\cellcolor[HTML]{BBDAFF}\textbf{Platform} & \cellcolor[HTML]{BBDAFF}\textbf{Diversity} & \cellcolor[HTML]{BBDAFF}\textbf{AGI} & \cellcolor[HTML]{BBDAFF}\textbf{Advanced Environment(s)} \\ \hline
		\multicolumn{1}{|l|}{ALE} & \cellcolor[HTML]{D5E8D4}Yes & Yes & \cellcolor[HTML]{F8CECC}No \\
		\multicolumn{1}{|l|}{Malmo Platform} & \cellcolor[HTML]{F8CECC}No & \cellcolor[HTML]{F8CECC}No & Yes \\
		\multicolumn{1}{|l|}{ViZDoom} & \cellcolor[HTML]{F8CECC}No & Yes & Yes \\
		\multicolumn{1}{|l|}{DeepMind Lab} & \cellcolor[HTML]{F8CECC}No & \cellcolor[HTML]{F8CECC}No & Yes \\
		\multicolumn{1}{|l|}{OpenAI Gym} & \cellcolor[HTML]{D5E8D4}Yes & Yes & \cellcolor[HTML]{F8CECC}No \\
		\multicolumn{1}{|l|}{OpenAI Universe} & \cellcolor[HTML]{D5E8D4}Yes & Yes & \cellcolor[HTML]{F8CECC}Partially \\
		\multicolumn{1}{|l|}{ELF} & \cellcolor[HTML]{F8CECC}No & \cellcolor[HTML]{F8CECC}No & Yes \\
		\multicolumn{1}{|l|}{(GYM-CAIR)} & \cellcolor[HTML]{D5E8D4}Yes & Yes & Yes \\ \hline
	\end{tabular}
	\caption{Summary of researched platforms}
	\label{tbl:sota_env_summary}
\end{table}

\section{Reinforcement Learning in Games}
\label{sec:sota:games}
Reinforcement Learning for games is a well-established field of research and is frequently used to measure how well an algorithm can perform within an environment. This section presents some of the most important achievements in Reinforcement Learning.

TD-Gammon is an algorithm capable of reaching an expert level of play in the board game \textit{Backgammon}~\cite{Tesauro1994, Tesauro1995}. The algorithm was developed by Gerald Tesauro in 1992 at IBM's Thomas J. Watson Research Center. TD-Gammon consists of a three-layer \gls{ANN} and is trained using an \gls{RL} technique called \textit{TD-Lambda}. TD-Lambda is a temporal difference learning algorithm invented by Richard S. Sutton~\cite{Sutton1990}. The \gls{ANN} iterates over all possible moves the player can perform and estimates the reward for that particular move. The action that yields the highest reward is then selected. TD-Gammon is one of the first algorithms to utilize self-play methods to improve the \gls{ANN} parameters. 

In late 2015, \textit{AlphaGO} became the first algorithm to win against a human professional Go player. AlphaGO is an \gls{RL} framework that uses Monte Carlo Tree search and two Deep Neural Networks for value and policy estimation~\cite{Silver2016}. Value refers to the expected future reward from a state assuming that the agent plays perfectly. The policy network attempts to learn which action is best in any given board configuration. The earliest versions of AlphaGO used training data from games played by human professionals. In the most recent version, \textit{AlphaGO Zero}, only self-play is used to train the AI~\cite{Silver2017a} In a recent update, AlphaGO was generalized to work for Chess and Shogi (Japanese Chess) only using 24 hours to reach superhuman level of play~\cite{Silver2017}

DOTA 2 is an advanced player versus player game where the player is controlling a hero unit. The game objective is to defeat the enemy heroes and destroy their base. In August 2017, OpenAI invented an \gls{RL} based AI that defeated professional players in one versus one game. Training was done only using self-play, and the algorithm learned how to exploit game mechanics to perform well.

DeepStack is an algorithm that can perform an expert level play in Texas Hold'em poker. This algorithm uses tree-search in conjunction with neural networks to perform sensible actions in the game \cite{Moravcik2017}. DeepStack is a general-purpose algorithm that aims to solve problems with imperfect information. 

There have been several other significant achievements in \gls{AI}, but these are not directly related to the use of \gls{RL} algorithms. These include Deep Blue\footnote{Deep Blue is not  \href{https://www.research.ibm.com/deepblue/meet/html/d.3.3a.html}{AI}} and Watson from IBM.

\part{Contributions}

\chapter{Environments}
\label{chap:env}
Simulated environments are a popular research method to conduct experiments on algorithms in computer science. These simulated environments are often tailored to the problem, and quickly proves, or disproves the capability of an algorithm. This chapter proposes four new game environments for deep learning research: \textit{FlashRL}, \textit{Deep Line Wars}, \textit{Deep RTS}, and \textit{Deep Maze}. The game \textit{Flappy Bird} is introduced as a validation environment for experiments conducted in Chapter~\ref{chap:results:dqn}.
\begin{figure}[!t]\centering
	\includegraphics[width=0.8\textwidth]{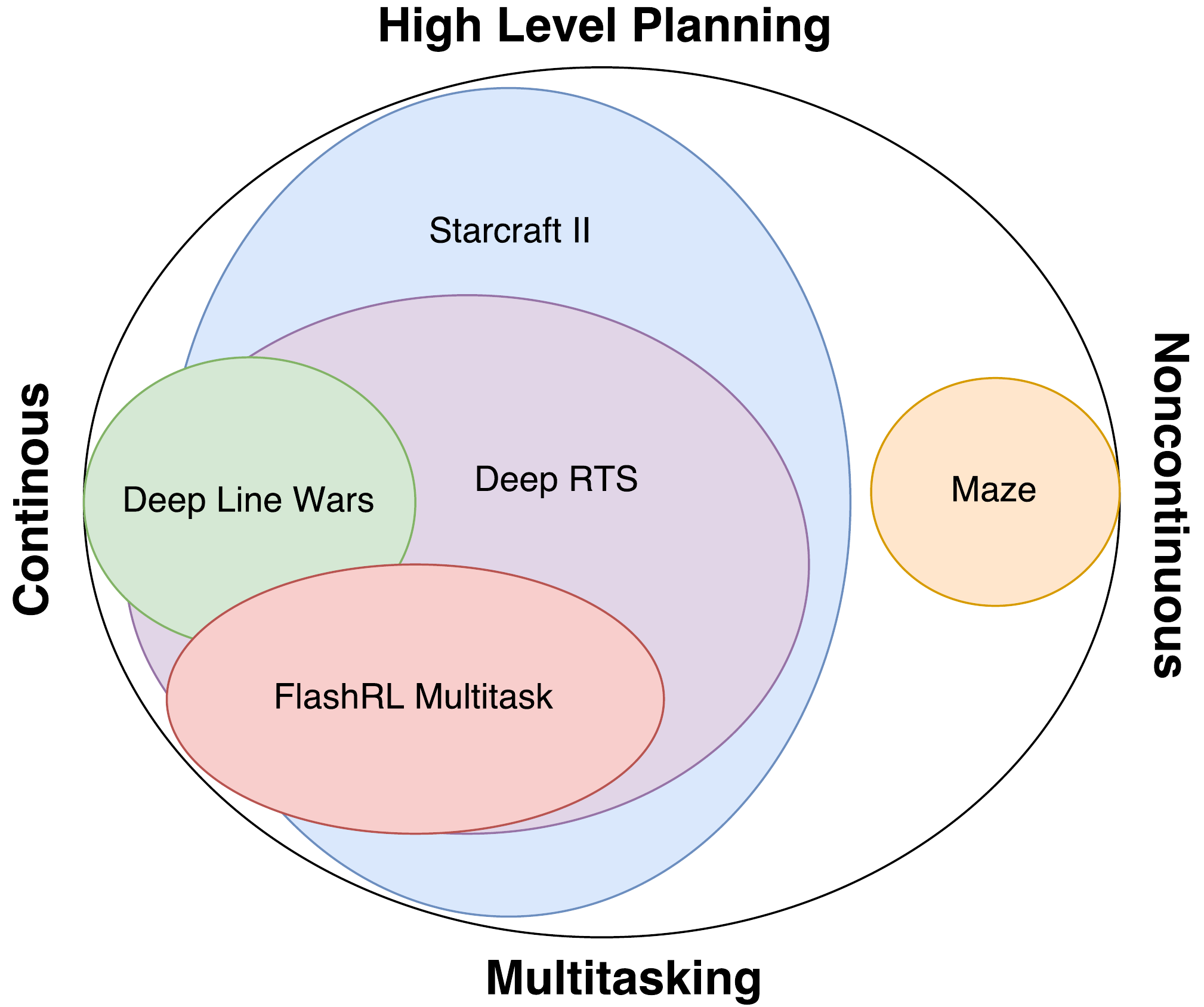}
	\caption{Environment field of focus}
	\label{fig:field_of_focus}
\end{figure}
Figure~\ref{fig:field_of_focus} illustrates that each of these environments has different goals, and the agent placed in these environments are challenged in several topics, for instance, multitasking, deep and shallow state interpretation and planning. This chapter creates a foundation for research into \gls{CapsNet} based \gls{RL}-algorithms in advanced game environments. 

\section{FlashRL}
\label{sec:env:flashrl}
\textit{Adobe Flash} is a multimedia software platform used for the production of applications and animations. The Flash run-time was recently declared deprecated by Adobe, and by 2020, no longer supported. Flash is still frequently used in web applications, and there are countless games created for this platform. Several web browsers have removed the support for  the Flash runtime, making it difficult to access the mentioned game environments. Flash games are an excellent resource for machine learning benchmarking, due to size and diversity of its game repository. It is therefore essential to preserve the Flash run-time as a platform for \gls{RL}.

\textit{Flash Reinforcement Learning} (FlashRL) is a novel platform that acts as an input/output interface between Flash games and \gls{DRL} algorithms. FlashRL enables researchers to interface against almost any Flash-based game environment efficiently.

\begin{figure}[htp]
	\centering
	\includegraphics[width=0.7\textwidth]{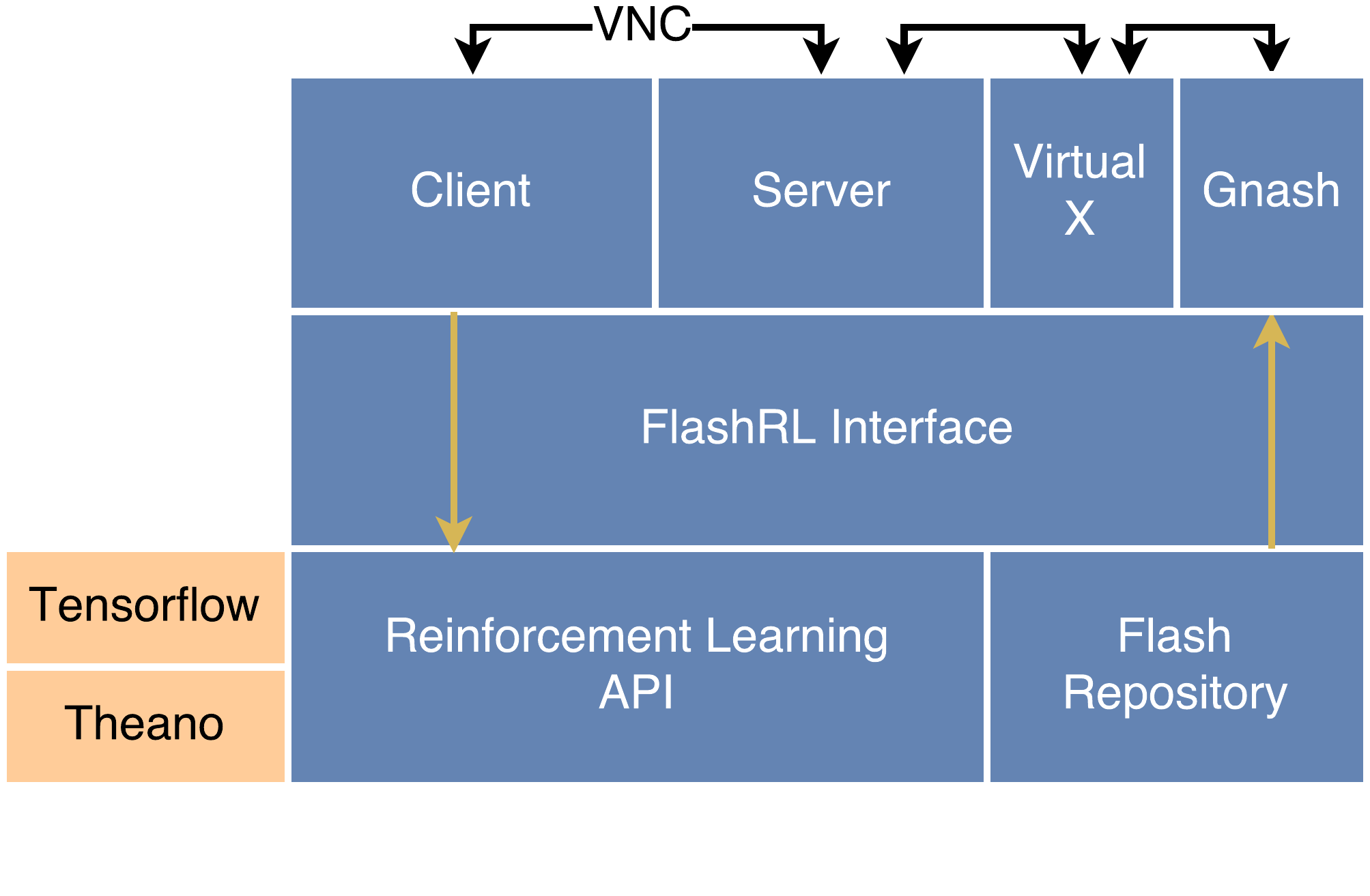}
	\caption{FlashRL: Architecture}
	\label{fig:flashrl_architecture}
\end{figure}
The learning platform is developed primarily for Linux based operating systems but is likely to run on Cygwin with few modifications. There are several key components that FlashRL uses to operate adequate, see Figure \ref{fig:flashrl_architecture}. FlashRL uses XVFB to create a virtual frame-buffer. The frame-buffer acts like a regular desktop environment, found in Linux desktop distributions~\cite{Hunt2004}. Inside the frame-buffer, a Flash game chosen by the researcher is executed by a third-party flash player, for example, \textit{Gnash}. A VNC server serves the frame-buffer and enable FlashRL to access display, mouse and keyboard via the VNC protocol.
The VNC Client \textit{pyVLC} was specially made for this FlashRL. The code base originates from python-vnc-viewer~\cite{Techtonik2015}. The last component of FlashRL is the Reinforcement Learning API that allows the developer to access the input/output of the pyVLC. This makes it easy to develop sequenced algorithms by using API callbacks or invoke commands manually with threading.

\begin{figure}[htp]
	\centering
	\includegraphics[width=0.8\textwidth]{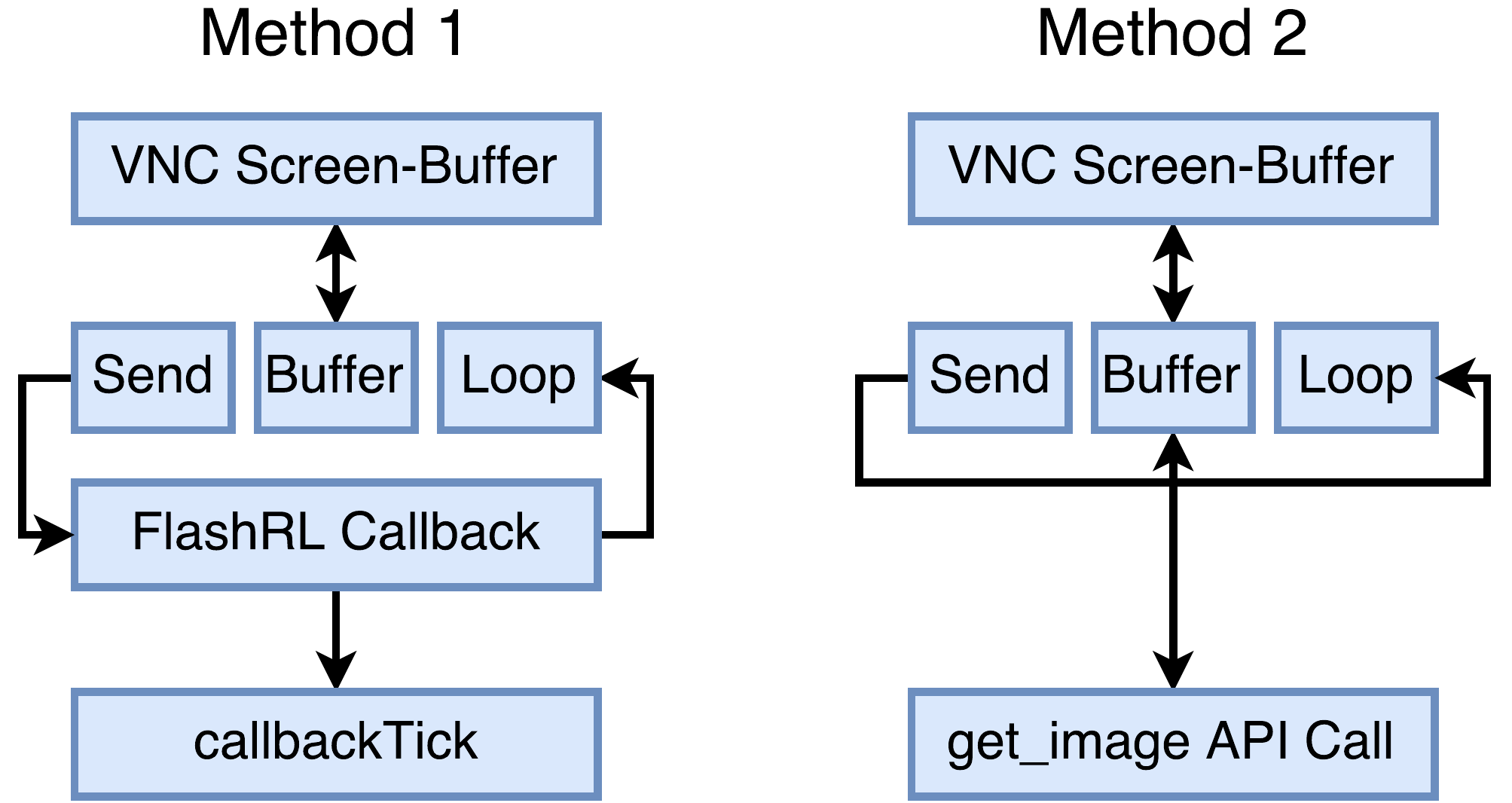}
	\caption{FlashRL: Frame-buffer Access Methods}
	\label{fig:execution_1}
\end{figure}

Figure~\ref{fig:execution_1} illustrates two methods of accessing the frame-buffer from the Flash environment. Both approaches are sufficient to perform \gls{RL}, but each has its strengths and weaknesses.
Method 1 sends frames at a fixed rate, for example at 60 frames per second. The second method does not set  any restrictions of how fast the frame-buffer can be captured. This is preferable for developers that do not require images from fixed time-steps because it demands less processing power per frame. The framework was developed with deep learning in mind and is proven to work well with Keras and Tensorflow \cite{Andersen2017a}.

\begin{figure}[!ht]
	\centering
	\includegraphics[width=0.8\textwidth]{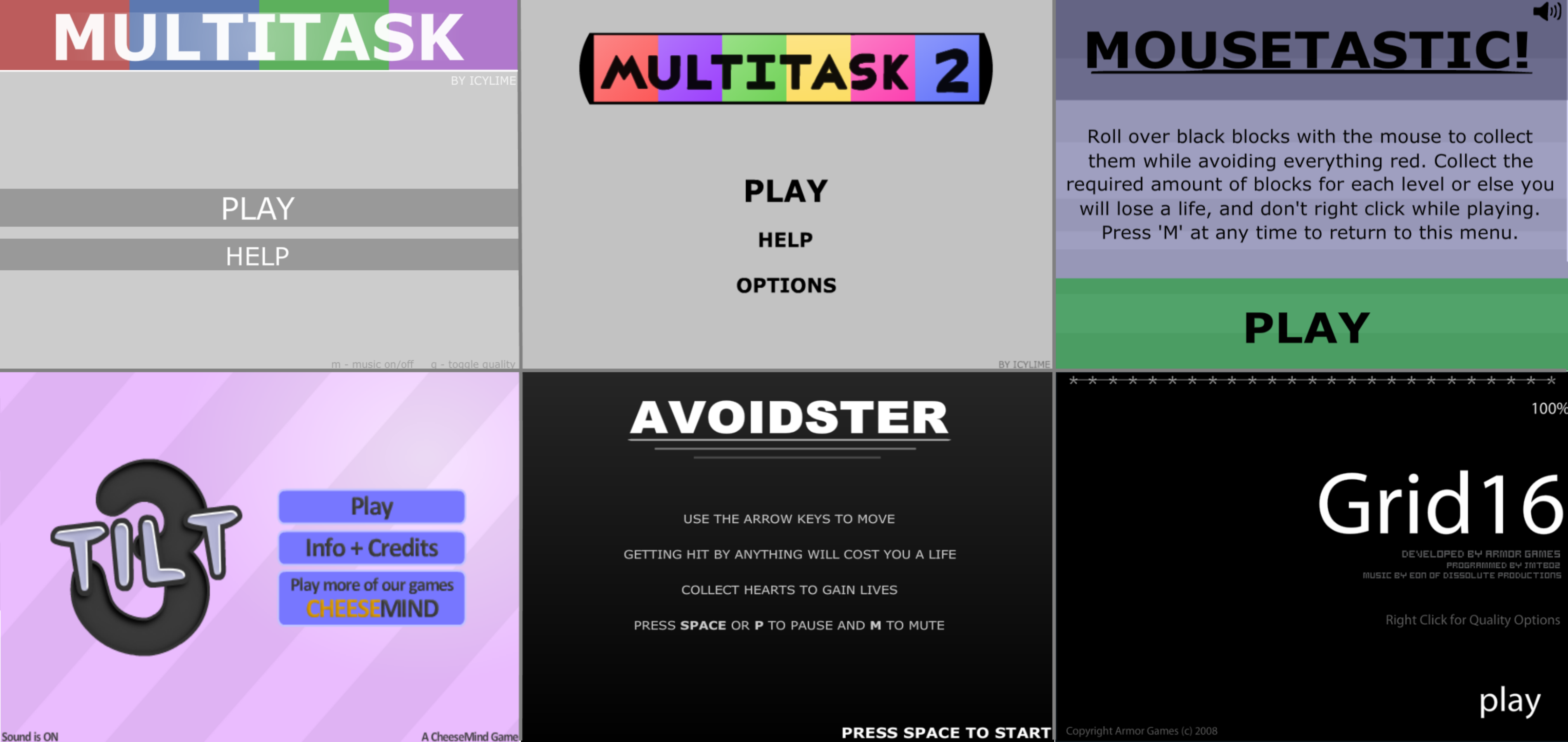}
	\caption{FlashRL: Available environments}
	\label{fig:game_list}
\end{figure}

There are close to a thousand game environments available for the first version of FlashRL. These game environments were gathered from different sources on the world wide web. FlashRL has a relatively small code-base and to preserve this size, the Flash repository is hosted at a remote site. Because of the large repository, not all games have been tested thoroughly. The game quality may therefore vary. Figure~\ref{fig:game_list} illustrates tested games that yield a great value for \gls{DRL} research.

\section{Deep Line Wars}
\label{sec:env:deeplinewars}

\begin{figure}
	\centering
	\includegraphics[width=0.90\linewidth]{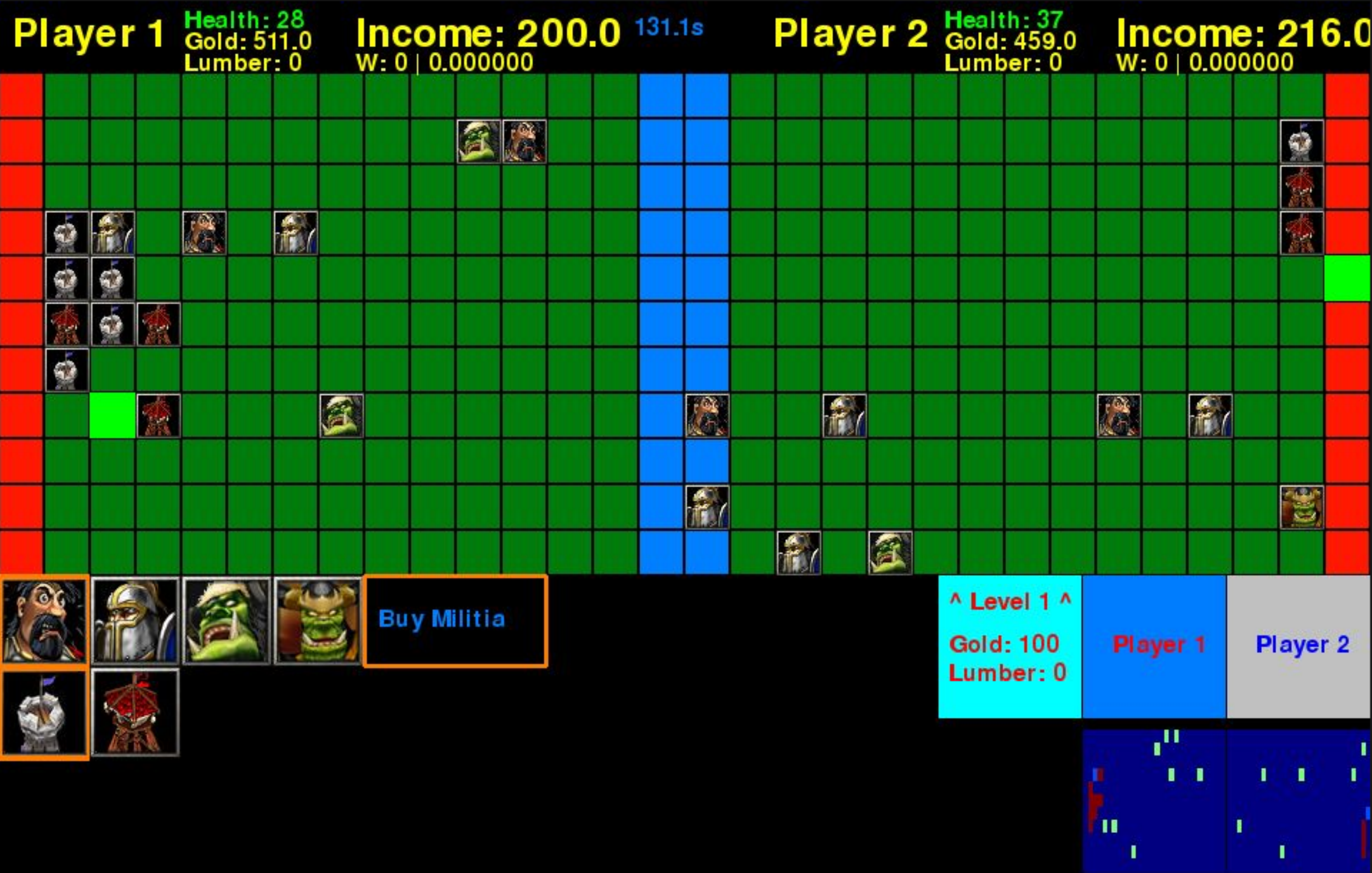}
	\caption{Deep Line Wars: Graphical User Interface}
	\label{fig:dlw}
\end{figure}

The game objective of Deep Line Wars is to invade the opposing player (hereby enemy) with mercenary units until all health points are depleted (see Figure \ref{fig:dlw}). For every friendly unit that enters the red area on the map, the enemy health pool is reduced by one. When a player purchases a mercenary unit, it spawns at a random location inside the red area of the owners base. Mercenary units automatically move towards the enemy base. To protect the base, players can construct towers that shoot projectiles at the opponents mercenaries. When a mercenary dies, a fair percentage of its gold value is awarded to the opponent. When a player sends a unit, the income is increased by a percentage of the units gold value. As a part of the income system, players gain gold at fixed intervals.\cite{Andersen2017}

To successfully master game mechanics of Deep Line Wars, the player (agent) must learn
\begin{itemize}
	\item offensive strategies of spawning units,
	\item defending against the opposing player's invasions, and
	\item maintain a healthy balance between offensive and defensive to maximize income
\end{itemize}
The game is designed so that if the player performs better than the opponent in these mechanics, he is guaranteed to win over the opponent.

\begin{figure}
	\centering
	\includegraphics[width=0.90\linewidth]{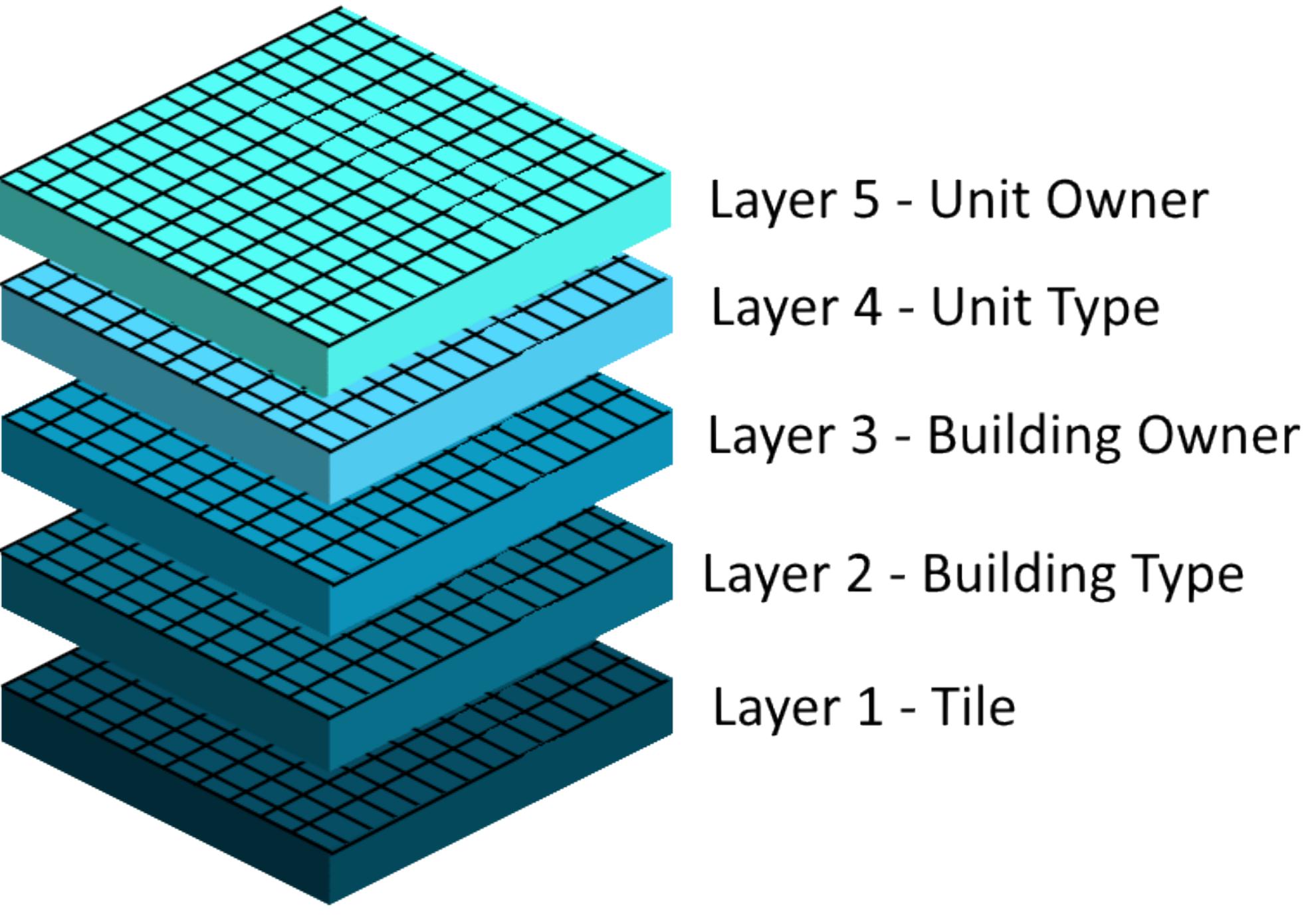}
	\caption{Deep Line Wars: Game-state representation}
	\label{fig:dlw_game_arch}
\end{figure}

\begin{figure}
	\centering
	\begin{minipage}{.5\textwidth}
		\centering
		\includegraphics[height=2.2cm]{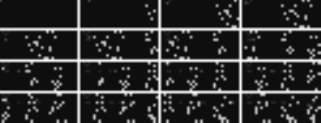}
	\end{minipage}%
	\begin{minipage}{.5\textwidth}
		\centering
		\includegraphics[height=2.2cm]{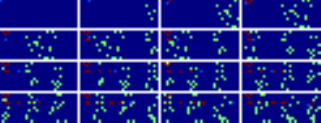}
	\end{minipage}
	\caption{Deep Line Wars: Game-state representation using heatmaps}
	\label{fig:dlw_heatmaps}
\end{figure}

Because the game is specifically targeted towards \gls{RL} research, the game-state is defined as a multi-dimensional matrix. This way, it is trivial to input the game-state directly into \gls{ANN} models. Figure \ref{fig:dlw_game_arch} illustrates how the game state is constructed. This state is later translated into graphics, seen in Figure \ref{fig:dlw}. It is beneficial to directly access this information because it requires less data preprocessing compared to using raw game images. Deep Line Wars also features abstract state representation using heat-maps, seen in Figure~\ref{fig:dlw_heatmaps}. By using heatmaps, the state-space is reduced by a magnitude, compared to raw images. Heatmaps can better represent the true objective of the game, enabling faster learning for \gls{RL} algorithms~\cite{Samek2015}.

\begin{table}[]
	\centering
	\begin{tabular}{|lll|}
		\hline
		\rowcolor[HTML]{BBDAFF} 
		\multicolumn{1}{|l|}{\cellcolor[HTML]{BBDAFF}Representation} & \multicolumn{1}{l|}{\cellcolor[HTML]{BBDAFF}Matrix Size} & Data Size \\ \hline
		Image                                                        & \(800\cdot600\cdot3\)                                    & 1440000   \\ \hline
		Matrix                                                       & \(10\cdot15\cdot5\)                                      & 750       \\ \hline
		Heatmap RGB                                                  & \(10\cdot15\cdot3\)                                      & 450       \\ \hline
		Heatmap Grayscale                                            & \(10\cdot15\cdot1\)                                      & 150       \\ \hline
	\end{tabular}
	\caption{Deep Line Wars: Representation modes}
	\label{tbl:dlw_representation_modes}
\end{table}
In Deep Line Wars, there are primarily four representation modes available for \gls{RL}.Table~\ref{tbl:dlw_representation_modes} shows that there is considerably lower data size for grayscale heatmaps. Effectively, the state-space can be reduced by 9600\%, when no data preprocessing is done. Heatmaps seen in~\ref{fig:dlw_heatmaps} define
\begin{itemize}
	\item red pixels as friendly buildings, 
	\item green pixels as enemy units, and
	\item teal pixels as the mouse cursor.
\end{itemize}
When using grayscale heatmaps, RGB values are squashed into a one-dimensional matrix with values ranging between 0 and 1. Economy drastically increases the complexity of Deep Line Wars, and it is challenging to present only using images correctly. Therefore a secondary data structure is available featuring health, gold, lumber, and income. This data can then be feed into a hybrid \gls{DL} model as an auxiliary input~\cite{Wan2017}.

\section{Deep RTS}
\label{sec:env:deeprts}
\gls{RTS} games are considered to be the most challenging games for \gls{AI} algorithms to master~\cite{Vinyals2017}. With colossal state and action-spaces, in a continuous setting, it is nearly impossible to estimate the computational complexity of games such as Starcraft II. 

The game objective of Deep RTS is to build a base consisting of a Town-Hall and then expand the base to gain the military power to defeat the opponents. Each of the players starts with a worker. Workers can construct buildings and gather resources to gain an economic advantage.

\begin{figure}[!ht]
	\centering
	\includegraphics[width=\linewidth]{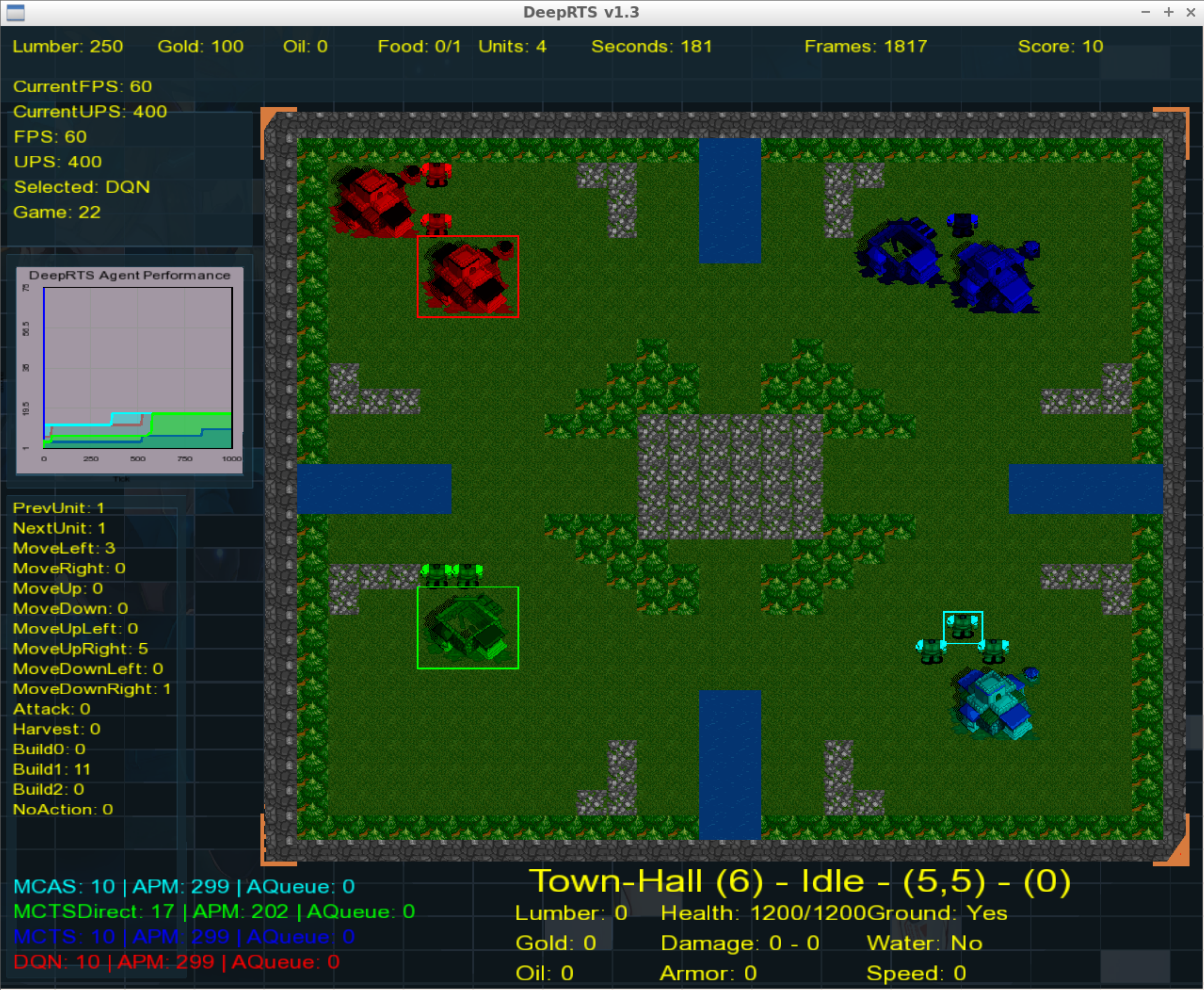}
	\caption{Deep RTS: Graphical User Interface}
	\label{fig:deeprts_game}
\end{figure}

The game mechanics consist of two main terminologies, \textit{Micro} and \textit{Macro} management. The player with the best ability to manage their resources, military, and defensive is likely to win the game. There is a considerable leap from mastering Deep Line Wars to Deep RTS, much because Deep RTS features more than two players.

\begin{table}[]
	\centering
	\begin{tabular}{|l|l|l|l|l|l|l}
		\hline
		\multicolumn{6}{|c|}{{\cellcolor[HTML]{BBDAFF}\textbf{Player Resources}}} \\
		\hline
		\textbf{Property:} & \textbf{Lumber} & \textbf{Gold} & \textbf{Oil} & \textbf{Food} & \textbf{Units} \\
		\hline
		\textbf{Value Range:} & 0 - $10^6$ & 0 - $10^6$ & 0 - $10^6$ & 0 - 200 & 0 - 200 \\
		\hline
	\end{tabular}
	\caption{Deep RTS: Player Resources}
	\label{tbl:deeprts_resources}
\end{table}

The game interface displays relevant statistics meanwhile a game session is running. These statistics show the \textit{action distribution}, \textit{player resources}, \textit{player scoreboard} and a \textit{live performance graph}.
The action distribution keeps track of which actions a player has performed in the game session. These statistics are stored to the hard-drive after a game has reached the terminal state. Player Resources (Table~\ref{tbl:deeprts_resources}), are shown at the top bar of the game. Player Scoreboard indicates the overall performance of each of the players, measured by kills, defensive points, offensive points and resource count. Deep RTS features several hotkeys for moderating the game-settings like game-speed and state representation. The hotkey menu is accessed by pressing the G-hotkey.

\begin{figure}
	\centering
	\includegraphics[width=10.5cm]{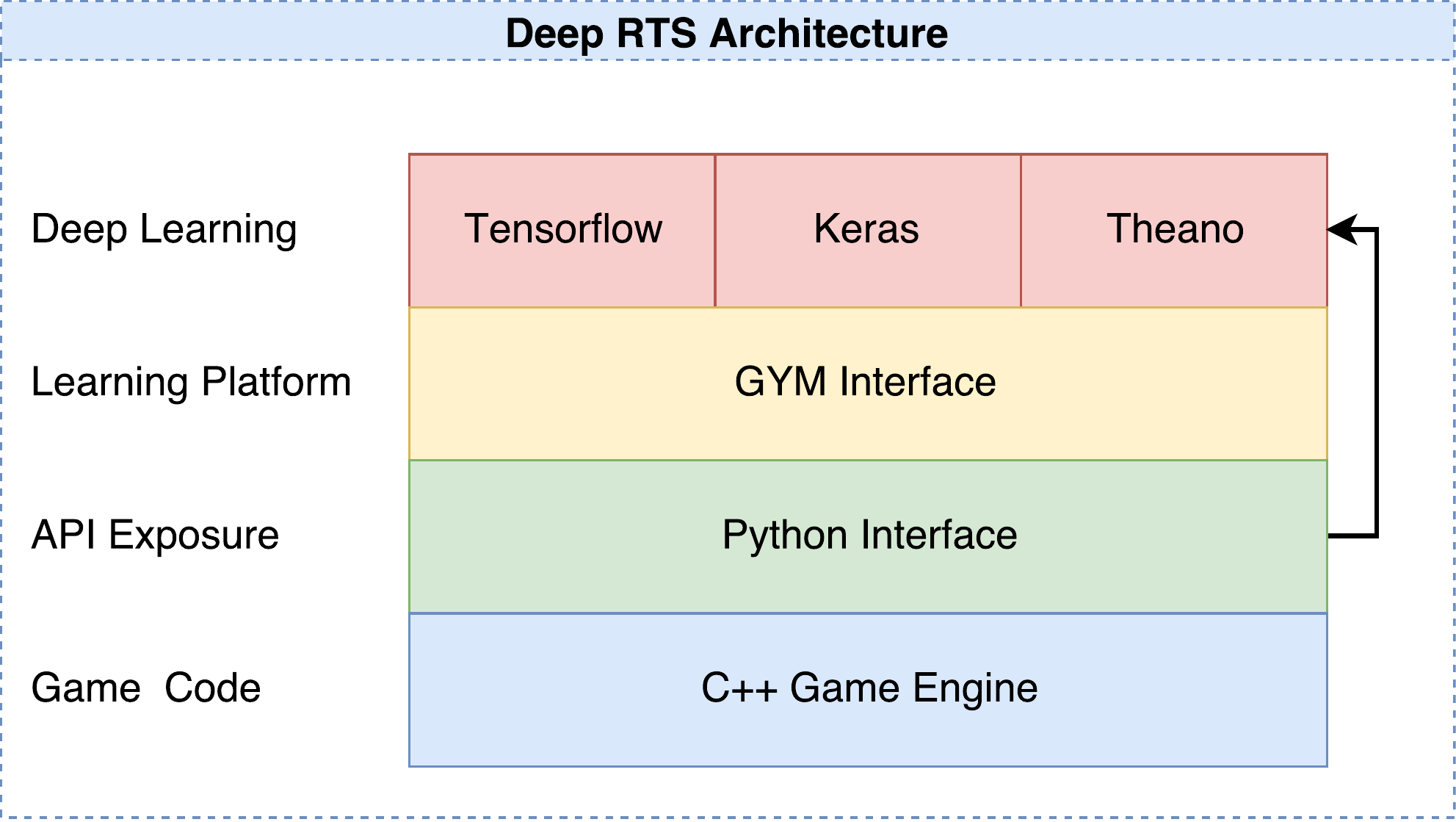}
	\caption{Deep RTS: Architecture}
	\label{fig:deeprts_arch}
\end{figure}

Deep RTS is an environment developed as an intermediate step between Atari 2600 and the famous game Starcraft II. 
It is designed to measure the performance in \gls{RL} algorithms, while also preserving the game goal. Deep RTS is developed for high-performance simulation of \gls{RTS} scenarios. The game engine is developed in C++ for performance but has an API wrapper for Python, seen in Figure~\ref{fig:deeprts_arch}. It has a flexible configuration to enable different AI approaches, for instance online and offline \gls{RL}. Deep RTS can represent the state as raw game images (C++) and as a matrix, which is compatible with both C++ and Python.

\section{Deep Maze}
\label{sec:env:maze}
Deep Maze is a game environment designed to challenge \gls{RL} agents in the \textit{shortest path problem}. Deep Maze defines the problem as follows:
\begin{itemize}
	\item How can the agent optimally navigate through any fully-observable maze?
\end{itemize}
The environment is simple, but becomes drastically more complicated when the objective is to find the optimal policy \(\pi^{\star}(s)\) \textit{where} s = state for all the maze configurations.

\begin{figure}
	\centering
	\includegraphics[width=\linewidth]{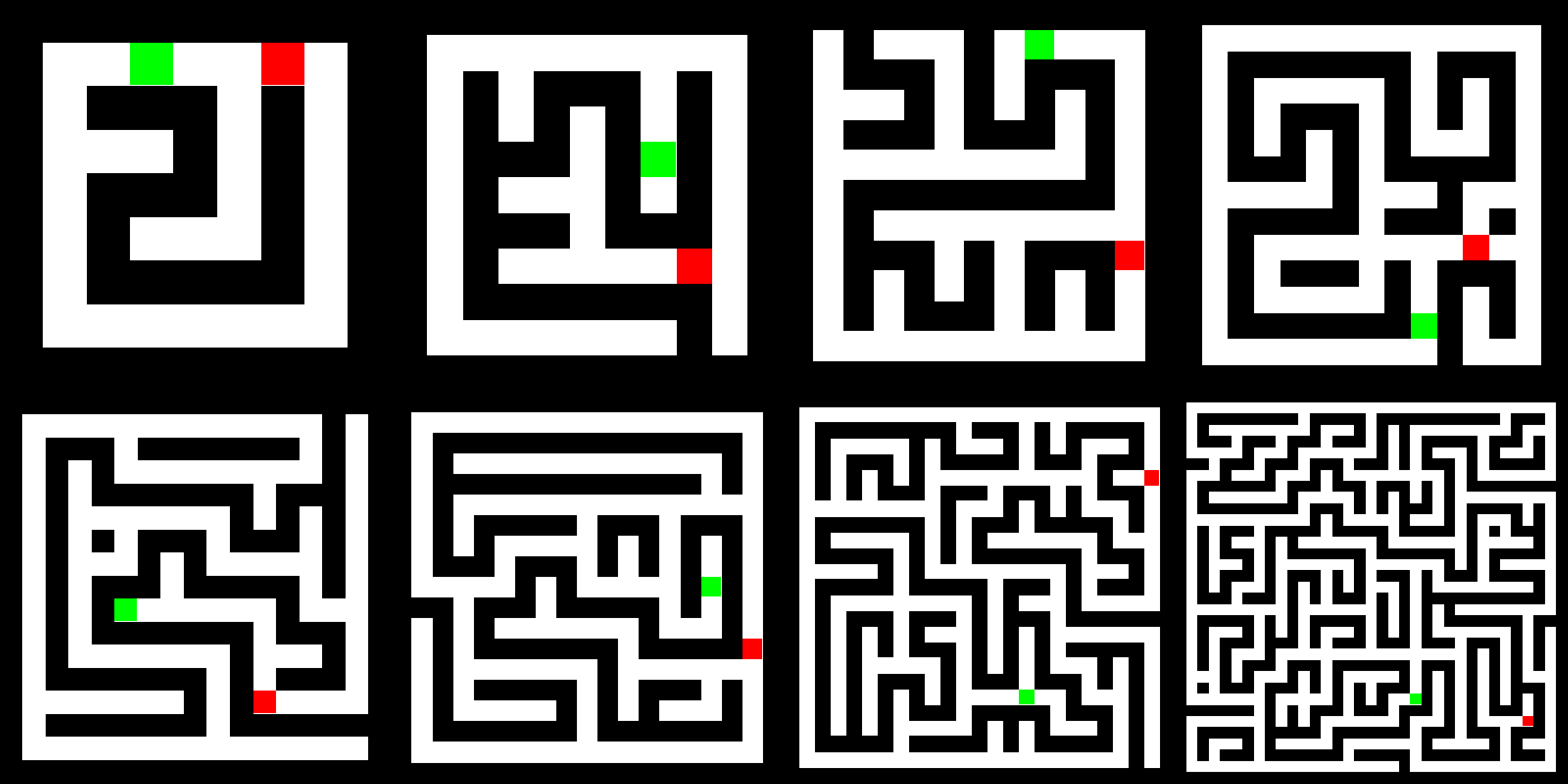}
	\caption{Deep Maze: Graphical User Interface}
	\label{fig:deep_maze}
\end{figure}

There are multiple difficulty levels for Deep Maze in two separate modes; deterministic and stochastic. In the deterministic mode, the maze structure is never changed from game to game. Stochastic mode randomizes the maze structure for every game played. There are multiple size configurations, ranging from \(7\times7~\text{to}~55\times55\) in width and height, seen in Figure~\ref{fig:deep_maze}.

\begin{figure}
	\centering
	\includegraphics[width=\linewidth]{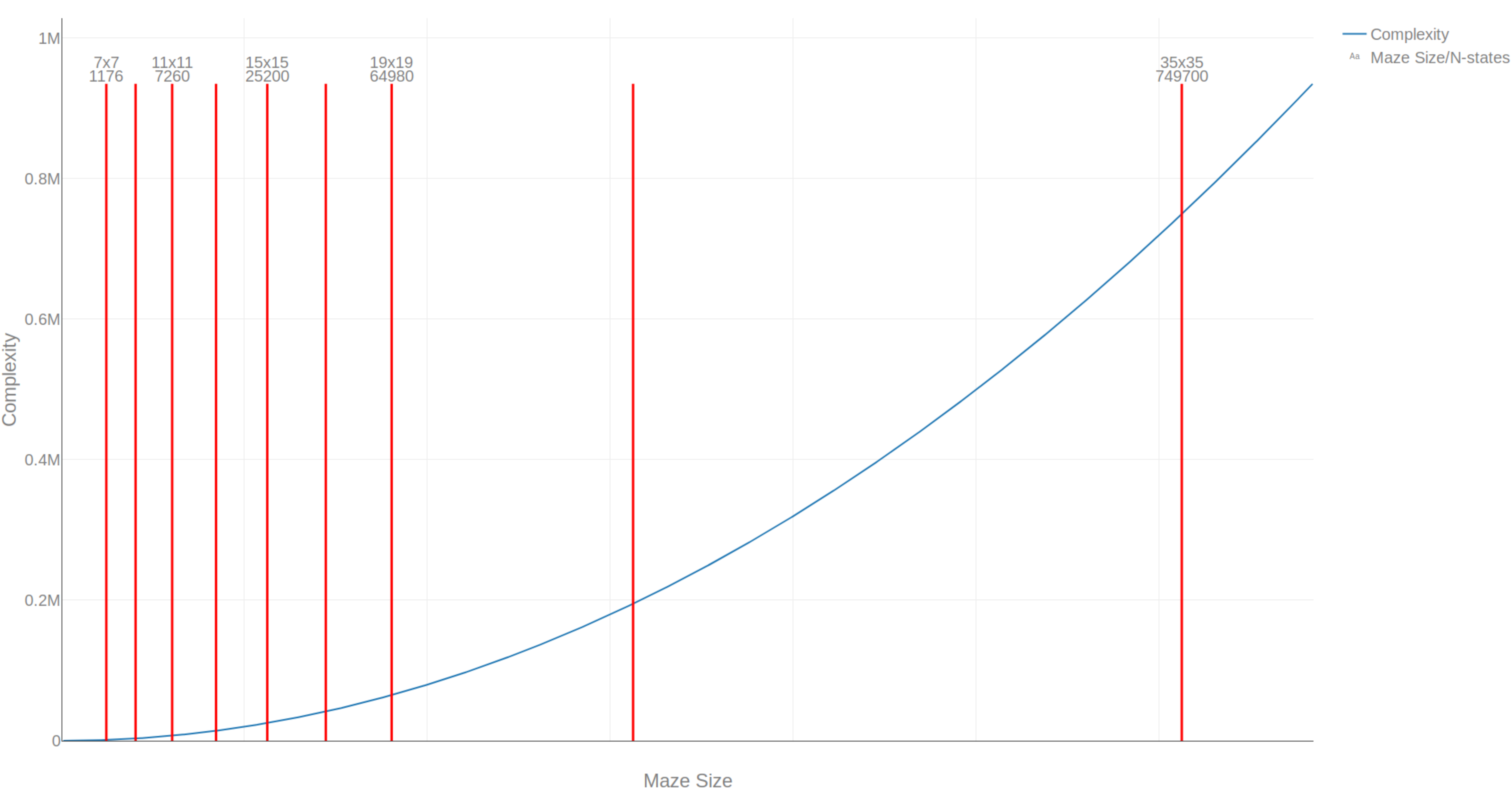}
	\caption{Deep Maze: State-space complexity}
	\label{fig:deep_maze:state_space}
\end{figure}

Figure~\ref{fig:deep_maze:state_space} illustrates how the theoretical maximum state-space set \(S\) of Deep Maze increase with maze size. This is calculated by performing following binomial: \(S = \binom{width\times height}{player + goal} = \binom{w\times h}{2}\). This is however reduced depending on the maze composition, where dense maze structures are generally less complex to solve theoretically.

The simulation is designed for performance so that each discrete time step is calculated with fewest possible CPU cycles. The simulation is estimated to run at 3 000 000 ticks per second with modern hardware. This allows for fast training of \gls{RL} algorithms.

From an \gls{RL} point of view, Deep Maze challenges an agent in state-interpretation and navigation, where the goal is to reach the terminal state in fewest possible time steps. It's a flexible environment that enables research in a single environment setting, as well as multiple scenarios played in sequence.

\section{Flappy Bird}
\label{sec:env:flappybird}

\begin{figure}[!t]\centering
	\includegraphics[width=0.8\textwidth]{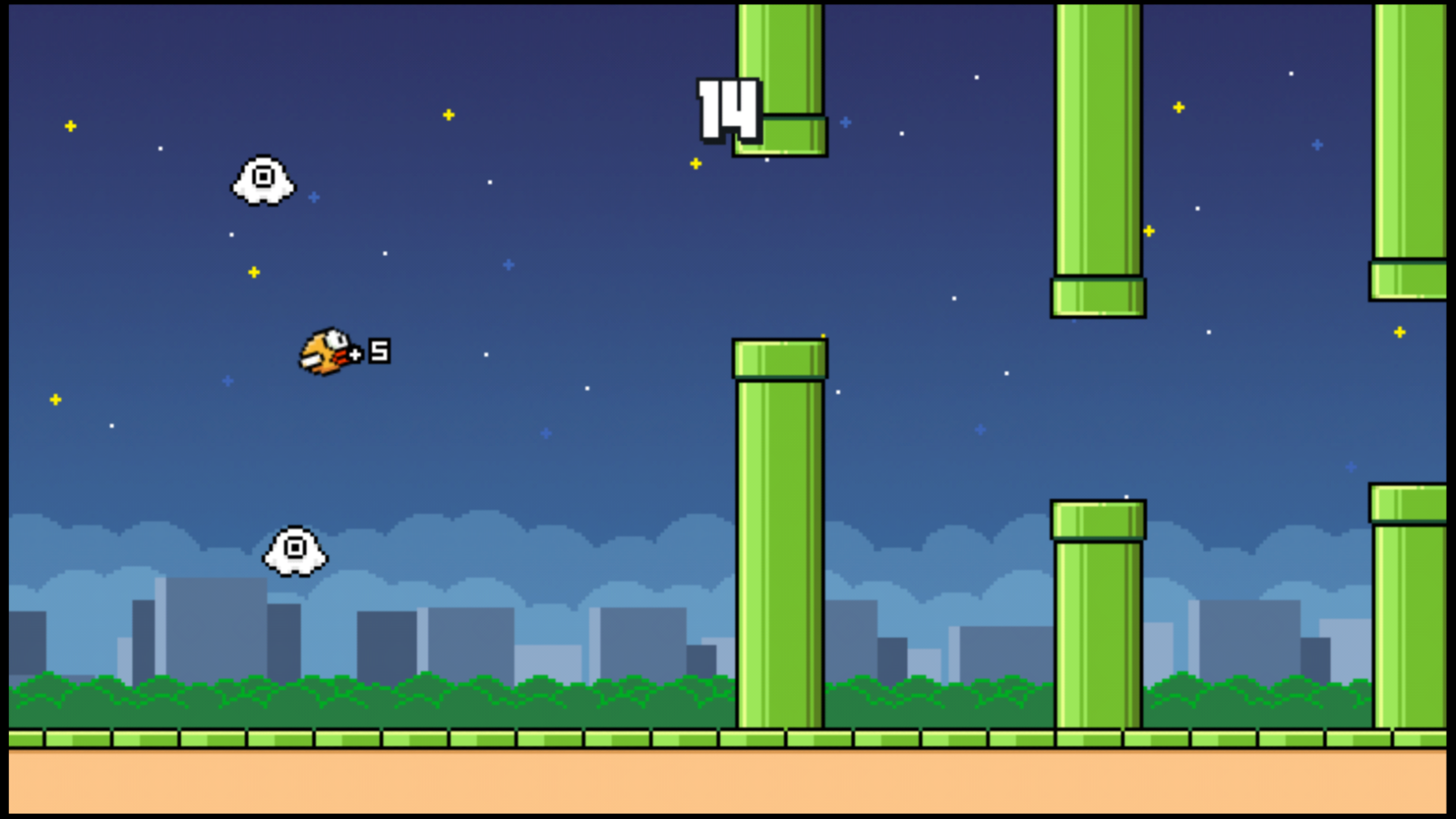}
	\caption{Flappy Bird: Graphical User Interface}
	\label{fig:flappybird:gui}
\end{figure}

Flappy Bird is a popular mobile phone game developed by Dong Nguyen in May 2013. The game objective is to control a bird by \textit{"flapping"} its wings to pass pipes, see Figure~\ref{fig:flappybird:gui}. The player is awarded one point for each pipe passed.

Flappy Bird is widely used in \gls{RL} research and was first introduced in \textit{Deep Reinforcement Learning for Flappy Bird}~\cite{Chen2015}. This report shows superhuman agent performance in the game using regular \gls{DQN} methods\footnote{Source code: \url{https://github.com/yenchenlin/DeepLearningFlappyBird}}.

OpenAI's gym platform implements Flappy Bird through PyGame Learning Environment\footnote{Available at: \url{https://github.com/ntasfi/PyGame-Learning-Environment}} (PLE). It supports both visual and non-visual state representation. The visual representation is an RGB image while the non-visual information includes vectorized data of the birds position, velocity, upcoming pipe distance, and position.

Figure~\ref{fig:flappybird:gui} illustrates the visual state representation of Flappy Bird. It is represented by an RGB Image with the dimension of $512\times288$. It is recommended that raw images are preprocessed to gray-scale and downscaled to $80\times80$. Flappy Bird is an excellent environment for \gls{RL} and provides adequate validation of new game environments introduced in this thesis. 


\chapter{Proposed Solutions}
\label{chap:solutions}

\begin{figure}
	\centering
	\includegraphics[width=\linewidth]{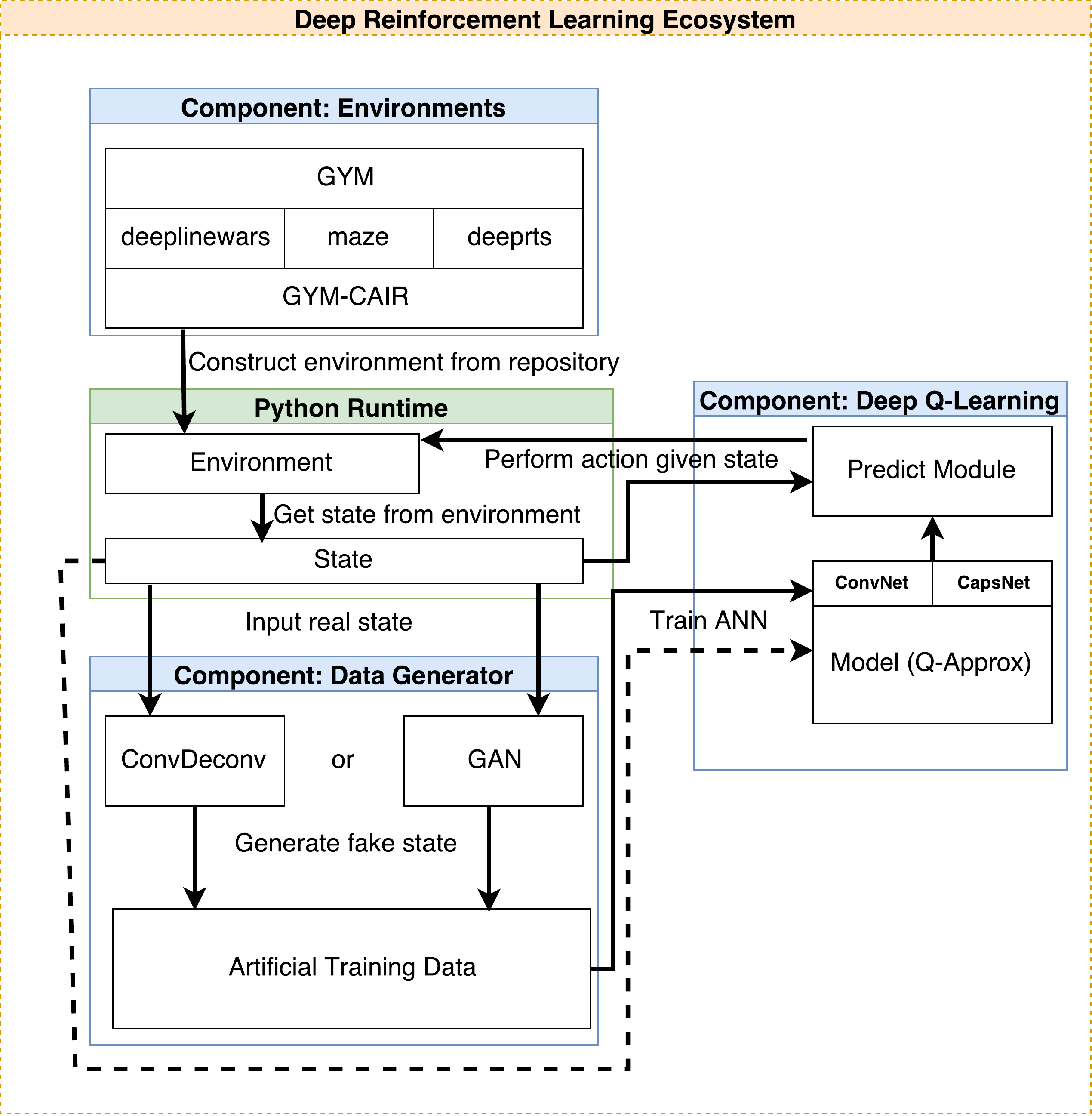}
	\caption{Proposed Deep Reinforcement Learning ecosystem}
	\label{fig:proposed_solution_ecosystem}
\end{figure}

Three key solutions are presented in this thesis. First is an architecture that provides a generic communication interface between the environments and the \gls{DRL} agents. Second is to apply Capsule Layers to \gls{DQN}, enabling the research into \gls{CapsNet} based \gls{RL} algorithms. The third is a novel technique for generating artificial training data for \gls{DQN} models. These components propose a \gls{DRL} ecosystem that is suited for research purposes, see Figure~\ref{fig:proposed_solution_ecosystem}.


\section{Environments}
\label{sec:solutions:environments}
OpenAI GYM is an open-source learning platform, exposing several game environments to the AI research community. There are many existing games available, but these are too simple because they have too easy game objectives. A game environment is \textit{registered} to the GYM platform by defining a \textit{scenario}. This scenario predefines the environment settings that determines the complexity. This type of registration is beneficial because it enables to construct multiple scenarios per game environment. An example of this would be the Maze environment, which contains scenarios for \textit{deterministic} and \textit{stochastic} gameplay for the different maze sizes. 

\begin{figure}
	\centering
	\includegraphics[width=10.5cm]{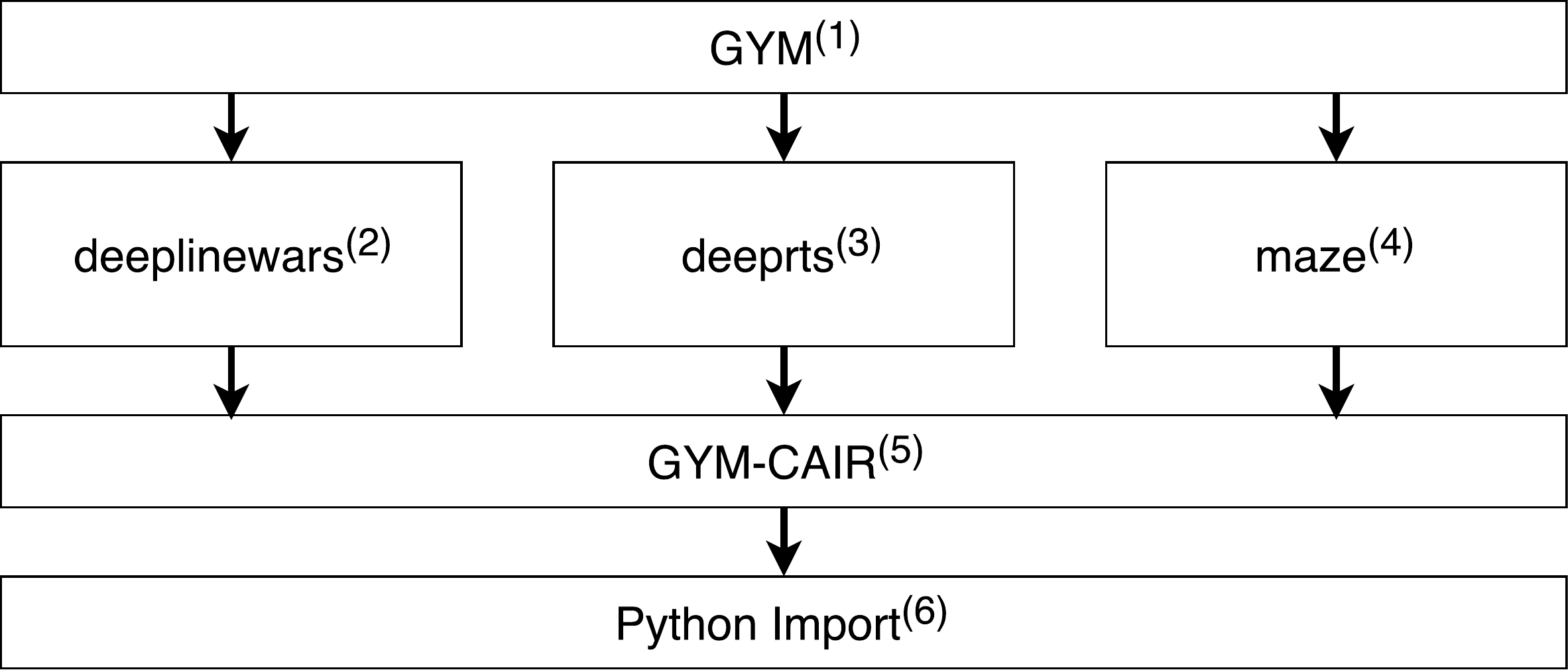}
	\caption{Architecture: gym-cair}
	\label{fig:gym_arch}
\end{figure}

Figure~\ref{fig:gym_arch} illustrates how the environment ecosystem is designed using OpenAI GYM. Environments are registered to the GYM\(_{(1)}\) platform.\\
Deep Line Wars\(_{(2)}\), Deep RTS\(_{(3)}\) and Maze\(_{(4)}\) are then added to a common repository, called \textit{gym-cair}\(_{(5)}\). This repository links together all environments, which can be imported via Python\(_{(6)}\).

\begin{algorithm}
	\caption{Generic GYM \gls{RL} routine}
	\label{alg:gym_flow}
	\begin{algorithmic}[1]
	\State $state_{x} = gym.reset$
	\State $terminal = False$

	\While {not $terminal$}
		\State $env.render$
		\State $a = env.action\_space.sample$
		\State $state_{x+1},~r_{x+1},~terminal,~info = env.step(a)$
		\State $state_{x} = state_{x+1}$
	\EndWhile
	\end{algorithmic}
\end{algorithm}

The benefit of using GYM is that all environments are constrained to a generic \gls{RL} interface, seen in Algorithm~\ref{alg:gym_flow}. The environment is initially reset by running \textit{gym.reset} function (Line 1). It is assumed that the environment does not start in a terminal state (Line 2). While the environment is not in a terminal state, the agent can perform actions (Line 5 and 6). This procedure is repeated until the environment reaches the terminal state.

By using this setup, it is far more trivial to perform experiments in the proposed environments. It also enables better comparison, because GYM ensures that the environment configuration is not altered while conducting the experiments.

\section{Capsule Networks}
\label{sec:solutions:capsnet}
Capsule Networks recently illustrated that a shallow \gls{ANN} could successfully classify the MNIST dataset of digits,  with state-of-the-art results, using considerably fewer parameters then in regular \gls{ConvNet}s. The idea behind \gls{CapsNet} is to interpret the input by identifying \textit{parts of the whole}, namely the objects of the input. \cite{Sabour2017} The objects are identified using Capsules that have the responsibility of finding specific objects in the whole. A capsule becomes active when the object it searches for exist.

It becomes significantly harder to use \gls{CapsNet} in \gls{RL}. The objective of \gls{RL} is to identify actions that are sensible to do in any given state. This means that actions become \textit{parts}, and the \textit{whole} becomes the state. Instead of classifying objects, the capsules now estimate a vector of the likelihood that an action is sensible to do in the current state.

Several issues need to be solved for \gls{CapsNet} to work properly in the environments outlined in Chapter~\ref{chap:env}. The first problem is the input size. The MNIST dataset of digits contains images of $28\times28\times1$ pixels, in contrast, game environments usually range between $64\times64\times1$ and $128\times128\times3$ pixels.

\begin{table}[]
	\centering
	\begin{tabular}{|l|ll|ll|}
		\hline
		\rowcolor[HTML]{BBDAFF} 
		Layer Name                & Output          & Params & Output          & Params \\ \hline
		Input                     & $28\times28\times1$   & 0          & $84\times84\times1$   & 0          \\
		Conv Layer       & $20\times20\times256$ & 20 992     & $76\times76\times256$ & 20 992     \\
		Primary Caps           & $6\times6\times256$   & 5 308 672  & $34\times34\times256$ & 5 308 672  \\
		Capsule Layer             & $16\times16$          & 2 359 296  & $16\times16$          & 75 759 616 \\
		Output                    & 16                    & 0          & 16                    & 0          \\ \hline
		\textbf{Parameters} & \multicolumn{2}{l|}{7 688 960}     & \multicolumn{2}{l|}{81 089 280}    \\ \hline
	\end{tabular}
	\caption{Capsule Networks: Dimension Comparison}
	\label{chap:capsnet:dimension_comparison}
\end{table}

\begin{figure}
	\centering
	\includegraphics[width=\linewidth]{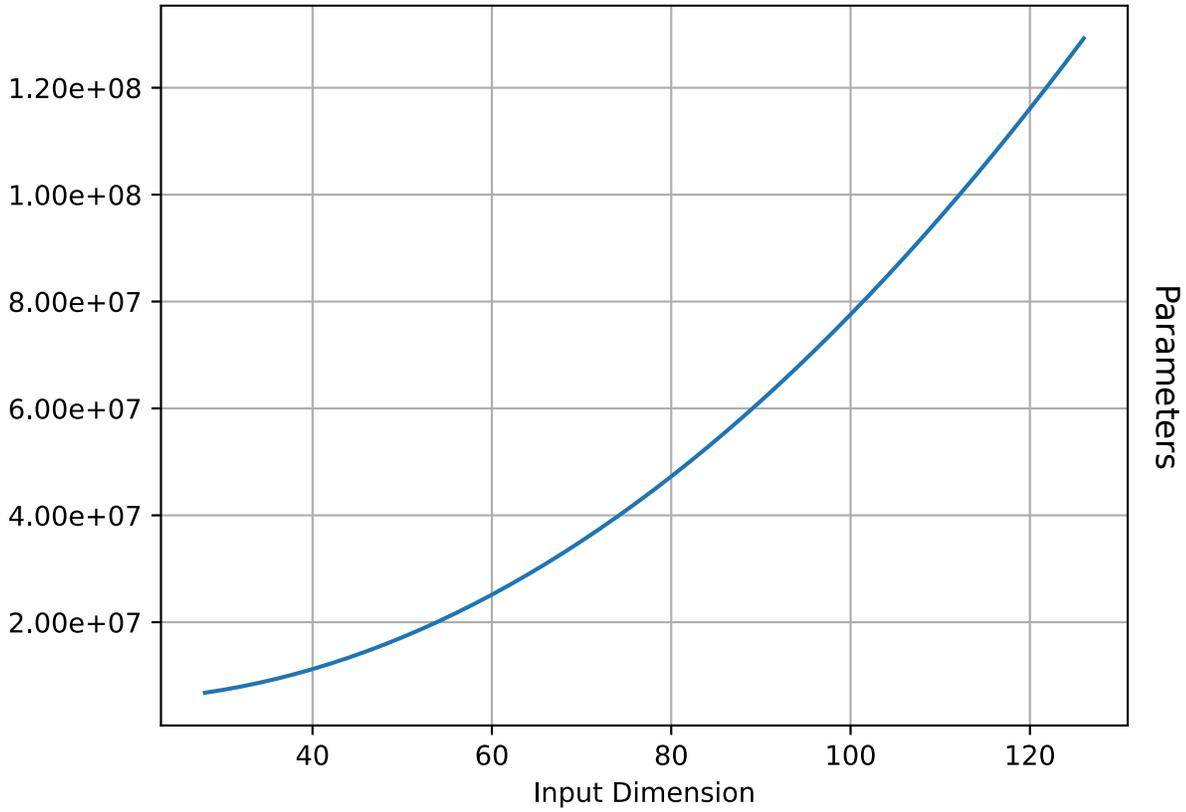}
	\caption{Capsule Networks: Parameter count for different input sizes}
	\label{fig:capsules:scalability}
\end{figure}

Table~\ref{chap:capsnet:dimension_comparison} illustrates the consequence of increasing the input size beyond the specified  $28\times28\times1$. By increasing the input size by a magnitude of 3 ($84\times84$), the model gains over $10\times$ parameters. Figure~\ref{fig:capsules:scalability} illustrates how parameters increase exponentially with the input size. In attempts to solve the scalability issue, several Convolutional Layers can be put in front of the \gls{CapsNet}. This enables the algorithm to extract feature maps from the original input, but it is crucial to not utilize any form of pooling prior the Capsule Layer. The whole reason to use Capsules is that it solves several problems with invariance that max-pooling possess.

\begin{figure}
	\centering
	\includegraphics[width=\linewidth]{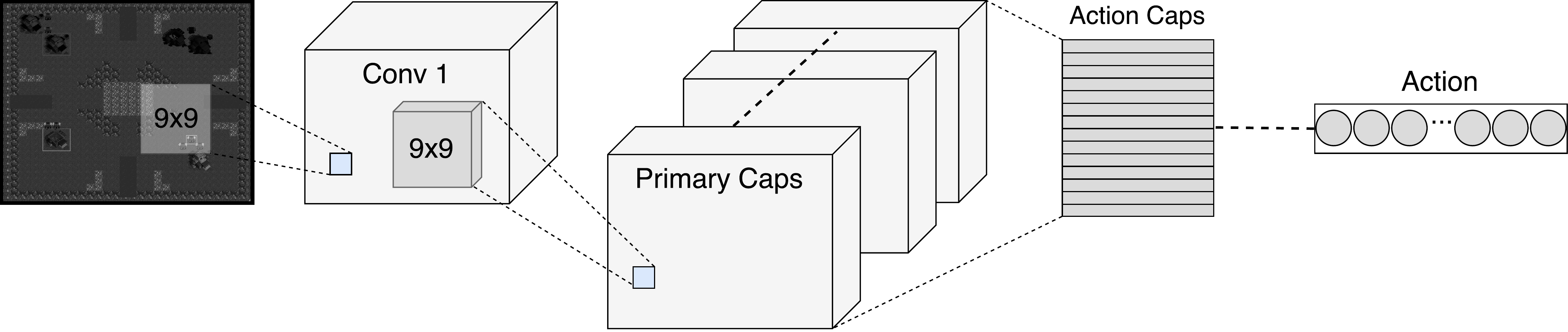}
	\caption{Capsule Networks: Architecture}
	\label{fig:capsules:architecture}
\end{figure}

Figure~\ref{fig:capsules:architecture} illustrates how a standard \gls{CapsNet} is structured, using a single Convolutional Layer. When a neural network is used, a question is defined to instruct the neural network to predict an answer. For a simple image classification task, the question is: \textit{what do you see in the image}. The neural network then tries to answer, by using its current knowledge: \textit{I see a bird}. The answer is then revealed to the neural network, which allows it to tune its response if it answered incorrectly. The same analogy can be used in an \gls{RL} problem.

The hope is that despite having several scalability issues, it is possible to accurately encode states into the correct capsules for each possible action in the environment. There are several methods to improve the training, but for this thesis, only primitive Q-Learning strategies will be used.


\section[Deep Q-Learning]{Deep Q-Learning\footnote{General knowledge of ANN, DQN, and CapsNet from Chapter \ref{chap:bg} is required.}}
\label{sec:solutions:dqn}

There are many different Deep Q-Learning algorithms available consisting of different hyper-parameters, network depth, experience replay strategies and learning rates.
The primary problem of \gls{DQN} is learning stability, and this is shown with the countless configurations found in the literature~\cite{Mnih2013, Mnih2015, Hausknecht2015, VanHasselt2015, Wang2015, Gu2016}. Refer to Section~\ref{sec:bg:rl:dql} for how the algorithm performs learning of the Q function.

\begin{table}[]
	\centering
\begin{tabular}{|llll|}
	\hline
	{\cellcolor[HTML]{BBDAFF}\textbf{}}& {\cellcolor[HTML]{BBDAFF}\textbf{Model}}           & {\cellcolor[HTML]{BBDAFF}\textbf{Paper}}                          & {\cellcolor[HTML]{BBDAFF}\textbf{Year}} \\ \hline
	1 & Vanilla DQN              & Mnih et al.~\cite{Mnih2013, Mnih2015}   & 2013/2015     \\ \hline
	2 & Deep Recurrent Q-Network & Hausknecht et al.~\cite{Hausknecht2015} & 2015          \\ \hline
	3 & Double DQN               & Hasselt et al.~\cite{VanHasselt2015}       & 2015          \\ \hline
	4 & Dueling DQN              & Wang et al.~\cite{Wang2015}             & 2015          \\ \hline
	5 & Continuous DQN           & Gu et al.~\cite{Gu2016}                 & 2016          \\ \hline
	6 & \multicolumn{3}{l|}{Deep Capsule Q-Network}                                        \\ \hline
	7 & \multicolumn{3}{l|}{Recurrent Capsule Q-Network}                                   \\ \hline
\end{tabular}
	\caption{Deep Q-Learning architectures in testbed}
	\label{tbl:tested_algorithms}
\end{table}

Models 1-4 (Figure \ref{tbl:tested_algorithms}) are the most commonly used \gls{DQN} architectures found in literature. Model 5 shows great potential in continuous environments, comparable to environments from Chapter~\ref{chap:env}. Models 6 and 7 are two novel approaches using Capsule Layers in conjunction with Convolution layers~\cite{Sabour2017, Xi2017}.

\begin{savenotes}
\begin{table}[]
	\centering
\begin{tabular}{|lllllll|}
	\hline
	\multicolumn{7}{|c|}{{\cellcolor[HTML]{BBDAFF}\textbf{Deep Q-Learning Models}}}                                                                                                                                                                                                                                                                                                                               \\ 
	\multicolumn{7}{|c|}{{{\cellcolor[HTML]{BBDAFF}\textit{(It is assumed that all models have a preceding input layer)}}}}                                                                                                                                                                                                                                                                                             \\ \hline
	& \textbf{Model}                                        & \textbf{Layer 1}                                    & \textbf{Layer 2}                                    & \textbf{Layer 3}                                    & \textbf{Layer 4}                                                       & \textbf{Layer 5}                                                             \\ \hline
	1 & \textbf{DQN}                                          & \begin{tabular}[c]{@{}l@{}}Conv\\ Relu\end{tabular} & \begin{tabular}[c]{@{}l@{}}Conv\\ Relu\end{tabular} & \begin{tabular}[c]{@{}l@{}}Conv\\ Relu\end{tabular} & \begin{tabular}[c]{@{}l@{}}Dense\\ Relu\end{tabular}                   & \begin{tabular}[c]{@{}l@{}}Output\\ Linear\end{tabular}                      \\ \hline
	2 & \textbf{DRQN}                                         & \begin{tabular}[c]{@{}l@{}}Conv\\ Relu\end{tabular} & \begin{tabular}[c]{@{}l@{}}Conv\\ Relu\end{tabular} & \begin{tabular}[c]{@{}l@{}}Conv\\ Relu\end{tabular} & LSTM                                                                   &                                                                              \\ \hline
	3 & \textbf{DDQN}                                         & \begin{tabular}[c]{@{}l@{}}Conv\\ Relu\end{tabular} & \begin{tabular}[c]{@{}l@{}}Conv\\ Relu\end{tabular} & \begin{tabular}[c]{@{}l@{}}Conv\\ Relu\end{tabular} & \begin{tabular}[c]{@{}l@{}}Dense\\ Relu\\ \\ Dense\\ Relu\end{tabular} & \begin{tabular}[c]{@{}l@{}}Output\\ Linear\\ \\ Output\\ Linear\end{tabular} \\ \hline
	4 & \textbf{DuDQN}                                        & \multicolumn{5}{l|}{\textit{Uses 2x DQN, Gradual updates from Target to Main}}                                                                                                                                                                                                                                             \\ \hline
	5 & \textbf{CDQN}                                         & \multicolumn{5}{l|}{\textit{Identical to DDQN but with different update strategy}}                                                                                                                                                                                                                                          \\ \hline
	6 & \textbf{DCQN}                                         & \begin{tabular}[c]{@{}l@{}}Conv\\ Relu\end{tabular} & \begin{tabular}[c]{@{}l@{}}Conv\\ Relu\end{tabular} & \begin{tabular}[c]{@{}l@{}}Conv\\ Relu\end{tabular} & Capsule                                                                & OutCaps                                                                      \\ \hline
	7 & \textbf{RCQN~\footnote{Time Distributed / Recurrent}} & \begin{tabular}[c]{@{}l@{}}Conv\\ Relu\end{tabular} & \begin{tabular}[c]{@{}l@{}}Conv\\ Relu\end{tabular} & \begin{tabular}[c]{@{}l@{}}Conv\\ Relu\end{tabular} & Capsule                                                                & OutCaps                                                                      \\ \hline
\end{tabular}
	\caption{Deep Q-Learning architectures}
	\label{tbl:dqn_architectures}
\end{table}
\end{savenotes}

\begin{table}[]
	\centering
	\begin{tabular}{|lll}
		\hline
		\multicolumn{3}{|c|}{{\cellcolor[HTML]{BBDAFF}\textbf{Deep Q-Learning Hyperparameters}}}                                                                        \\ 
		{\cellcolor[HTML]{BBDAFF}\textbf{Parameter}} & {\cellcolor[HTML]{BBDAFF}\textbf{Value Range}}                                                   & \multicolumn{1}{l|}{{\cellcolor[HTML]{BBDAFF}\textbf{Default}}}   \\ \hline
		Learning Rate      & 0.0-1.0                                                                & \multicolumn{1}{l|}{$1\mathrm{e}{-04}$} \\ \hline
		Discount Factor    & 0.0-1.0                                                                & \multicolumn{1}{l|}{0.99}               \\ \hline
		Loss Function      & {[}Huber, MSE{]}                                                       & \multicolumn{1}{l|}{Huber}              \\ \hline
		Optimizer          & {[}SGD, Adam, RMSProp{]}                                               & \multicolumn{1}{l|}{Adam}               \\ \hline
		Batch Size         & $1\rightarrow\infty$                                                   & \multicolumn{1}{l|}{32}                 \\ \hline
		Memory Size        & $1\rightarrow\infty$                                                   & \multicolumn{1}{l|}{1 000 000}          \\ \hline
		$\epsilon_{min}$   & $0.0\rightarrow1.0$                                                    & \multicolumn{1}{l|}{0.10}               \\ \hline
		$\epsilon_{max}$   & $0.0\rightarrow1.0~\textbf{and} > \epsilon_{min}$                      & \multicolumn{1}{l|}{1.0}                \\ \hline
		$\epsilon_{start}$ & $\epsilon_{start} \in \left \{\epsilon_{min}, \epsilon_{max}\right \}$ & \multicolumn{1}{l|}{1.0}                \\ \hline
	\end{tabular}
	\caption{Deep Q-Learning hyper-parameters}
	\label{tbl:dqn_hyperparameters}
\end{table}

Models 1-7 are implemented in the Keras/Tensorflow framework according to the definitions found in the illustrated papers. Table~\ref{tbl:dqn_architectures} shows the architecture of the \gls{DQN} models found in Table~\ref{tbl:tested_algorithms}. Filter and stride count is intentionally left out because these are considered as hyper-parameters. Hyperparameters are manually tuned by trial and error. Table~\ref{tbl:dqn_hyperparameters} outlines the parameters that are tuned individually for each of the architectures.

\section{Artificial Data Generator}
\label{sec:solutions:datagen}
The Artificial Data Generator component from Figure~\ref{fig:proposed_solution_ecosystem} is an attempt to shorten the exploration phase in \gls{RL}. By generating artificial training data, the hope is that \gls{DQN} models can learn features that were never experienced within the environment. The proposed algorithm could be able to predict these future states, $s_{i + 1}$ given $s_{i}$ conditioned on action $a$ in the generator function $s_{i + 1} = G(s_{i}|a)$~\cite{Mirza2014}. The initial plan was to utilize adversarial generative networks but was not able to generate conditioned states successfully. Instead, an architecture called Conditional Convolution Deconvolution Network was developed that use \gls{SDG} to update parameters (Section~\ref{sec:solutions:datagen:ccdn}).

\label{sec:solutions:datagen:ccdn}
\textbf{Conditional Convolution Deconvolution Network} (\gls{CCDN}) is an architecture that tries to predict the consequence of a condition applied to an image. A state is conditioned on a action to predict future game states.

\begin{figure}
	\centering
	\includegraphics[width=7.5cm]{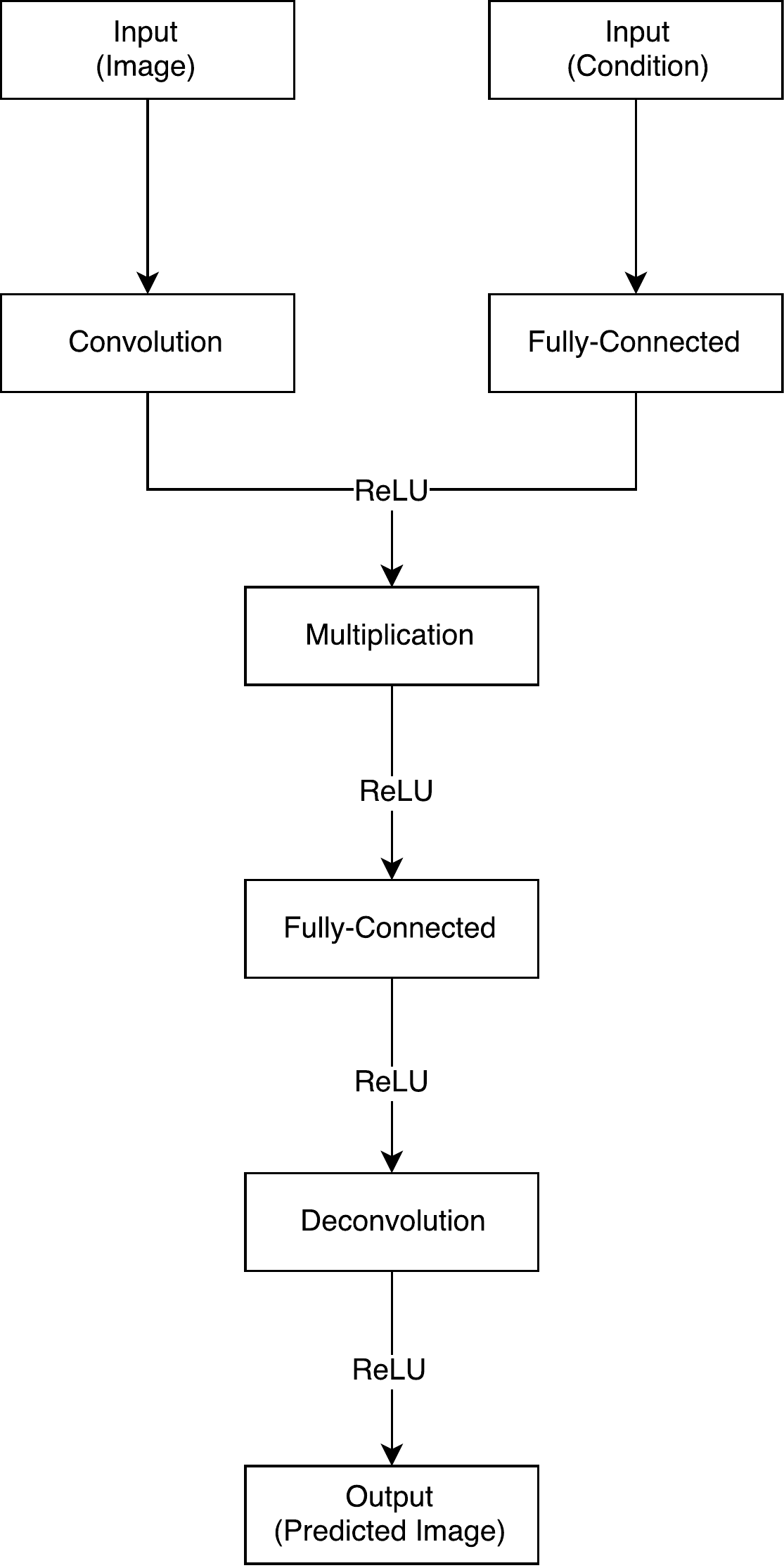}
	\caption{Architecture: \gls{CCDN}}
	\label{fig:conv_deconv_architecture}
\end{figure}

Figure~\ref{fig:conv_deconv_architecture} illustrates the general idea of the model. The model is designed using two input streams, image and condition steam. The image stream is a series of convolutional layers following a fully-connected layer. The conditional stream contains fully-connected layers where the last layer matches the number neurons in the last layer of the image steam. These streams are then multiplied following a fully-connected layer that encodes the conditioned states. The conditioned state is then reconstructed using deconvolutions. The output layer is the final reconstructed image of the predicted next state given condition.

The process of training this model is supervised as it consumes data from the experience replay buffer gathered by \gls{RL} agents. The model is trained by fetching a memory from the experience replay memory \textit{($s_{i}$, a, $s_{i + 1}$)} where $s_{i}$ is the current state, $a$ is the action, and $s_{i + 1}$ is the transition $T(s_{i}, a)$. \gls{CCDN} generates an artificial state $\hat{s}_{i + 1}$ by using the generator model $G(s_{i} | a, \theta)$. The parameters $\theta$ is optimized using \gls{SDG}, and the loss is calculated using \gls{MSE}. 

\begin{equation}
\label{eq:prop:mse}
MSE = \frac{1}{n}\sum_{i=1}^{n}(S_{i+1} - \hat{S}_{i+1})^{2}
\end{equation}

Equation~\ref{eq:prop:mse} (Equation~\ref{eq:loss:mse}; the predicted value \textit{y} is now denoted \textit{S}) is simple in that the loss is decreased when the value of the predicted state $\hat{S}_{i+1}$ gets closer to $S_{i+1}$. 

Table~\ref{tbl:ccdn_state_generation} illustrates how states are generated using \gls{CCDN}. It is assumed that it is possible to transition between states given an action, to create training data. When sufficient training data is collected, the recorded state data is used to estimate future states. In this example, there is a $2\times2$ grid where the agent is a red square with the actions, \textit{up, down, left, and right}. This yields a theoretical maximum state-space of 4 states with $256$ possible transitions (4 actions per cell = $4^4$ possible state and action combinations). When a portion of the state-space is explored trough random-play the \gls{CCDN} algorithm can train by comparing the predictions against real data. The goal is for the model to converge towards learning the transition function of the environment, to continue generating future states without any interaction with the environment. It is likely that the model are able to converge towards the optimal solution for more than a single time-step in the future.

\begin{table}[]
	\centering
	\begin{tabular}{lllllllll}
		&                                                  &                                                  &                                         & \multicolumn{2}{l}{}                                                                                &                                           & \multicolumn{2}{l}{}                                                                                \\
		& \cellcolor[HTML]{FE0000}{\color[HTML]{FE0000} 1} & \cellcolor[HTML]{333333}{\color[HTML]{333333} 0} &                                         & \cellcolor[HTML]{333333}{\color[HTML]{333333} 0} & \cellcolor[HTML]{FE0000}{\color[HTML]{FE0000} 1} &                                           & \cellcolor[HTML]{333333}{\color[HTML]{333333} 0} & \cellcolor[HTML]{333333}{\color[HTML]{333333} 0} \\
		\multirow{-2}{*}{Real States}                          & \cellcolor[HTML]{333333}{\color[HTML]{333333} 0} & \cellcolor[HTML]{333333}{\color[HTML]{333333} 0} & \multirow{-2}{*}{$T(s_{0}, A_{right})$} & \cellcolor[HTML]{333333}{\color[HTML]{333333} 0} & \cellcolor[HTML]{333333}{\color[HTML]{333333} 0} & \multirow{-2}{*}{$T(s_{1}, A_{down})$}    & \cellcolor[HTML]{333333}{\color[HTML]{333333} 0} & \cellcolor[HTML]{FE0000}{\color[HTML]{FE0000} 1} \\
		& $s_{0}$                                          &                                                  &                                         & \multicolumn{2}{l}{$s_{1}$}                                                                         &                                           & \multicolumn{2}{l}{$s_{2}$}                                                                         \\ \hline
		&                                                  &                                                  &                                         & \multicolumn{2}{l}{}                                                                                &                                           & \multicolumn{2}{l}{}                                                                                \\ \cline{2-3}
		\multicolumn{1}{l|}{}                                  & \multicolumn{2}{c|}{}                                                                               &                                         & \cellcolor[HTML]{333333}{\color[HTML]{333333} 0} & \cellcolor[HTML]{FE0000}{\color[HTML]{FE0000} 1} &                                           & \cellcolor[HTML]{333333}{\color[HTML]{333333} 0} & \cellcolor[HTML]{333333}{\color[HTML]{333333} 0} \\
		\multicolumn{1}{l|}{\multirow{-2}{*}{Generated States}} & \multicolumn{2}{c|}{\multirow{-2}{*}{N/A}}                                                          & \multirow{-2}{*}{$G(s_{0},A_{right})$}  & \cellcolor[HTML]{333333}{\color[HTML]{333333} 0} & \cellcolor[HTML]{333333}{\color[HTML]{333333} 0} & \multirow{-2}{*}{$G(\hat{s}_1,A_{down})$} & \cellcolor[HTML]{333333}{\color[HTML]{333333} 0} & \cellcolor[HTML]{FE0000}{\color[HTML]{FE0000} 1} \\ \cline{2-3}
		&                                                  &                                                  &                                         & \multicolumn{2}{l}{$\hat{s}_1$}                                                                     &                                           & \multicolumn{2}{l}{$\hat{s}_2$}                                                                    
	\end{tabular}
	\caption{Proposed prediction cycle for \gls{CCDN}}
	\label{tbl:ccdn_state_generation}
\end{table}

\part{Experiments and Results}

\chapter[Conditional Convolution Deconvolution Network]{Conditional Convolution Deconvolution Network}
\chaptermark{CCDN}
\label{chap:results:ccdn}
This chapter presents Conditional Convolution Deconvolution Network (\gls{CCDN}). The purpose of the \gls{CCDN} algorithm is to generate artificial training data for Deep Reinforcement Learning algorithms. The data is generated from the game environments introduced in Chapter~\ref{chap:env}. The goal is to generate high-quality training data that can be used to train algorithms without self-play. The algorithm is used on the following game environments:
\begin{itemize}
	\item FlashRL: Multitask (Section~\ref{sec:env:flashrl}),
	\item Deep Line Wars (Section~\ref{sec:env:deeplinewars}),
	\item Deep Maze (Section~\ref{sec:env:maze}), and
	\item Flappy Bird (Section~\ref{sec:env:flappybird}).
\end{itemize}
Deep RTS (Section~\ref{sec:env:deeprts}) is excluded from these tests because it does not support image state-representation\footnote{Deep RTS image state-representation is planned for \hyperref[sec:conclusion:future_work]{future work}}

A dataset consisting of 100 000 unique state transitions is collected for all environments using random-play strategies. A training set is created, consisting of 60 000 transitions (60\%), and the remaining 40\% as a test. The training for \gls{CCDN} took approximately 160 hours per game environment when using hardware listed in Appendix~\ref{appendix:hwspec}.

\renewcommand{\sectionbreak}{}

\section{Introduction}
\label{sec:results:datagen:ccdn}
\gls{CCDN} predicts the future states by conditioning current state on a action. Figure~\ref{fig:conv_deconv_architecture} illustrates the architecture used in these experiments. To calculate the loss, \gls{MSE} (Equation~\ref{eq:prop:mse}) was used during training. The model is tested using 32, 64, 128, 256, and 512 neurons in the fully-connected layer. Depending on the neuron count, the model has approximately 13 000 000 to 67 000 000 parameters in total.

It is beneficial to have a significant amount of parameters because it allows the model to encode more data. The drawback of this is that the model uses more memory, and takes longer to train. The aim is to train the model for 10 000 epoch at a maximum of 168 hours. For this reason, the algorithm used 32 neurons in the hidden layers which gave reasonably good results for some environments.

Conditioned actions are not shown in the generated images from these experiments. This is because the precision is still too coarse, and the generated future states are yet too far from the ground truth. These results are impressive for some environments, and there is a possibility that the generated samples can be used in conjunction with real samples to train \gls{DQN} models.

\renewcommand{\sectionbreak}{\clearpage}

\section{Deep Line Wars}
\label{sec:results:datagen:dlw}
Deep Line Wars show excellent results when generating data using \gls{CCDN} to generate future states \(\hat{s} = G(s|a)\). Table~\ref{tbl:ccdn:dlw} illustrates the transition from real states (Left side) to generated future states by training \gls{CCDN} using \gls{SDG}.

\begin{figure}
	\centering
	\includegraphics[width=0.7\linewidth]{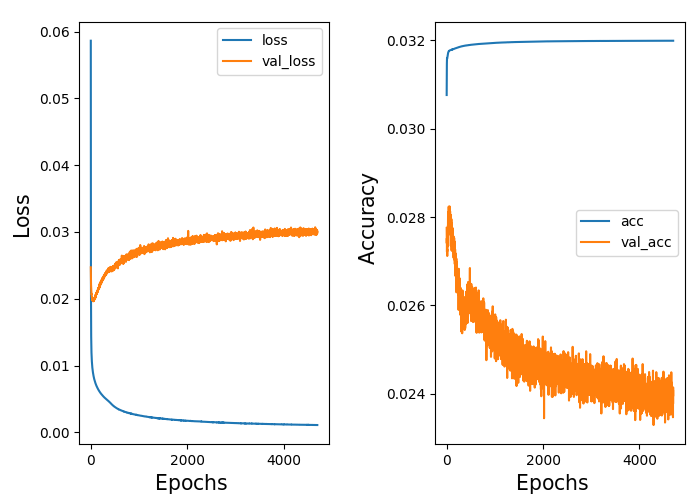}
	\caption{\gls{CCDN}: Deep Line Wars: Training Performance}
	\label{fig:ccdn:dlw_perf}
\end{figure}
\gls{CCDN} was not able to converge, but it is possible that this is due to our low neuron count of 32 in the fully-connected layer. Figure~\ref{fig:ccdn:dlw_perf} shows that the loss inclined gradually while the accuracy declined. Loss and accuracy do not reflect the generated images seen in Table~\ref{tbl:ccdn:dlw}. The observed transitions at Day 5-6.5 illustrate realistic transition behavior between states. Observations indicate that \gls{CCDN} learns input features like:
\begin{itemize}
	\item Background intensity (Represents health points)
	\item Possible mouse position (White square)
	\item Possible unit positions
	\item Building positions
\end{itemize}
The model is still not able to correctly predict the movement of units. This could potentially be solved by stacking several state transitions before predicting future states~\cite{Chen2015}. This could be done using \gls{ConvNet}s or the use of recurrence in neural networks (RNN).

\begin{table}[]
	\centering
	\begin{tabular}{|c|l|}
		\hline
		\multicolumn{2}{|c|}{\cellcolor[HTML]{BBDAFF}\textbf{Deep Line Wars: Conditioned State Transitions}}                                                                         \\ \hline
		Day 1                                                                                   & \multicolumn{1}{c|}{Day 4}                                         \\
		\multicolumn{1}{|l|}{\includegraphics[width=0.37\linewidth]{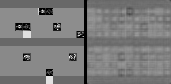}} & \includegraphics[width=0.37\linewidth]{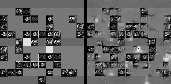}  \\
		Day 1.5                                                                                 & \multicolumn{1}{c|}{Day 4.5}                                       \\
		\multicolumn{1}{|l|}{\includegraphics[width=0.37\linewidth]{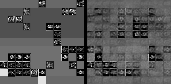}} & \includegraphics[width=0.37\linewidth]{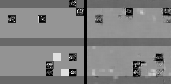}  \\
		Day 2                                                                                   & \multicolumn{1}{c|}{Day 5}                                         \\
		\multicolumn{1}{|l|}{\includegraphics[width=0.37\linewidth]{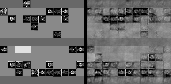}} & \includegraphics[width=0.37\linewidth]{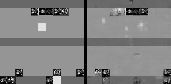}  \\
		Day 2.5                                                                                 & \multicolumn{1}{c|}{Day 5.5}                                       \\
		\multicolumn{1}{|l|}{\includegraphics[width=0.37\linewidth]{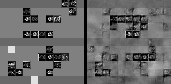}} & \includegraphics[width=0.37\linewidth]{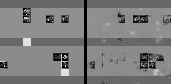} \\
		Day 3                                                                                   & \multicolumn{1}{c|}{Day 6}                                         \\
		\multicolumn{1}{|l|}{\includegraphics[width=0.37\linewidth]{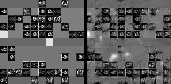}} & \includegraphics[width=0.37\linewidth]{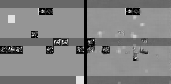} \\
		Day 3.5                                                                                 & \multicolumn{1}{c|}{Day 6.5}                                       \\
		\multicolumn{1}{|l|}{\includegraphics[width=0.37\linewidth]{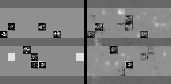}} & \includegraphics[width=0.37\linewidth]{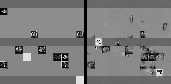} \\ \hline
	\end{tabular}
	\caption{CCDN: Deep Line Wars}
	\label{tbl:ccdn:dlw}
\end{table}

\section{Deep Maze}
\label{sec:results:datagen:ccdn:deepmaze}

\begin{figure}
	\centering
	\includegraphics[width=0.7\linewidth]{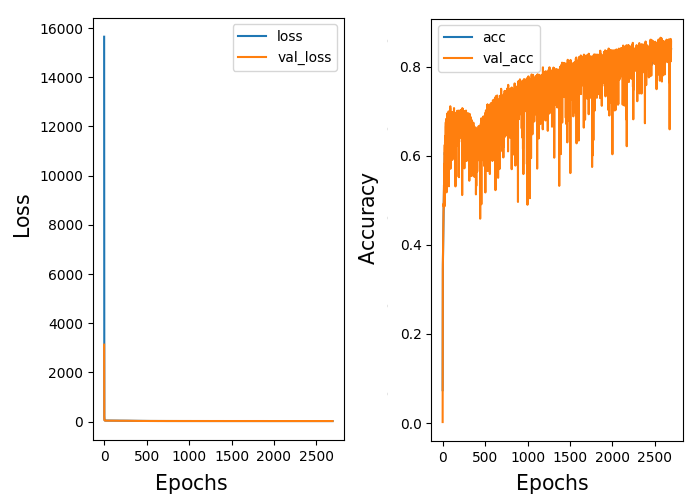}
	\caption{\gls{CCDN}: Deep Maze: Training Performance}
	\label{fig:ccdn:maze_perf}
\end{figure}

Deep Maze should be considered as one of the more straightforward environments to generate high-quality training data because it has the simplest state-space. From Table~\ref{fig:ccdn:maze} it is clear that \gls{CCDN} recognized features like the maze structure early in the training process. Figure~\ref{fig:ccdn:maze_perf} illustrates that \gls{CCDN} converged quickly, having a loss near 0 at the 5th epoch of training. High accuracy was reported during training when using \gls{MSE} as the loss function. By inspecting the produced images manually, it was clear that \gls{CCDN} did not learn how to predict the position of the player inside the maze. Hallways inside the maze did not show any sensible information about the actual location of the player. Instead, the maze hallways were generated as random noise. There are however some parts of the maze that presents less noise, indicating that the player did not visit these locations as frequently.

\begin{table}[]
	\centering
	\begin{tabular}{|c|l|}
		\hline
		\multicolumn{2}{|c|}{\cellcolor[HTML]{BBDAFF}\textbf{Deep Maze: Conditioned State Transitions}}                                                                         \\ \hline
		Day 1                                                                                   & \multicolumn{1}{c|}{Day 4}                                         \\
		\multicolumn{1}{|l|}{\includegraphics[width=0.37\linewidth]{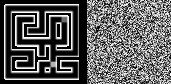}} & \includegraphics[width=0.37\linewidth]{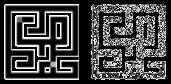}  \\
		Day 1.5                                                                                 & \multicolumn{1}{c|}{Day 4.5}                                       \\
		\multicolumn{1}{|l|}{\includegraphics[width=0.37\linewidth]{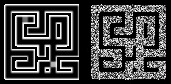}} & \includegraphics[width=0.37\linewidth]{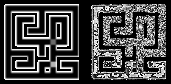}  \\
		Day 2                                                                                   & \multicolumn{1}{c|}{Day 5}                                         \\
		\multicolumn{1}{|l|}{\includegraphics[width=0.37\linewidth]{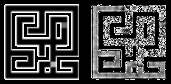}} & \includegraphics[width=0.37\linewidth]{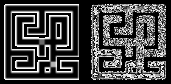}  \\
		Day 2.5                                                                                 & \multicolumn{1}{c|}{Day 5.5}                                       \\
		\multicolumn{1}{|l|}{\includegraphics[width=0.37\linewidth]{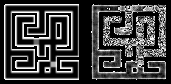}} & \includegraphics[width=0.37\linewidth]{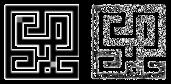} \\
		Day 3                                                                                   & \multicolumn{1}{c|}{Day 6}                                         \\
		\multicolumn{1}{|l|}{\includegraphics[width=0.37\linewidth]{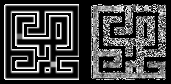}} & \includegraphics[width=0.37\linewidth]{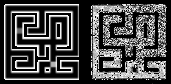} \\
		Day 3.5                                                                                 & \multicolumn{1}{c|}{Day 6.5}                                       \\
		\multicolumn{1}{|l|}{\includegraphics[width=0.37\linewidth]{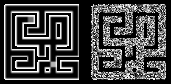}} & \includegraphics[width=0.37\linewidth]{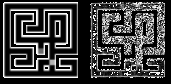} \\ \hline
	\end{tabular}
	\caption{CCDN: Deep Maze}
	\label{fig:ccdn:maze}
\end{table}

\section{FlashRL: Multitask}
\label{sec:results:datagen:ccdn:flashrl}

\begin{figure}
	\centering
	\includegraphics[width=0.7\linewidth]{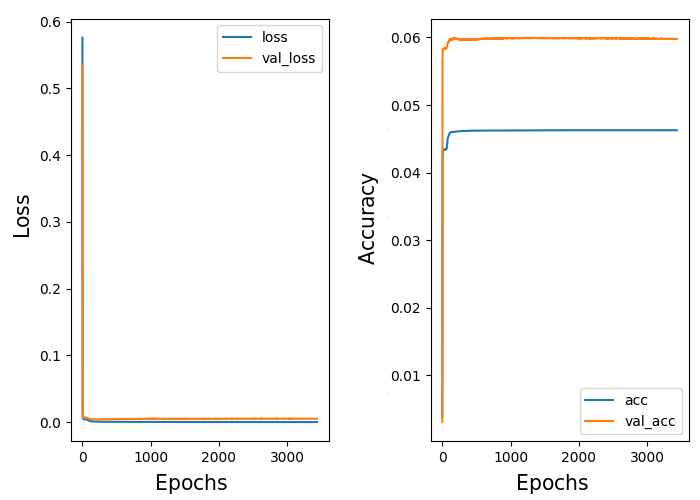}
	\caption{\gls{CCDN}: FlashRL: Training Performance}
	\label{fig:ccdn:flashrl_perf}
\end{figure}

\gls{CCDN} produced high-quality state transitions when applying it to Flash RL: Multitask. Since Multitask is an environment consisting of several different scenes (Menu, Stage 1, Stage 2), it was expected that it would fail to generate sensible output. Table~\ref{fig:ccdn:flashrl} illustrates that \gls{CCDN} was able to extract features from all states and map it to correct action. Transitions from Day 2.5 and Day 3.5 illustrates a slight change in the paddle tilt and the position of the ball. This shows that the algorithm can to some extend understand game mechanics. In addition to this, \gls{CCDN} can draw the menu including a slight change in the mouse position.
The results show that \gls{CCDN} can learn to extract:
\begin{itemize}
	\item The current scene layout
	\item Primitive physics
\end{itemize}

Figure~\ref{fig:ccdn:flashrl_perf} illustrates that \gls{CCDN} did not reach more than 5\% accuracy at training time even though the loss was close to zero. It is not clear what is causing the training instability because measuring loss of the images manually using \gls{MSE} gave far better accuracy for most training samples. The results indicate that \gls{CCDN} did indeed learn to extract features from the Multitask environment.

\begin{table}[]
	\centering
	\begin{tabular}{|c|l|}
		\hline
		\multicolumn{2}{|c|}{\cellcolor[HTML]{BBDAFF}\textbf{Flash RL: Conditioned State Transitions}}                                                                         \\ \hline
		Day 1                                                                                   & \multicolumn{1}{c|}{Day 4}                                         \\
		\multicolumn{1}{|l|}{\includegraphics[width=0.37\linewidth]{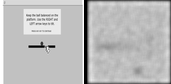}} & \includegraphics[width=0.37\linewidth]{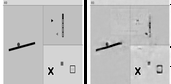}  \\
		Day 1.5                                                                                 & \multicolumn{1}{c|}{Day 4.5}                                       \\
		\multicolumn{1}{|l|}{\includegraphics[width=0.37\linewidth]{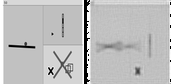}} & \includegraphics[width=0.37\linewidth]{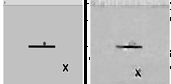}  \\
		Day 2                                                                                   & \multicolumn{1}{c|}{Day 5}                                         \\
		\multicolumn{1}{|l|}{\includegraphics[width=0.37\linewidth]{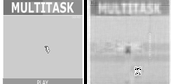}} & \includegraphics[width=0.37\linewidth]{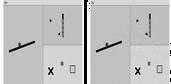}  \\
		Day 2.5                                                                                 & \multicolumn{1}{c|}{Day 5.5}                                       \\
		\multicolumn{1}{|l|}{\includegraphics[width=0.37\linewidth]{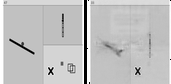}} & \includegraphics[width=0.37\linewidth]{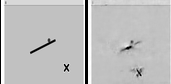} \\
		Day 3                                                                                   & \multicolumn{1}{c|}{Day 6}                                         \\
		\multicolumn{1}{|l|}{\includegraphics[width=0.37\linewidth]{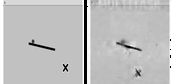}} & \includegraphics[width=0.37\linewidth]{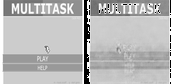} \\
		Day 3.5                                                                                 & \multicolumn{1}{c|}{Day 6.5}                                       \\
		\multicolumn{1}{|l|}{\includegraphics[width=0.37\linewidth]{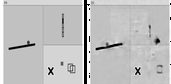}} & \includegraphics[width=0.37\linewidth]{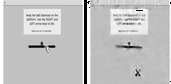} \\ \hline
	\end{tabular}
	\caption{CCDN: FlashRL: Multitask}
	\label{fig:ccdn:flashrl}
\end{table}

\section{Flappy Bird}
\label{sec:results:datagen:ccdn:flappybird}

\begin{figure}
	\centering
	\includegraphics[width=0.7\linewidth]{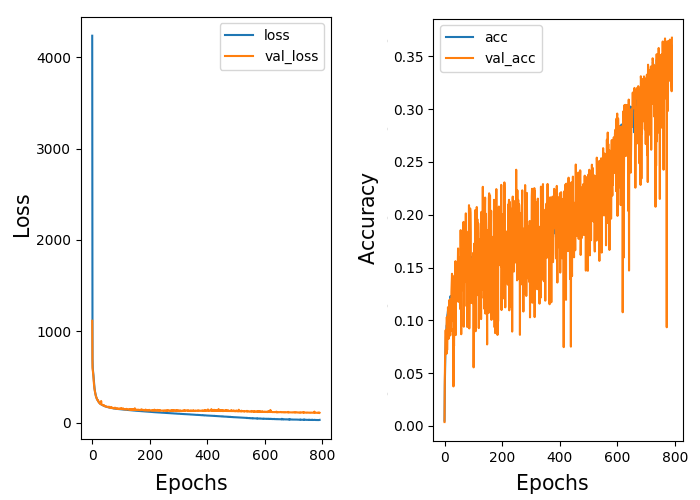}
	\caption{\gls{CCDN}: Flappy Bird: Training Performance}
	\label{fig:ccdn:flappybird_perf}
\end{figure}

Table~\ref{fig:ccdn:flappybird} illustrates the generated transitions for the third party game Flappy Bird. Figure~\ref{fig:ccdn:flappybird_perf} show that \gls{CCDN} has a gradual decrease in the loss while the accuracy increases to approximately 35\%. Flappy Bird has the highest accuracy for the tested game environments, but observations shows that \gls{CCDN} is only able to generate noise.

The reason is that Flappy Bird has a scrolling background, meaning that \gls{CCDN} must encode a lot more data than in the other environments. Because of this, \gls{CCDN} could not determine how to generate future state representations for this game. 

It is expected that this problem could be mitigated by performing data preprocessing. Literature indicates that \gls{RL} algorithms often strip away the background to simplify the game-state~\cite{Chen}. Also, it is likely that \gls{CCDN} could successfully encode Flappy Bird with additional parameters, but this would increase the training time to several weeks.

\begin{table}[]
	\centering
	\begin{tabular}{|c|}
		\hline
		\rowcolor[HTML]{BBDAFF} 
		\textbf{Flappy Bird: Conditioned States}                                                        \\ \hline
		Day 1                                                                                         \\
		\multicolumn{1}{|l|}{\includegraphics[width=0.60\linewidth]{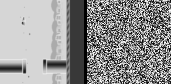}} \\
		Day 3                                                                                         \\
		\multicolumn{1}{|l|}{\includegraphics[width=0.60\linewidth]{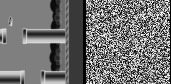}} \\
		Day 4                                                                                        \\
		\multicolumn{1}{|l|}{\includegraphics[width=0.60\linewidth]{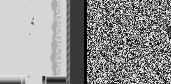}} \\
		Day 6                                                                                         \\
		\multicolumn{1}{|l|}{\includegraphics[width=0.60\linewidth]{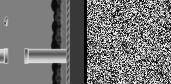}} \\ \hline
	\end{tabular}
	\caption{CCDN: Flappy Bird}
	\label{fig:ccdn:flappybird}
\end{table}

\section{Summary}
\gls{CCDN} is a novel algorithm suited for generating artificial training data for \gls{RL} algorithms and shows great potential for some environments. The results indicate that \gls{CCDN} has issues in game environments with a sparse state-space representation. Flappy Bird illustrates the problem well because \gls{CCDN} generates noise instead of future states for action and state pairs. One method to combat this problem may be to increase the neuron count for the fully-connected layer in the \gls{CCDN} model.

\gls{ANN} based algorithms frequently suffer from training instability. The results show that the \gls{CCDN} algorithm was not able to accurately determine the loss using regular \gls{MSE}. This could potentially be the cause of the training instability because the optimizer would not be able to determine how well it is doing when updating network parameters. It is likely that replacing the \gls{MSE} loss function could improve the generated images drastically.

The results presented in this Chapter shows excellent potential in using \gls{CCDN} for generation of artificial training data for game environments. It shows excellent performance in Deep Line Wars and Flash RL: Multitask and could potentially reduce the required amount of exploration in \gls{RL} algorithms

\chapter{Deep Q-Learning}
\label{chap:results:dqn}
This chapter presents experimental results of the research done using Deep Q-Learning with \gls{CapsNet} and \gls{ConvNet} based models. The goal is to use \gls{CapsNet} in Deep Q-Learning to solve the environments from Chapter~\ref{chap:env}.

\gls{RL} algorithms are known to be computationally intensive and are thus difficult to train for environments with large state-spaces~\cite{Duan2016}. Models are trained using hardware specified in Appendix~\ref{appendix:hwspec}. Chapter~\ref{chap:solutions} proposed 7 \gls{DQN} architectures that could potentially control an agent well within the environments. Model 1 and 6 from Table~\ref{tbl:tested_algorithms} was selected as the primary research area to limit the scope of this thesis \footnote{Training time for 7 models in 5 environments: $7\times5\times7 = 245$ days (Approx 7 days per experiment)}. To increase training stability for all environments, hyper-parameters from Table~\ref{tbl:dqn_hyperparameters} is tuned further per environment. The datasets are populated with 20\% artificial training data, generated from \gls{CCDN}. Table~\ref{tbl:dqn:hyper-parameters} illustrates updated hyper-parameters that performed best when experimenting with \gls{CapsNet} and \gls{ConvNet} based models. The \gls{DQN} models use \gls{SDG} to optimize its parameters. Initial training data is sampled using random-play strategies, gradually moving into exploitation using $\epsilon$-greedy.
\begin{table}
	\begin{tabular}{|l|c|c|c|c|c|}
		\hline 
		{\cellcolor[HTML]{BBDAFF}\textbf{Environment}} & {\cellcolor[HTML]{BBDAFF}\textbf{$\alpha$}} & {\cellcolor[HTML]{BBDAFF}\textbf{$\gamma$}} & {\cellcolor[HTML]{BBDAFF}\textbf{$\epsilon$-decay}} & {\cellcolor[HTML]{BBDAFF}\textbf{Batch Size}} & {\cellcolor[HTML]{BBDAFF}\textbf{Dataset-Size}} \\ 
		\hline 
		Deep Line Wars & 3e-5 &  0.98 &   0.005 & 16 & 1M \\ 
		\hline 
		Deep Maze & 3e-5 &  0.98 &   0.005 & 16 & 1M \\ 
		\hline 
		FlashRL:Multitask & 1e-4 &  0.98 &  0.005 & 16 & 1M \\ 
		\hline
		Deep RTS & 3e-5  &  0.98 &   0.005 & 16 & 1M \\ 
		\hline 
		Flappy Bird  & 2e-4 &  0.98 &   0.005 & 16 & 1M \\ 
		\hline 
	\end{tabular} 
	\caption{DQN: Hyper-parameters}
	\label{tbl:dqn:hyper-parameters}
\end{table}

Experiments conducted in this thesis are available at \url{http://github.com/UIA-CAIR}. 

\section{Experiments}

\begin{figure}[!htp]
	\centering
	\includegraphics[width=0.8\linewidth]{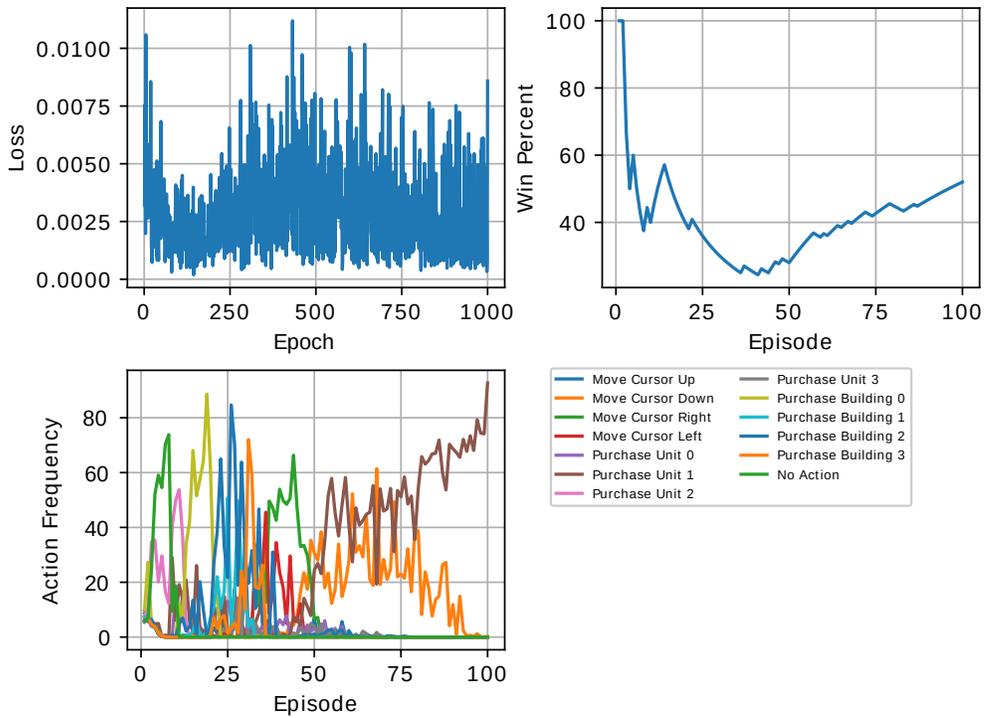}
	\caption{\gls{DQN}-\gls{CapsNet}: Deep Line Wars}
	\label{fig:dqn_capsnet:dlw}
\end{figure}

\begin{figure}[!htp]
	\centering
	\includegraphics[width=0.8\linewidth]{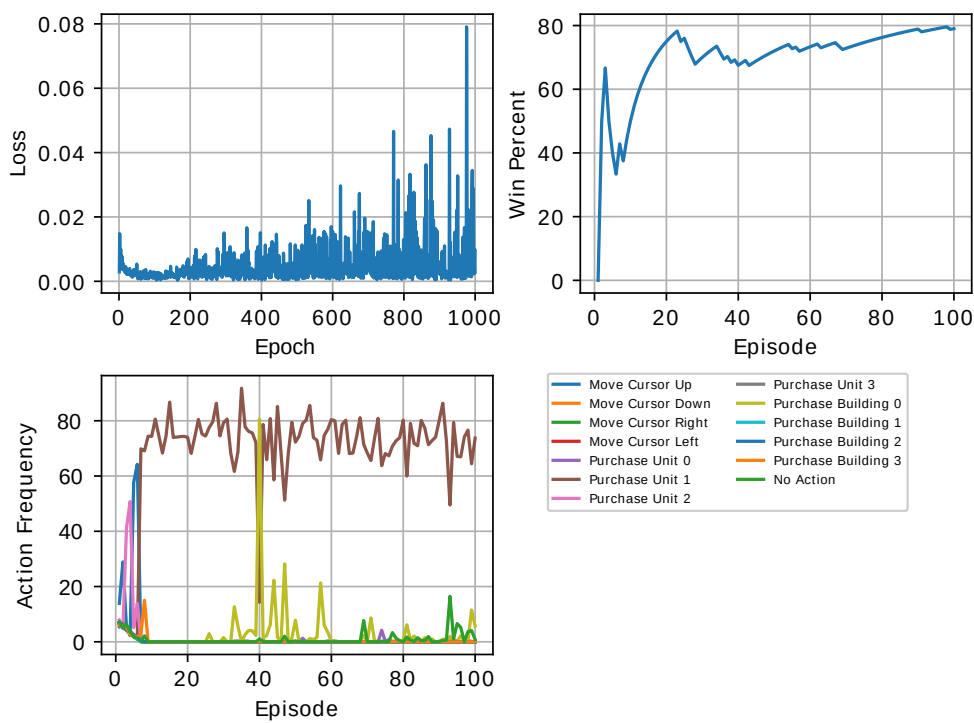}
	\caption{\gls{DQN}-\gls{ConvNet}: Deep Line Wars}
	\label{fig:dqn_cnn:dlw}
\end{figure}

\clearpage
\begin{figure}[!htp]
	\centering
	\includegraphics[width=0.8\linewidth]{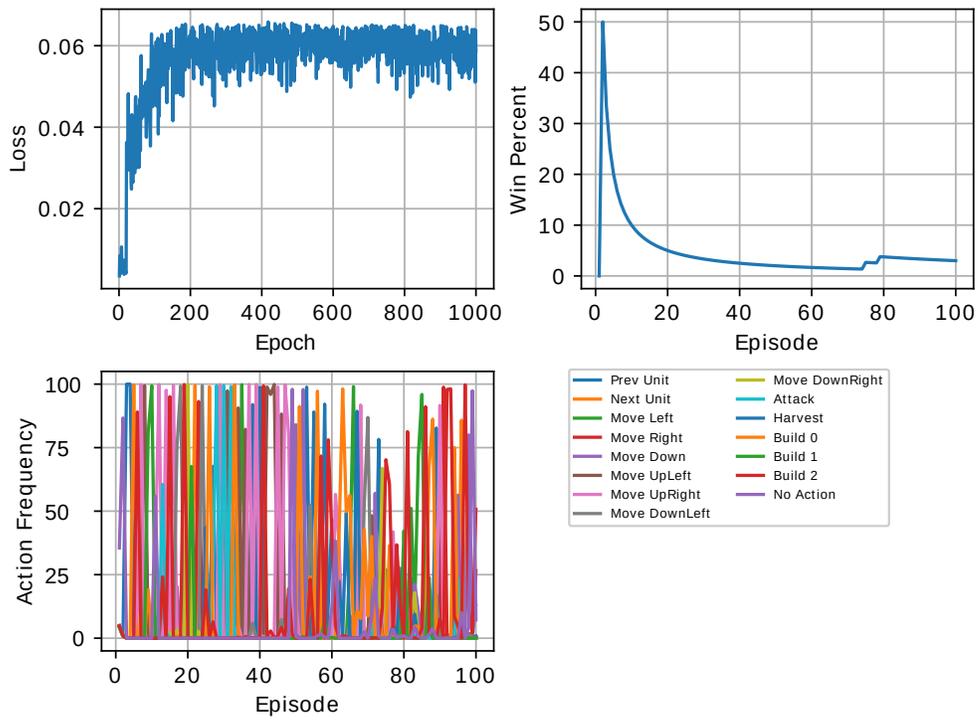}
	\caption{\gls{DQN}-\gls{CapsNet}: Deep RTS}
	\label{fig:dqn_capsnet:deeprts}
\end{figure}

\begin{figure}[!htp]
	\centering
	\includegraphics[width=0.8\linewidth]{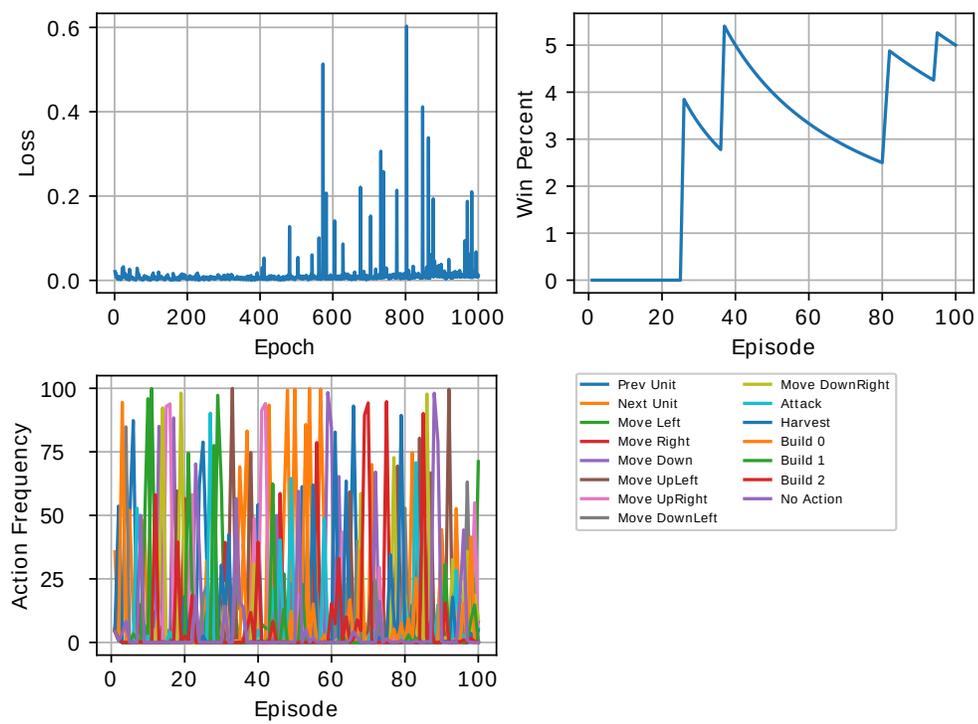}
	\caption{\gls{DQN}-\gls{ConvNet}: Deep RTS}
	\label{fig:dqn_cnn:deeprts}
\end{figure}

\clearpage
\begin{figure}[!htp]
	\centering
	\includegraphics[width=0.8\linewidth]{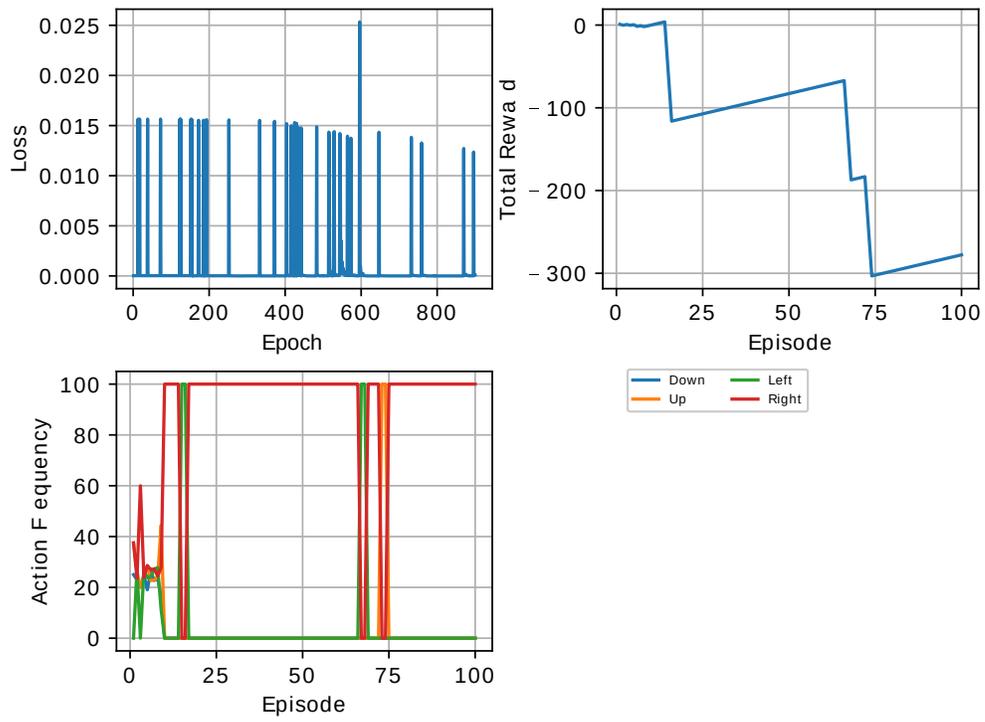}
	\caption{\gls{DQN}-\gls{CapsNet}: Deep Maze}
	\label{fig:dqn_capsnet:maze}
\end{figure}

\begin{figure}[!htp]
	\centering
	\includegraphics[width=0.8\linewidth]{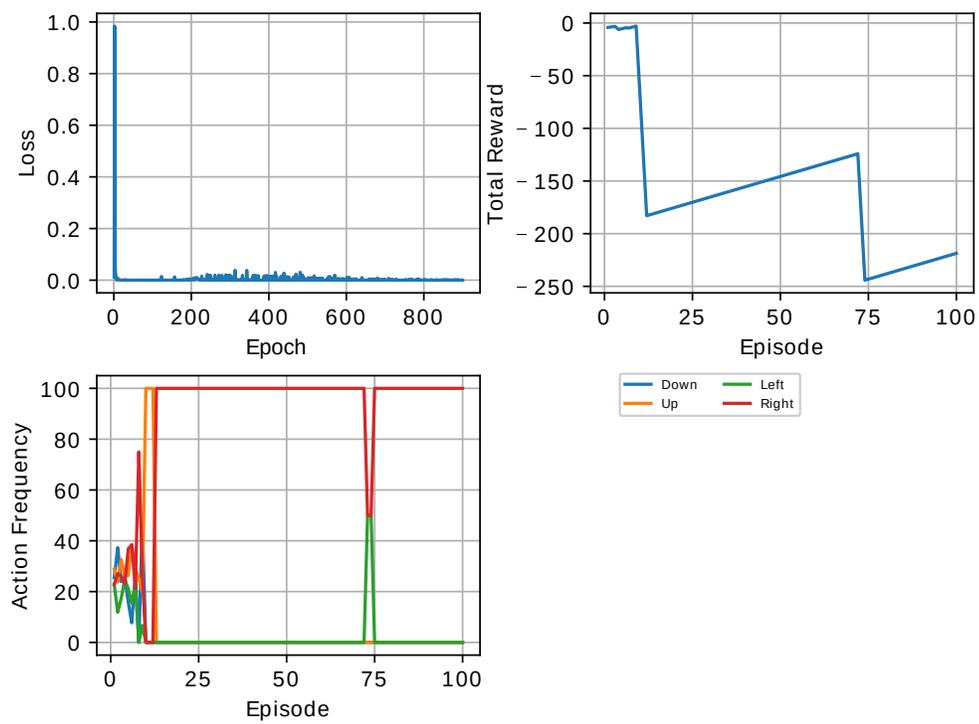}
	\caption{\gls{DQN}-\gls{ConvNet}: Deep Maze}
	\label{fig:dqn_cnn:maze}
\end{figure}



\clearpage
\begin{figure}[!htp]
	\centering
	\includegraphics[width=0.8\linewidth]{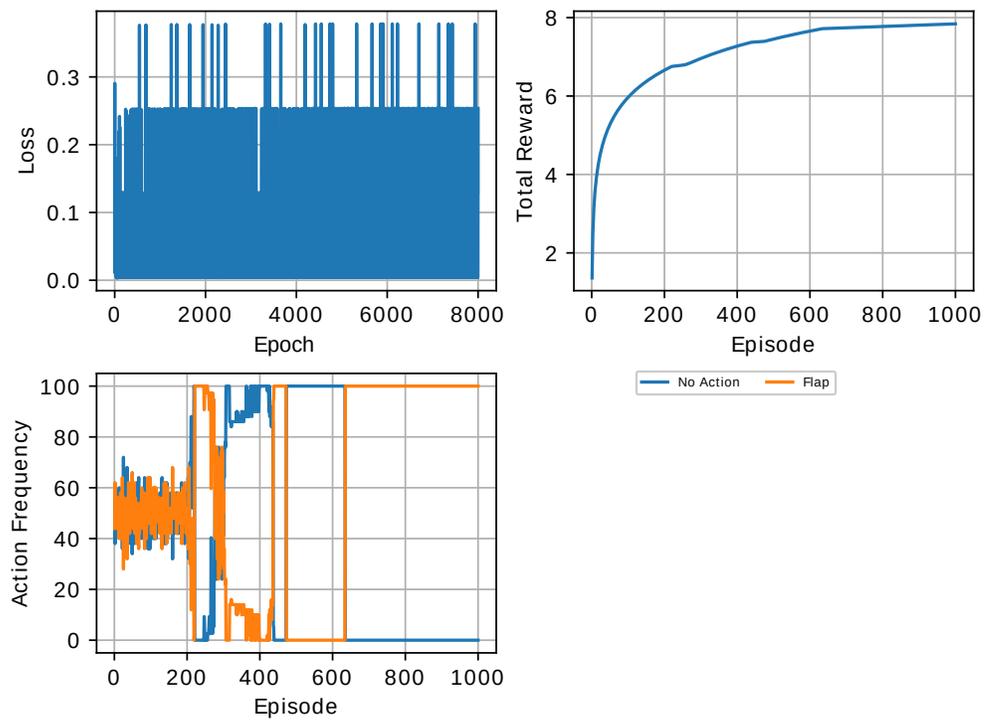}
	\caption{\gls{DQN}-\gls{CapsNet}: Flappy Bird}
	\label{fig:dqn_capsnet:flappybird}
\end{figure}

\begin{figure}[!htp]
	\centering
	\includegraphics[width=0.8\linewidth]{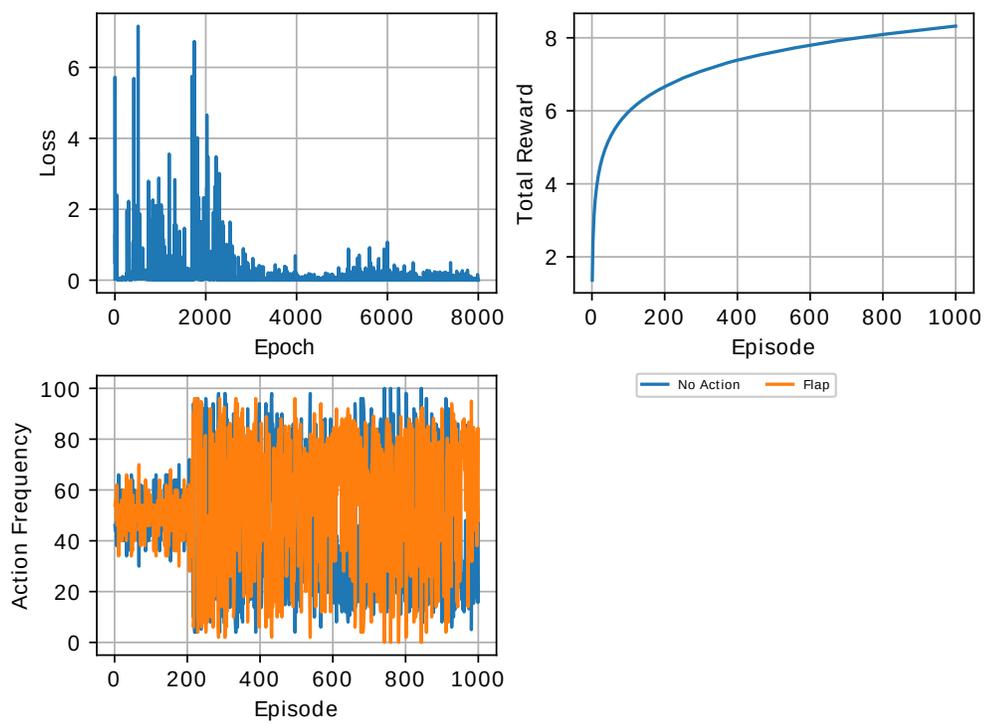}
	\caption{\gls{DQN}-\gls{ConvNet}: Flappy Bird}
	\label{fig:dqn_cnn:flappybird}
\end{figure}

\clearpage
\renewcommand{\sectionbreak}{}

\section{Deep Line Wars}
For Deep Line Wars, both agents illustrated relatively strong capabilities when it comes to exploiting game mechanics and finding the opponents weakness. The opponent is a random-play agent, that builds an uneven defense, sending units without any economic considerations. Figure~\ref{fig:dqn_capsnet:dlw} and Figure~\ref{fig:dqn_cnn:dlw} show that both agents find the opponents weakness to be defense. 

\begin{figure}[!htp]
	\centering
	\includegraphics[width=0.8\linewidth]{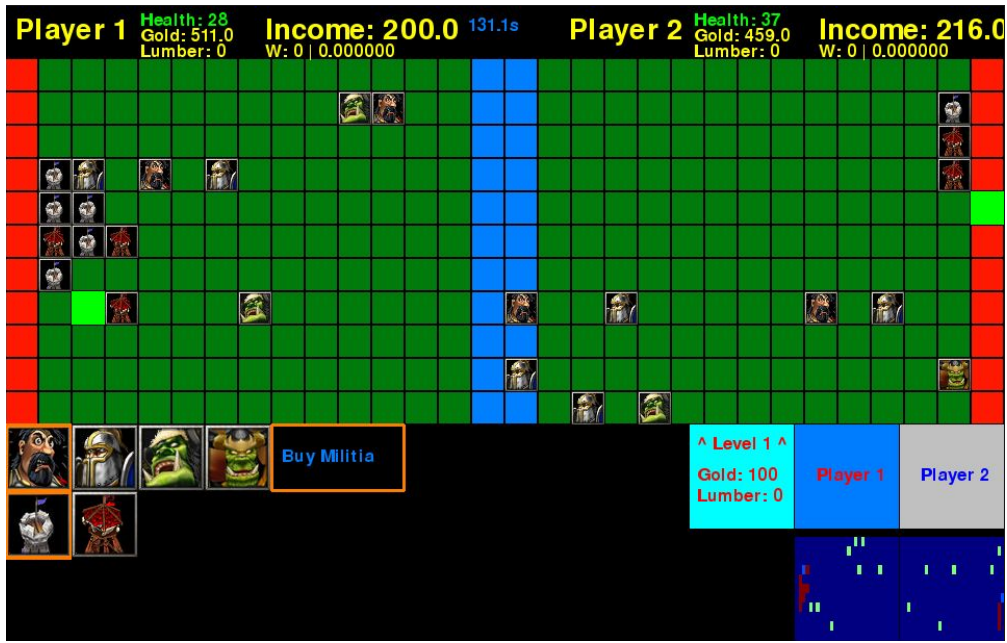}
	\caption{\gls{DQN}-\gls{CapsNet}: Agent building defensive due to low health in Deep Line Wars}
	\label{fig:dqn:dlw_ingame}
\end{figure}

Results shows that the game mechanics are not balanced, making the \textit{Purchase Unit 1} the obvious choice for offensive actions. This unit is strong enough to survive most defenses and does the most damage to the opponents health pool. The \gls{ConvNet} agent performs better in a period of 100 episodes, and both agents can master the random-play opponent.

\section{Deep RTS}
Deep RTS shows exciting results, where \gls{DQN}-\gls{CapsNet} starts at a low loss with a high total reward, slowly diverging in reward and loss. The results show that \gls{DQN}-\gls{CapsNet} and \gls{DQN}-\gls{ConvNet} perform comparably. It is not clear why \gls{DQN}-\gls{CapsNet} diverged, but the high action-space is a likely candidate. It is difficult to see any sense in the determination of action-state mapping, but some observations indicated that the agent favor gathering instead of military actions.

\section{Deep Maze}
The goal of Deep Maze is to find the shortest path from start to goal in a $25\times25$ labyrinth. Figure~\ref{fig:dqn_capsnet:maze} and Figure~\ref{fig:dqn_cnn:maze} shows that \gls{DQN}-\gls{CapsNet} had issues with the training stability. The algorithm is tested with several different hyper-parameter configurations, but there was no solution to remedy this. \gls{DQN}-\gls{ConvNet} did not indicate any issues during the training. Both agents had issues finding the shortest path, looking at the total reward, both agents had a negative score. For each move done after reaching the optimal move count, a negative reward is given the agent. Observations show that both agents have similar performance in this experiment.

\begin{figure}[!htp]
	\centering
	\includegraphics[width=0.5\linewidth]{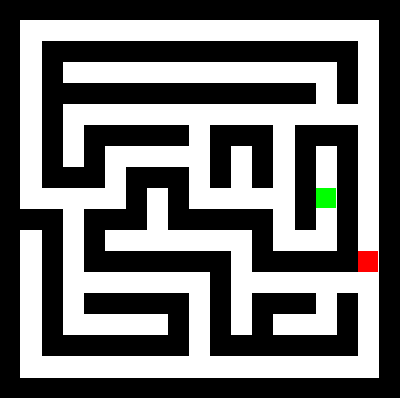}
	\caption{\gls{DQN}-\gls{CapsNet}: Agent attempting to find the shortest path in a $25\times25$ \textit{Deep Maze}}
	\label{fig:dqn:deepmaze_ingame}
\end{figure}

Figure~\ref{fig:dqn:deepmaze_ingame} illustrates an in-game image of the $25\times25$ map used in this experiment. The green square is the start area, while the red is the goal. The optimal path for this experiment is a series of 21 actions.

\section{FlashRL: Multitask}
For FlashRL: Multitask, the \gls{DQN}-\gls{CapsNet} was not able to compete with \gls{DQN}-\gls{ConvNet}. It was not able to learn how to control the first paddle. The results are for this reason not included for this environment. Refer to \textbf{Publication B} for results using \gls{DQN}-\gls{ConvNet}.

\section{Flappy Bird}
Flappy Bird is a difficult environment for an agent to master because the state-space is large due to the scrolling background. In literature, the training time for this environment is between one and four days. For this experiment, the agent trained for seven days, in the hope that both agents would converge. Figure~\ref{fig:dqn_capsnet:flappybird} and Figure~\ref{fig:dqn_cnn:flappybird} shows that both agents performed well, where \gls{DQN}-\gls{ConvNet} scored 0.4 points higher. For each pipe, the bird passes, 0.1 points are awarded to the total reward.

\section{Summary}
Looking at the results, it is clear that \gls{DQN}-\gls{CapsNet} overestimates actions for almost all environments. Instead of having a sensible distribution of actions, it often chooses to favor a particular move after a short period of training.

Recent state-of-the-art suggests that self-play using dueling methods may increase stability and performance in the long-term~\cite{Wang2015}, but this was not possible due to GPU memory limitations. It is clear that \gls{DQN}-\gls{CapsNet} can work for other tasks then image recognition, but there are still many challenges to solve before it can perform comparably to \gls{DQN}-\gls{ConvNet}. A significant issue is that Capsules do not scale well with several outputs (actions), resulting in a model that quickly becomes too large for the GPU memory to handle.
The upcoming paragraphs summarize the findings of the experiments conducted using \gls{DQN}-\gls{CapsNet} and \gls{DQN}-\gls{ConvNet}.

\textbf{Training Loss}
\\
An interesting observation during the training was that none of the models had a gradual decline in the loss during training. This may be because the state-space was quite large for all environments in the test-bed. Some investigation revealed that environments with sparse input had a more significant loss increase. By comparing Figure~\ref{fig:dqn_capsnet:dlw} and Figure~\ref{fig:dqn_capsnet:deeprts}, it is clear that Deep RTS has far more loss compared to Deep Line Wars when predicting the best Q-Value for an action. Since \gls{CapsNet} primarily detects objects, it is likely that the sudden jumps in loss (Figure~\ref{fig:dqn_capsnet:maze}) can be explained by several capsules changing its prediction vector at the same time. A possible improvement would be to decay the learning rate throughout the training period.
It is likely that the training loss issues can be managed for models with several new hyper-parameter configurations.

\textbf{Action Frequency}
\\
Results shows that \gls{CapsNet} tends to overestimate actions drastically in environments with few actions (Deep Maze and Flappy Bird). It is possible that this is because a Capsule looks for \textit{parts in the whole}. Since \gls{CapsNet} is positional invariant, one explanation may be that the model classifies states by looking for the existence of an object, instead of the likelihood of the best action. For Flappy Bird, the model determines that the agent should use \textit{Flap} as long as the bird exists in the input. For environments with large action-spaces, observations show a more consistent action frequency.

\textbf{Agent Performance}
\\
Table~\ref{tbl:dqn:summary} shows that \gls{DQN}-\gls{CapsNet} does indeed perform above random-play agents in selected environments, but falls behind compared to \gls{DQN}-\gls{ConvNet}. For all environments, a higher score is better. In Deep Line Wars, the reward increases as the agent keep surviving the game or defeat the enemy. The \gls{CapsNet} agent has approximately 57\% win chance while \gls{ConvNet} wins in 78\% of the games against a random-play agent. 

In Deep RTS, the accumulated score is measured during the first 600 seconds of the game. This is typically resource harvesting, as the agent was never able to create long-term strategies. In early training, \gls{CapsNet} accumulated far more resources then \gls{ConvNet}, but it gradually declined while training. This means that the model diverged from the optimal solution. It is likely that this is because the model starts to overestimate action Q-values. In comparison, results show that both models perform comparably while performing well beyond the capability of random-play agents.

In Deep Maze, none of the agents were able to find the optimal path to the goal. Additional experiments were conducted and showed better results for smaller mazes (9x9 and 11x11). For 25x25 the \gls{CapsNet} used on average 275 additional actions to reach the goal, while \gls{ConvNet} performed marginally better using 225 actions.

The \gls{CapsNet} agent is able to perform well in Flappy Bird. With only 0.4 points less then \gls{ConvNet}, it is clear that both agents perform at the same level of expertise. It is possible that the \gls{CapsNet} agent could achieve far better results if a solution is found for the Q-Value overestimation problem.

\begin{table}
	\centering
	\begin{tabular}{|l|c|c|c|}
		\hline 
		{\cellcolor[HTML]{BBDAFF}\textbf{Environment}} & {\cellcolor[HTML]{BBDAFF}\textbf{Random}} & {\cellcolor[HTML]{BBDAFF}\textbf{\gls{DQN}-\gls{CapsNet}}} & {\cellcolor[HTML]{BBDAFF}\textbf{\gls{DQN}-\gls{ConvNet}}} \\ 
		\hline 
		Deep Line Wars & 50 & 57  & 78  \\ 
		\hline 
		Deep RTS & 1.4 & 5.0 & 5.1  \\ 
		\hline 
		Deep Maze & -600 & -275 & -225   \\ 
		\hline 
		FlashRL: Multitask & 14 & N/A & 300  \\ 
		\hline
		Flappy Bird & 1.4 & 7.9 & 8.3 \\
		\hline 
	\end{tabular} 
	\caption{Comparison of \gls{DQN}-\gls{CapsNet}, \gls{DQN}-\gls{ConvNet}, and Random accumulative reward (Higher is better)}
	\label{tbl:dqn:summary}
\end{table}

\renewcommand{\sectionbreak}{\clearpage}

\chapter{Conclusion and Future Work}
\label{chap:conclusion}

\textit{This thesis conclusively shows that Capsules are viable to use in advanced game environments. It is further shown that capsules do not scale as well as convolutions, implying that capsule networks alone will not be able to play even more advanced games without improved scalability.}

This thesis has focused on \textbf{\ThesisTitle}. This work presents several new game environments that are tailored for research into \gls{RL} algorithms in the \gls{RTS} genre. This contribution could potentially lead to a groundbreaking performance in advanced game environments that could enable \gls{RL} agents to perform well in games like Starcraft II. The combination of Capsule Networks and Deep Q-Learning illustrated comparable results to regular \gls{ConvNet}s, in regards to stability, on the new learning platform. As a secondary goal, a generative model was implemented, \gls{CCDN}, which successfully generates future state representations in the majority of the test environments.

Since Capsule Networks are a novel research area that is in its early infant stage, more research is required to determine its capabilities in \gls{RL} for advanced game environments. This chapter presents the thesis conclusion and future work for the continuation of a PhD thesis in \gls{DRL}.

\section{Conclusion}
\label{sec:conclusion}

\textbf{Hypothesis 1: } \textit{Generative modeling using deep learning is capable of generating artificial training data for games with sufficient quality.}
\begin{itemize}
	\item[] Our work shows that it is indeed possible to generate artificial training data using deep learning. Our work shows that it is of sufficient quality to perform off-line training of deep neural networks.
\end{itemize}
\textbf{Hypothesis 2: } \textit{\gls{CapsNet} can be used in Deep Q-Learning with comparable performance to \gls{ConvNet} based models.}
\begin{itemize}
	\item[] The research shows that \gls{CapsNet} can be directly adapted to work with Deep Q-Learning, but the stability is inferior to regular \gls{ConvNet}. Some experiments show comparable results to \gls{ConvNet}s, but it is not clear how \gls{CapsNet}s do reasoning in an \gls{RL} environment.
\end{itemize}

\textbf{Goal 1: } \textit{Investigate the state-of-the-art research in the field of Deep Learning, and learn how Capsule Networks function internally.}
\begin{itemize}
	\item[] A thorough survey of the state-of-the-art in deep learning was outlined in Chapter~\ref{chap:sota}. Much of the performed work was inspired by previous research, which enabled several exciting discoveries in \gls{RL}. Results show that it is possible to combine \gls{CapsNet} with other algorithms.
\end{itemize}

\textbf{Goal 2: } \textit{Design and develop game environments that can be used for research into \gls{RL} agents for the \gls{RTS} game genre.}
\begin{itemize}
	\item[] The thesis outline four new game environments that target research into \gls{RL} agents for \gls{RTS} games. 
	
	Deep RTS is a Warcraft II clone that is suited for an agent of high-quality play. It requires the agent to do actions in a high-dimensional environment that is continuously moving. Since the Deep RTS state is of such high-dimension, it is still not feasible to master this environment.
	
	Deep Line Wars was created to enable research on a simpler scale, this enabled research into some of the \gls{RTS} aspects, found in Deep RTS.
	
	To simplify it even further, Deep Maze was created to only account for trivial state interpretations. Flash RL was created as a side project, enabling research into a vast library of Flash games.
	
	Together, these game environments create a platform that allows for in-depth research into \gls{RL} problems in the \gls{RTS} game genre.
\end{itemize}

\textbf{Goal 3: } \textit{Research generative modeling and implement an experimental architecture for generating artificial training data for games.}
\begin{itemize}
	\item[] \gls{CCDN} is introduced as a novel architecture for generating artificial future states from a game, using present state and action to learn the transition function of an environment. Early experimental results are presented in this work, showing that it has potential to successfully train a neural network based model.
\end{itemize}

\textbf{Goal 4: } \textit{Research the novel \gls{CapsNet} architecture for MNIST classification and apply this to \gls{RL} problems.}
\begin{itemize}
	\item[] Section~\ref{sec:solutions:capsnet} outlines the research into \gls{CapsNet} in scenarios that are different from the MNIST experiments conducted by Sabour et al\cite{Sabour2017}. The objective of Capsules is redefined so that it could work for \gls{RL} related problems.
\end{itemize}

\textbf{Goal 5: } \textit{Combine Deep-Q Learning and \gls{CapsNet} and perform experiments on environments from Achievement 2.}
\begin{itemize}
	\item[] In Chapter~\ref{chap:results:dqn}, \gls{DQN} and \gls{CapsNet} were successfully combined and illustrated that it has the potential to perform well in several advanced game environments. Although these results only show minor agent intelligence, it is an excellent beginning for further research into this type of deep models.
\end{itemize}

\textbf{Goal 6: } \textit{Combine the elements of Goal 3 and Goal 5. The goal is to train an \gls{RL} agent with artificial training data successfully.}
\begin{itemize}
	\item[] Results shows that training data generated with \gls{CCDN} can be used in conjunction with real data to train an \gls{DQN} algorithm successfully.
\end{itemize}

All of the goals defined in the scope of this thesis were accomplished. Although the results are not astounding for all goals, it enables further research into several new deep learning fields. The work presented in this thesis enables further research into \gls{CapsNet} based \gls{RL} in advanced game environments. Because of the new learning platform, researchers can better perform research into \gls{RTS} games. It is possible that the work from this thesis could be the foundation for novel \gls{RL} algorithms in the future.

\section{Future Work}
\label{sec:conclusion:future_work}

\textbf{Environments}
\begin{enumerate}

	\item Continue work on Flash RL, enabling it to replace OpenAI Universe Flash.
	\item Propose partnership with ELF\footnote{ELF Source-code: \url{https://github.com/facebookresearch/ELF}} and implement Deep RTS and Deep Line Wars into ELF.
	\item Develop a full-fledged platform that expands beyond gym-cair.
	\item Implement Image state-representation for Deep RTS.
\end{enumerate}

\textbf{Generative Modeling}
\begin{enumerate}
	\item Additional experiments with hyper-parameters with the existing models.
	\item Attempt to stabilize training.
	\item Investigate if it is possible to use adversarial methods to train the generative model.
	\item Identify and solve the issue with the loss function in \gls{CCDN}.
\end{enumerate}

\textbf{Deep Capsule Q-Learning}
\begin{enumerate}
	\item Improve stability of current architecture, enabling less data. preprocessing for the algorithm to function.
	\item Improve the scalability of Capsules for large action spaces.
	\item Do additional experiments with multiple configurations to find the cause of the training instability.
	\item More research into combining Capsules with \gls{RL} algorithms.
\end{enumerate}

\textbf{Planned Publications\footnote{Proposed Publication titles may change in final versions}}
\begin{enumerate}
	\item Deep RTS: A Real-time Strategy game for Reinforcement Learning.
	\item CCDN: Towards infinite training data using generative models.
	\item DCQN: Using Capsules in Deep Q-Learning.
\end{enumerate}

\cleardoublepage
\renewcommand{\bibname}{References}
\bibliography{library}

\begin{thebibliography}{10}

\bibitem{Andersen2017a}
Per-Arne Andersen, Morten Goodwin, and Ole-Christoffer Granmo.
\newblock {FlashRL: A Reinforcement Learning Platform for Flash Games}.
\newblock {\em Norsk Informatikkonferanse}, 2017.

\bibitem{Andersen2017}
Per~Arne Andersen, Morten Goodwin, and Ole~Christoffer Granmo.
\newblock {Towards a deep reinforcement learning approach for tower line wars}.
\newblock In {\em Lecture Notes in Computer Science (including subseries
  Lecture Notes in Artificial Intelligence and Lecture Notes in
  Bioinformatics)}, volume 10630 LNAI, pages 101--114, 2017.

\bibitem{Beattie2016}
Charles Beattie, Joel~Z. Leibo, Denis Teplyashin, Tom Ward, Marcus Wainwright,
  Heinrich K{\"{u}}ttler, Andrew Lefrancq, Simon Green, V{\'{i}}ctor
  Vald{\'{e}}s, Amir Sadik, Julian Schrittwieser, Keith Anderson, Sarah York,
  Max Cant, Adam Cain, Adrian Bolton, Stephen Gaffney, Helen King, Demis
  Hassabis, Shane Legg, and Stig Petersen.
\newblock {DeepMind Lab}.
\newblock dec 2016.

\bibitem{Bellemare2012}
Marc~G. Bellemare, Yavar Naddaf, Joel Veness, and Michael Bowling.
\newblock {The arcade learning environment: An evaluation platform for general
  agents}.
\newblock {\em IJCAI International Joint Conference on Artificial
  Intelligence}, 2015-Janua:4148--4152, 2015.

\bibitem{Brockman2016}
Greg Brockman, Vicki Cheung, Ludwig Pettersson, Jonas Schneider, John Schulman,
  Jie Tang, and Wojciech Zaremba.
\newblock {OpenAI Gym}.
\newblock jun 2016.

\bibitem{Chen2015}
Kevin Chen.
\newblock {Deep Reinforcement Learning for Flappy Bird}.
\newblock page~6, 2015.

\bibitem{VanHasselt2015}
Wenliang Chen, Min Zhang, Yue Zhang, and Xiangyu Duan.
\newblock {Exploiting meta features for dependency parsing and part-of-speech
  tagging}.
\newblock {\em Artificial Intelligence}, 230:173--191, sep 2016.

\bibitem{Ciardo1997}
Gianfranco Ciardo and Andrew~S. Miner.
\newblock {Storage alternatives for large structured state spaces}.
\newblock In {\em Lecture Notes in Computer Science (including subseries
  Lecture Notes in Artificial Intelligence and Lecture Notes in
  Bioinformatics)}, volume 1245, pages 44--57, 1997.

\bibitem{Dai2017}
Bo~Dai, Sanja Fidler, Raquel Urtasun, and Dahua Lin.
\newblock {Towards Diverse and Natural Image Descriptions via a Conditional
  GAN}.
\newblock mar 2017.

\bibitem{doya2002}
Kenji Doya, Kazuyuki Samejima, Ken-ichi Katagiri, and Mitsuo Kawato.
\newblock {Multiple model-based reinforcement learning}.
\newblock {\em Neural computation}, 14(6):1347--1369, 2002.

\bibitem{Even-dar2003}
Eyal Even-dar, Shie Mannor, and Yishay Mansour.
\newblock {Action Elimination and Stopping Conditions for Reinforcement
  Learning}.
\newblock {\em Icml}, 7:1079--1105, 2003.

\bibitem{Gu2015}
Jiuxiang Gu, Zhenhua Wang, Jason Kuen, Lianyang Ma, Amir Shahroudy, Bing Shuai,
  Ting Liu, Xingxing Wang, Gang Wang, Jianfei Cai, and Tsuhan Chen.
\newblock {Recent advances in convolutional neural networks}.
\newblock {\em Pattern Recognition}, dec 2017.

\bibitem{Chen}
Shixiang Gu, Ethan Holly, Timothy Lillicrap, and Sergey Levine.
\newblock {Deep reinforcement learning for robotic manipulation with
  asynchronous off-policy updates}, 2017.

\bibitem{Gu2016}
Shixiang Gu, Timothy Lillicrap, Ilya Sutskever, and Sergey Levine.
\newblock {Continuous Deep Q-Learning with Model-based Acceleration}.
\newblock mar 2016.

\bibitem{Gupta2016}
Abhishek Gupta, Clemens Eppner, Sergey Levine, and Pieter Abbeel.
\newblock {Learning dexterous manipulation for a soft robotic hand from human
  demonstrations}.
\newblock In {\em IEEE International Conference on Intelligent Robots and
  Systems}, volume 2016-Novem, pages 3786--3793, 2016.

\bibitem{Hausknecht2015}
Matthew Hausknecht and Peter Stone.
\newblock {Deep Recurrent Q-Learning for Partially Observable MDPs}.
\newblock jul 2015.

\bibitem{Hinton2006}
Geoffrey~E. Hinton, Simon Osindero, and Yee-Whye Teh.
\newblock {A Fast Learning Algorithm for Deep Belief Nets}.
\newblock {\em Neural Computation}, 18(7):1527--1554, 2006.

\bibitem{Hunt2004}
Harold~L Hunt and I~I~Jon Turney.
\newblock {Cygwin/X Contributor's Guide}.
\newblock 2004.

\bibitem{Goodfellow2016}
Aaron~Courville {Ian Goodfellow, Yoshua Bengio}.
\newblock {\em {Deep Learning}}, volume 521.
\newblock MIT Press, 2017.

\bibitem{Johnson2016}
Matthew Johnson, Katja Hofmann, Tim Hutton, and David Bignell.
\newblock {The malmo platform for artificial intelligence experimentation}.
\newblock {\em IJCAI International Joint Conference on Artificial
  Intelligence}, 2016-Janua:4246--4247, 2016.

\bibitem{Kaelbling1996}
Leslie~Pack Kaelbling, Michael~L. Littman, and Andrew~W. Moore.
\newblock {Reinforcement learning: A survey}.
\newblock {\em Journal of Artificial Intelligence Research}, 4:237--285, 1996.

\bibitem{LeCun1991}
I~Kanter, Y~LeCun, and S~Solla.
\newblock {Second-order properties of error surfaces: learning time and
  generalization}.
\newblock {\em Advances in Neural Information Processing Systems (NIPS 1990)},
  3:918--924, 1991.

\bibitem{Goodfellow2014}
Shohei Kinoshita, Takahiro Ogawa, and Miki Haseyama.
\newblock {LDA-based music recommendation with CF-based similar user
  selection}.
\newblock {\em 2015 IEEE 4th Global Conference on Consumer Electronics, GCCE
  2015}, pages 215--216, jun 2016.

\bibitem{Konidaris2006}
George Konidaris and Andrew~G. Barto.
\newblock {Autonomous shaping}.
\newblock {\em International Conference on Machine Learning}, pages 489--496,
  2006.

\bibitem{Krizhevsky2012}
Alex Krizhevsky, Ilya Sutskever, and Geoffrey~E Hinton.
\newblock {ImageNet Classification with Deep Convolutional Neural Networks},
  2012.

\bibitem{Watkins1989}
Ben~J.A. Kr{\"{o}}se.
\newblock {\em {Learning from delayed rewards}}.
\newblock PhD thesis, King's College, Cambridge, UK, 1995.

\bibitem{LeCun1989}
Y.~LeCun, B.~Boser, J.~S. Denker, D.~Henderson, R.~E. Howard, W.~Hubbard, and
  L.~D. Jackel.
\newblock {Backpropagation Applied to Handwritten Zip Code Recognition}.
\newblock {\em Neural Computation}, 1(4):541--551, dec 1989.

\bibitem{Lecun1998}
Yann LeCun, L{\'{e}}on Bottou, Yoshua Bengio, and Patrick Haffner.
\newblock {Gradient-based learning applied to document recognition}.
\newblock {\em Proceedings of the IEEE}, 86(11):2278--2323, 1998.

\bibitem{Lee2015}
Chen-Yu Lee, Patrick~W. Gallagher, and Zhuowen Tu.
\newblock {Generalizing Pooling Functions in Convolutional Neural Networks:
  Mixed, Gated, and Tree}.
\newblock sep 2015.

\bibitem{Li2017}
Yuxi Li.
\newblock {Deep Reinforcement Learning: An Overview}.
\newblock {\em arXiv}, pages 1--30, 2017.

\bibitem{Lin1993}
L~J Lin.
\newblock {Reinforcement Learning for Robots Using Neural Networks}.
\newblock {\em Report, CMU}, pages 1--155, 1993.

\bibitem{Mnih2016}
Bj{\"{o}}rn Lindstr{\"{o}}m, Ida Selbing, Tanaz Molapour, and Andreas Olsson.
\newblock {Racial Bias Shapes Social Reinforcement Learning}.
\newblock {\em Psychological Science}, 25(3):711--719, feb 2014.

\bibitem{Radford2015}
Chunhui Liu, Aayush Bansal, Victor Fragoso, and Deva Ramanan.
\newblock {Do Convolutional Neural Networks act as Compositional Nearest
  Neighbors?}
\newblock {\em arXiv}, pages 1--15, 2017.

\bibitem{Mirowski2016}
Piotr Mirowski, Razvan Pascanu, Fabio Viola, Hubert Soyer, Andrew~J. Ballard,
  Andrea Banino, Misha Denil, Ross Goroshin, Laurent Sifre, Koray Kavukcuoglu,
  Dharshan Kumaran, and Raia Hadsell.
\newblock {Learning to Navigate in Complex Environments}.
\newblock nov 2016.

\bibitem{Mirza2014}
Mehdi Mirza and Simon Osindero.
\newblock {Conditional Generative Adversarial Nets}.
\newblock nov 2014.

\bibitem{Mnih2013}
Volodymyr Mnih, Koray Kavukcuoglu, David Silver, Alex Graves, Ioannis
  Antonoglou, Daan Wierstra, and Martin Riedmiller.
\newblock {Playing Atari with Deep Reinforcement Learning}.
\newblock dec 2013.

\bibitem{Mnih2015}
Volodymyr Mnih, Koray Kavukcuoglu, David Silver, Andrei~A. Rusu, Joel Veness,
  Marc~G. Bellemare, Alex Graves, Martin Riedmiller, Andreas~K. Fidjeland,
  Georg Ostrovski, Stig Petersen, Charles Beattie, Amir Sadik, Ioannis
  Antonoglou, Helen King, Dharshan Kumaran, Daan Wierstra, Shane Legg, and
  Demis Hassabis.
\newblock {Human-level control through deep reinforcement learning}.
\newblock {\em Nature}, 518(7540):529--533, feb 2015.

\bibitem{Moravcik2017}
Matej Morav{\v{c}}{\'{i}}k, Martin Schmid, Neil Burch, Viliam Lis{\'{y}},
  Dustin Morrill, Nolan Bard, Trevor Davis, Kevin Waugh, Michael Johanson, and
  Michael Bowling.
\newblock {DeepStack: Expert-Level Artificial Intelligence in No-Limit Poker}.
\newblock jan 2017.

\bibitem{Nair2010}
Fernando Naclerio, Marco Seijo-Bujia, Eneko Larumbe-Zabala, and Conrad~P.
  Earnest.
\newblock {Carbohydrates alone or mixing with beef or whey protein promote
  similar training outcomes in resistance training males: A double-blind,
  randomized controlled clinical trial}.
\newblock {\em International Journal of Sport Nutrition and Exercise
  Metabolism}, 27(5):408--420, 2017.

\bibitem{Simonyan2014}
Bruno~A. Olshausen and David~J. Field.
\newblock {Sparse coding of sensory inputs}.
\newblock {\em Current Opinion in Neurobiology}, 14(4):481--487, sep 2004.

\bibitem{Kempka2016}
Etienne Perot, Maximilian Jaritz, Marin Toromanoff, and Raoul~De Charette.
\newblock {End-to-End Driving in a Realistic Racing Game with Deep
  Reinforcement Learning}.
\newblock {\em IEEE Computer Society Conference on Computer Vision and Pattern
  Recognition Workshops}, 2017-July:474--475, may 2017.

\bibitem{Pralyt1994}
Laurent Praly and Yuan Wang.
\newblock {Stabilization in spite of matched unmodeled dynamics and an
  equivalent definition of input-to-state stability}.
\newblock {\em Mathematics of Control, Signals, and Systems}, 9(1):1--33, 1996.

\bibitem{Wang2015}
Carlos Ramirez-Perez and Victor Ramos.
\newblock {SDN meets SDR in self-organizing networks: Fitting the pieces of
  network management}.
\newblock {\em IEEE Communications Magazine}, 54(1):48--57, nov 2016.

\bibitem{Rumelhart1986}
David~E. Rumelhart, Geoffrey~E. Hinton, and Ronald~J. Williams.
\newblock {Learning representations by back-propagating errors}.
\newblock {\em Nature}, 323(6088):533--536, oct 1986.

\bibitem{Sabour2017}
Sara Sabour, Nicholas Frosst, and Geoffrey~E Hinton.
\newblock {Dynamic Routing Between Capsules}.
\newblock {\em Nips}, oct 2017.

\bibitem{Salimans2016}
Tim Salimans, Ian Goodfellow, Wojciech Zaremba, Vicki Cheung, Alec Radford, and
  Xi~Chen.
\newblock {Improved Techniques for Training GANs}.
\newblock jun 2016.

\bibitem{Samek2015}
Wojciech Samek, Alexander Binder, Gregoire Montavon, Sebastian Lapuschkin, and
  Klaus~Robert Muller.
\newblock {Evaluating the Visualization of What a Deep Neural Network Has
  Learned}.
\newblock {\em IEEE Transactions on Neural Networks and Learning Systems}, sep
  2016.

\bibitem{Scherer2010}
Dominik Scherer, Andreas M{\"{u}}ller, and Sven Behnke.
\newblock {Evaluation of pooling operations in convolutional architectures for
  object recognition}.
\newblock In {\em Lecture Notes in Computer Science (including subseries
  Lecture Notes in Artificial Intelligence and Lecture Notes in
  Bioinformatics)}, volume 6354 LNCS, pages 92--101, 2010.

\bibitem{Silver2016}
David Silver, Aja Huang, Chris~J. Maddison, Arthur Guez, Laurent Sifre, George
  {Van Den Driessche}, Julian Schrittwieser, Ioannis Antonoglou, Veda
  Panneershelvam, Marc Lanctot, Sander Dieleman, Dominik Grewe, John Nham, Nal
  Kalchbrenner, Ilya Sutskever, Timothy Lillicrap, Madeleine Leach, Koray
  Kavukcuoglu, Thore Graepel, and Demis Hassabis.
\newblock {Mastering the game of Go with deep neural networks and tree search}.
\newblock {\em Nature}, 529(7587):484--489, 2016.

\bibitem{Silver2017}
David Silver, Thomas Hubert, Julian Schrittwieser, Ioannis Antonoglou, Matthew
  Lai, Arthur Guez, Marc Lanctot, Laurent Sifre, Dharshan Kumaran, Thore
  Graepel, Timothy Lillicrap, Karen Simonyan, and Demis Hassabis.
\newblock {Mastering Chess and Shogi by Self-Play with a General Reinforcement
  Learning Algorithm}.
\newblock dec 2017.

\bibitem{Silver2017a}
David Silver, Julian Schrittwieser, Karen Simonyan, Ioannis Antonoglou, Aja
  Huang, Arthur Guez, Thomas Hubert, Lucas Baker, Matthew Lai, Adrian Bolton,
  Yutian Chen, Timothy Lillicrap, Fan Hui, Laurent Sifre, George {Van Den
  Driessche}, Thore Graepel, and Demis Hassabis.
\newblock {Mastering the game of Go without human knowledge}.
\newblock {\em Nature}, 550(7676):354--359, 2017.

\bibitem{Sutton1990}
Richard~S Sutton and Andrew~G Barto.
\newblock {Time-Derivative Models of Pavlovian Reinforcement}.
\newblock {\em Learning and Computational Neuroscience: Foundations of Adaptive
  Networks}, (Mowrer 1960):497--537, 1990.

\bibitem{Sutton1998}
R.S. Sutton and A.G. Barto.
\newblock {\em {Reinforcement Learning: An Introduction}}, volume~9.
\newblock MIT Press, 1998.

\bibitem{Bellman1957}
W.W. Swart, C.E. Gearing, and T.~Var.
\newblock {\em {A dynamic programming—integer programming algorithm for
  allocating touristic investments}}, volume~27.
\newblock 1972.

\bibitem{Techtonik2015}
Techtonik.
\newblock python-vnc-viewer.
\newblock https://github.com/techtonik/python-vnc-viewer, 2015.

\bibitem{Tesauro1994}
Gerald Tesauro.
\newblock {TD-Gammon, a Self-Teaching Backgammon Program, Achieves Master-Level
  Play}.
\newblock {\em Neural Computation}, 6(2):215--219, 1994.

\bibitem{Tesauro1995}
Gerald Tesauro.
\newblock {Temporal difference learning and TD-Gammon}.
\newblock {\em Communications of the ACM}, 38(3):58--68, 1995.

\bibitem{Tian2017}
Yuandong Tian, Qucheng Gong, Wenling Shang, Yuxin Wu, and C.~Lawrence Zitnick.
\newblock {ELF: An Extensive, Lightweight and Flexible Research Platform for
  Real-time Strategy Games}.
\newblock jul 2017.

\bibitem{VanSeijen2017}
Harm van Seijen, Mehdi Fatemi, Joshua Romoff, Romain Laroche, Tavian Barnes,
  and Jeffrey Tsang.
\newblock {Hybrid Reward Architecture for Reinforcement Learning}.
\newblock jun 2017.

\bibitem{Vinyals2017}
Oriol Vinyals, Timo Ewalds, Sergey Bartunov, Petko Georgiev, Alexander~Sasha
  Vezhnevets, Michelle Yeo, Alireza Makhzani, Heinrich K{\"{u}}ttler, John
  Agapiou, Julian Schrittwieser, John Quan, Stephen Gaffney, Stig Petersen,
  Karen Simonyan, Tom Schaul, Hado van Hasselt, David Silver, Timothy
  Lillicrap, Kevin Calderone, Paul Keet, Anthony Brunasso, David Lawrence,
  Anders Ekermo, Jacob Repp, and Rodney Tsing.
\newblock {StarCraft II: A New Challenge for Reinforcement Learning}.
\newblock aug 2017.

\bibitem{Wan2017}
Fang Wan and Chaoyang Song.
\newblock {Logical Learning Through a Hybrid Neural Network with Auxiliary
  Inputs}.
\newblock may 2017.

\bibitem{Duan2016}
Jiang Wang, Yang Song, Thomas Leung, Chuck Rosenberg, Jingbin Wang, James
  Philbin, Bo~Chen, and Ying Wu.
\newblock {Learning fine-grained image similarity with deep ranking}.
\newblock {\em Proceedings of the IEEE Computer Society Conference on Computer
  Vision and Pattern Recognition}, pages 1386--1393, apr 2014.

\bibitem{He2015}
Songtao Wu, Shenghua Zhong, and Yan Liu.
\newblock {Deep residual learning for image steganalysis}.
\newblock {\em Multimedia Tools and Applications}, pages 1--17, dec 2017.

\bibitem{Xi2017}
Edgar Xi, Selina Bing, and Yang Jin.
\newblock {Capsule Network Performance on Complex Data}.
\newblock dec 2017.

\bibitem{Yann1998}
Lecun Yann.
\newblock {Efficient backprop}.
\newblock {\em Neural networks: tricks of the trade}, 53(9):1689--1699, 1998.

\end{thebibliography}
\bibliographystyle{plain}
\addcontentsline{toc}{chapter}{\numberline{}References}

\appendix
\chapter*{Appendices}
\addcontentsline{toc}{chapter}{Appendices}
\renewcommand{\thesection}{\Alph{section}}

\section{Hardware Specification}
\label{appendix:hwspec}
	\begin{table}[H]
	\begin{tabular}{|l*{1}{r}|r}
		\hline
		Operating System & Ubuntu 17.10  \\
		\hline
		Processor            & Intel i7-7700K \\
		\hline
		Memory           & 64GB DDR4  \\
		\hline
		Graphics     & 1x NVIDIA GeForce 1080TI   \\
		\hline
	\end{tabular}
\end{table}

\part{Publications}
\setcounter{chapter}{0}
\renewcommand{\chaptername}{Paper}
\renewcommand{\thechapter}{\Alph{chapter}}

\chapter{Towards a Deep Reinforcement Learning Approach for Tower Line Wars}
\publication{Towards a Deep Reinforcement Learning Approach for Tower Line Wars}
\includepdf[pages=-]{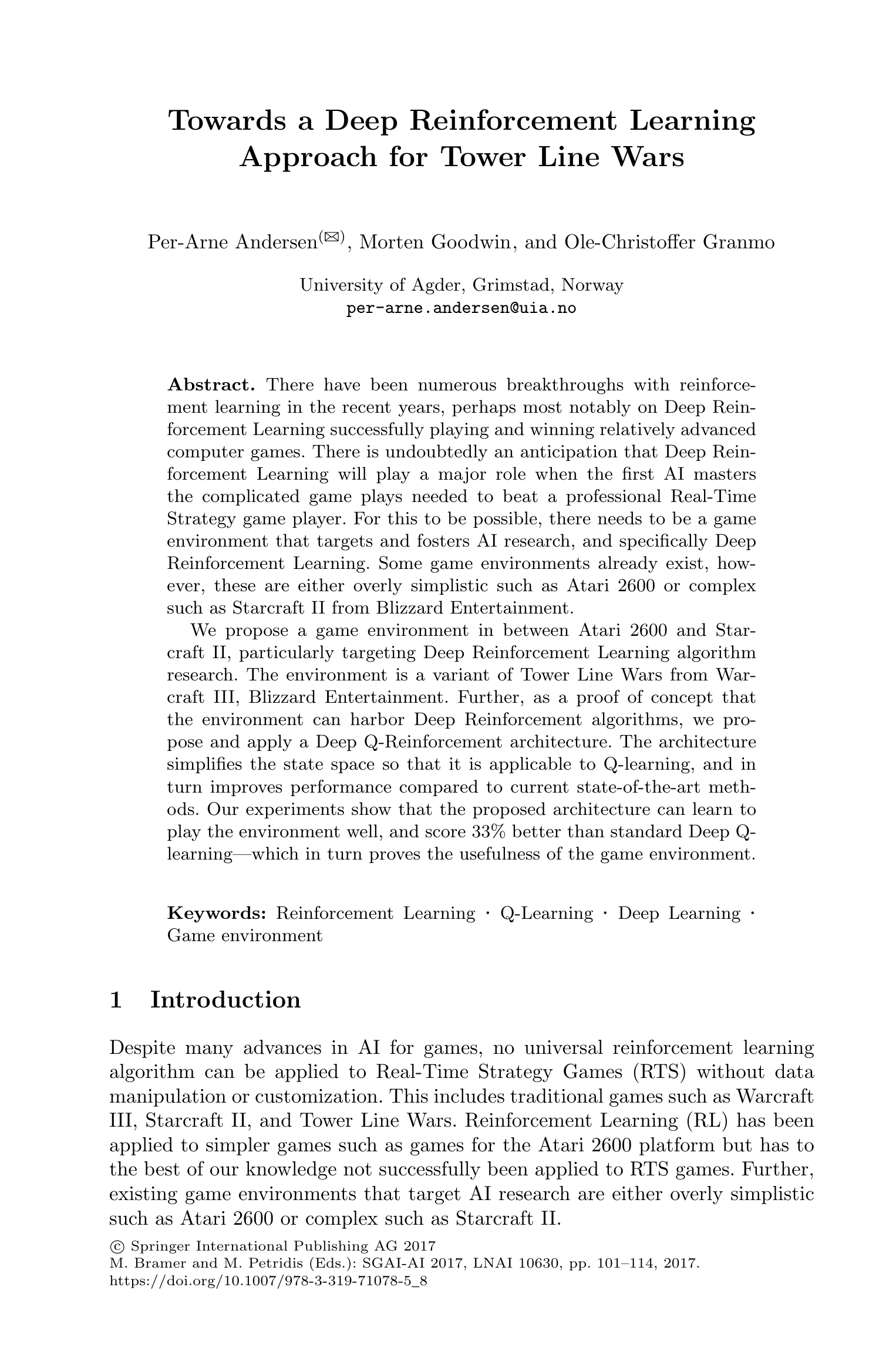}

\chapter{FlashRL: A Reinforcement Learning Platform for Flash Games}
\publication{FlashRL: A Reinforcement Learning Platform for Flash Games}
\includepdf[pages=-]{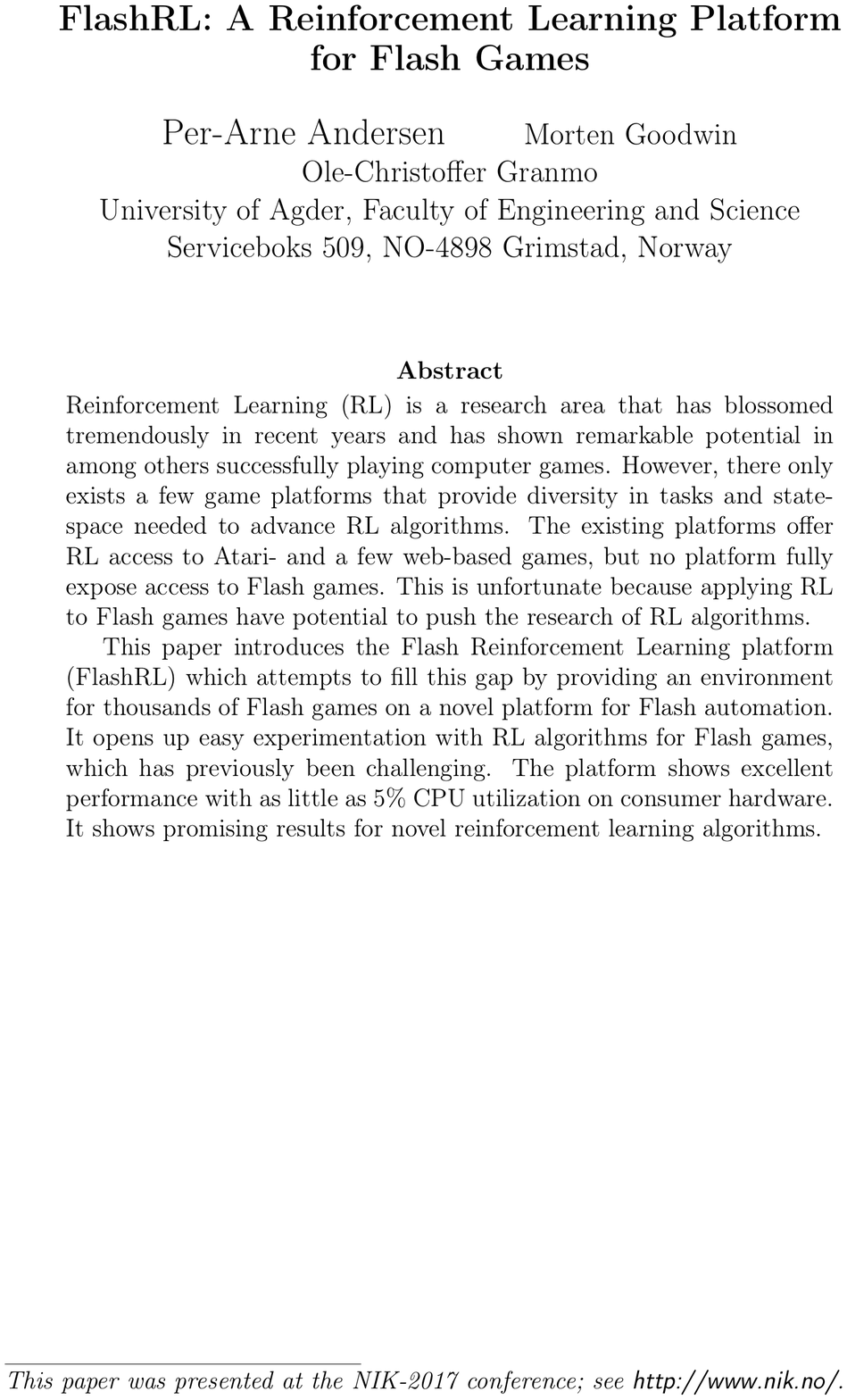}

\makebackcover

\end{document}